\documentclass[journal]{IEEEtran}

\usepackage{graphicx}
\usepackage{array}
\usepackage{makecell}
\usepackage{multirow}
\usepackage{amsmath,amsfonts}
\usepackage[normalem]{ulem}
\usepackage[most]{tcolorbox}
\usepackage{colortbl}
\usepackage{enumitem}
\usepackage{url}
\usepackage{enumerate}
\usepackage{booktabs}
\usepackage{xspace}

\usepackage{color}
\usepackage{threeparttable}
\usepackage{ulem}
\usepackage{amssymb}

\usepackage{float}
\usepackage[ruled,linesnumbered]{algorithm2e}
\usepackage{balance}
\usepackage[caption=false,font=normalsize,labelfont=sf,textfont=sf]{subfig}
\usepackage{textcomp}
\usepackage{stfloats}
\usepackage{verbatim}
\usepackage{balance}
\usepackage{hyperref}
\hypersetup{
    colorlinks=true,        % 使用颜色而非边框
    linkcolor=blue,         % 内部引用颜色
    citecolor=green,        % 引用文献颜色
    urlcolor=cyan,          % URL 颜色
    pdfborder={0 0 0},      % 移除边框宽度（三个0表示无边框）
}

\usepackage[T1]{fontenc}
\usepackage{algorithmic}

\newcommand{\myformer}{Pets }

\newcommand{\Rmnum}[1]{\expandafter\@slowromancap\romannumeral #1@}

\newcommand{\red}[1]{{\textcolor{red}{#1}}}

\def\BibTeX{{\rm B\kern-.05em{\sc i\kern-.025em b}\kern-.08em
    T\kern-.1667em\lower.7ex\hbox{E}\kern-.125emX}}

\definecolor{green_bg}{RGB}{180, 215, 183}
\definecolor{green_hyp}{RGB}{160, 196, 143}

\begin{document}

\title{Energy-Aware Pattern Disentanglement: \\ A Generalizable Pattern Assisted Architecture for Multi-task Time Series Analysis}

\author{
Xiangkai Ma, Xiaobin Hong, Wenzhong Li, Sanglu Lu
\vspace{-20pt}
\thanks{
Xiangkai Ma, Xiaobin Hong, Wenzhong Li, Sanglu Lu are with the State Key Laboratory for Novel Software Technology, Nanjing University, Nanjing 210023, China.
(e-mail: \{xiangkai.ma, xiaobinhong\}@smail.nju.edu.cn, \{lwz, sanglu\}@nju.edu.cn)
}
\thanks{Corresponding Authour: Wenzhong Li (lwz@nju.edu.cn).}
}

\markboth{IEEE TRANSACTIONS ON KNOWLEDGE AND DATA ENGINEERING}
{}

%%%%%%%%%%%%%%%%%%%%%%%%%%%%%%%%%%%%%%%%%%%%%%%%%%%%%%%%%%%%%%%%%%%%%%%%%%%%%%%%%
\maketitle
\begin{abstract}
Time series analysis has found widespread applications in areas such as weather forecasting, anomaly detection, and healthcare. 
While deep learning approaches have achieved significant success in this field, existing methods often adopt a ``one-model one-task'' architecture, limiting their generalization across different tasks.
To address these limitations, we perform local energy analysis in the time–frequency domain to more precisely capture and disentangle transient and non-stationary oscillatory components. Furthermore, our representational analysis reveals that generative tasks tend to capture long-period patterns from low-frequency components, whereas discriminative tasks focus on high-frequency abrupt signals, which constitutes our core contribution.
Concretely, we propose Pets, a novel ``one-model many-tasks'' architecture based on the General fluctuation Pattern Assisted (GPA) framework that is adaptable to versatile model structures for time series analysis. 
Pets integrates a Fluctuation Pattern Assisted (FPA) module and a Context-Guided Mixture of Predictors (MoP). The FPA module facilitates information fusion among diverse fluctuation patterns by capturing their dependencies and progressively modeling these patterns as latent representations at each layer. Meanwhile, the MoP module leverages these generalizable pattern representations to guide and regulate the reconstruction of distinct fluctuations hierarchically by energy proportion. 
Pets demonstrates strong versatility and achieves state-of-the-art performance across 60 benchmarks on various tasks, including forecasting, imputation, anomaly detection, and classification, while demonstrating strong generalization and robustness.
% The demonstrations and source code are publicly available \url{https://github.com/Xiang-Kai/Pets}.
\end{abstract}
\begin{IEEEkeywords}
Temporal-Frequency Analysis, Plug-and-Play Architecture, Fluctuation Pattern Decoupling
\end{IEEEkeywords}

\section{Introduction} \label{section:introduction}
%%1. background of ts
%%Bengio2015ScheduledSF,Chen2001FreewayPM,
Time series analysis plays a pivotal role across numerous domains~\cite{Fan2023DishTSAG,Sezer2019FinancialTS}, such as traffic planning~\cite{Thissen2003UsingSV,Yin2021DeepLO}, encompassing healthcare diagnostics~\cite{Harutyunyan2017MultitaskLA,Tashiro2021CSDICS}, and missing data imputation~\cite{LopezAlcaraz2022DiffusionbasedTS,Fortuin2019GPVAEDP}.
Extensive applications necessitate the models to capture clear, precise, and universal compound fluctuation patterns from the input series and utilize them in various domains and task scenarios~\cite{Liu2023TimesURLSC,Woo2024UnifiedTO}.

%%4. Fig.1. the frequency mapping of diverse task
\begin{figure*}[t]
\begin{center}
\centerline{\includegraphics[width=2.0\columnwidth]{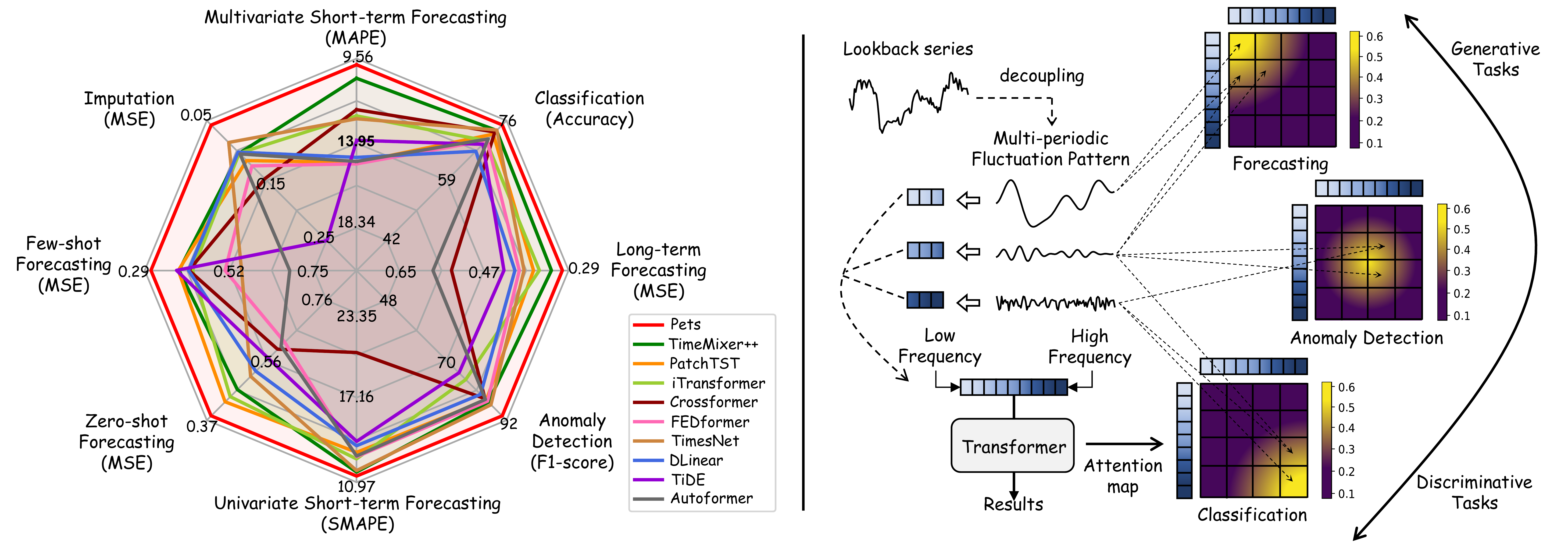}}
\vspace{-10pt}
\caption{
(\textbf{a}) \myformer demonstrates state-of-the-art performances on all 8 tasks. 
(\textbf{b}) We visualize the attention score for diverse tasks. Specifically, The original sequence is decoupled into diverse fluctuation sequences, each undergoing the patch operation to yield a set of tokens. These three token sets are then combined, and attention score between any two tokens are calculated. 
}\label{fig:intro_1}
\end{center}
\vspace{-15pt}
\end{figure*}

%%2. existing deep model in ts, challenge of tokenization in transformer
Recent years, the research endeavor of applying deep models to time series diverse tasks has achieved remarkable success~\cite{IsmailFawaz2018DeepLF,Zhao2020MultivariateTA,zhou2021informer}.
Architectures meticulously designed with fundamental backbones are regarded as the principal focus in time series investigations~\cite{Aseeri2023EffectiveRF,liu2023itransformer,Luo2024ModernTCNAM,Nie2023PatchTST,wu2023timesnet,Zeng2022DLinear}. 
% Initial methods employed RNN~\cite{Shi2015ConvolutionalLN,Aseeri2023EffectiveRF} and TCN~\cite{wu2023timesnet,wang2023micn,Luo2024ModernTCNAM} to capture the temporal dependency. Nevertheless, due to the limitations of Markov chains, RNNs tend to forget long-term dependency information. Meanwhile, TCNs have difficulty capturing global information owing to the constraints of the receptive field. 
% Recently, models based on the transformer architecture~\cite{Nie2023PatchTST,liu2023itransformer} have utilized the self-attention mechanism to establish global dependency among the observed sequences. Meanwhile, the lightweight MLP-based models~\cite{Zeng2022DLinear,Xu2023FITSMT} have demonstrated outstanding potential in both the prediction preciseness and the computational complexity.
Besides, enhancement strategies based on frequency assistance~\cite{yi2023FreTS,Yi2024FilterNetHF,zhou2022fedformer} and multi-period decoupling~\cite{Liu2021SCINetTS,liu2021pyraformer,Wang2025TimeMixerAG,Wang2024TimeMixerDM} have been integrated into time series analysis.
% , with the focus on disentangling trends, seasonal or disparate scale components from the observed sequences. 
The effectiveness of these strategies roots the fact that the isolated time point manifests a superimposed state of multiple periodic-fluctuation patterns, which constitutes an inherent challenge~\cite{Wang2024TimeMixerDM}.
% Nevertheless, in contrast to the tokens in language tasks that are strictly affiliated with distinct contexts~\cite{Jin2023TimeLLMTS}, the isolated time point manifests a superimposed state of multiple periodic-fluctuation patterns.
Furthermore, prior approaches follow the ``one-model one-task'' architecture, where dedicated models were designed for specific tasks. For example, time series prediction models cannot be directly applied to classification task~\cite{dong2023simmtm}, and vice versa. 
To address these challenges, this paper proposes a novel \textbf{``one-model many-tasks''} architecture by generalizing task-universal representations from the superposition state observation.
% For instance, there exist disparate fluctuation modalities across three cyclical magnitudes: daily, monthly, and yearly. These challenges present significant obstacles for time series to comprehensively harness the inductive biases of the attention mechanism and convolution~\cite{Wang2025TimeMixerAG,Wang2024TimeMixerDM}.

%%3. We need to establish a PETs
%%% shadow={2.5pt}{-2.5pt}{0pt}{opacity=5,black}, 
% \vspace{-1mm}
% \begin{tcolorbox}[notitle, rounded corners, colframe=gray, colback=white, boxrule=2pt, boxsep=0pt, left=0.15cm, right=0.17cm, enhanced, toprule=2pt, before skip=0.65em, after skip=0.75em]
% \emph{{\centering 
%   {\fontsize{8.5pt}{13.2pt}\selectfont 
%   These challenges encourage us to establish a General fluctuation Pattern to Assist temporal architecture (GPA).
%   }\\
% }}
% \end{tcolorbox}
% \vspace{-1mm}

%%5. the challenge of establishment about PETs
Our core idea is employing the General fluctuation Pattern Assisted architecture (GPA) to improve the model's capacity to discern and generalize universal patterns from observed sequences in a superimposed state. Decoupling observations of mixed multi-periodic series into a combination of multiple fluctuating modes will facilitate the adaptive selection of beneficial decomposition patterns for diverse downstream tasks. 
% Various time series analysis tasks would benefit from explicitly modeling diverse inherent fluctuation patterns. Nevertheless, disparate downstream tasks customarily capture the apprehension from particular fluctuation patterns~\cite{Gao2024UniTSAU,Wang2025TimeMixerAG}.  
To illustrate the sensitivity of tasks to different fluctuation patterns, Figure~\ref{fig:intro_1} (b) visualizes the attention scores both between and within decoupled fluctuation patterns across various tasks. Notably, the model tends to capture diverse pattern dependencies tailored to specific downstream tasks. For forecasting tasks, the model prefers learning dependencies among low-frequency fluctuation tokens. In contrast, for classification tasks, the model leans towards capturing abrupt signals in high-frequency patterns to better serve discriminative purposes.

% These impediments exemplify the challenges associated with constructing a flexible fluctuation pattern enhancement architecture that is universally applicable across a multiplicity of tasks and time series data.

%%7. how to establish a PETs
We proposes \textbf{P}attern \textbf{E}nhanced \textbf{T}ime \textbf{S}eries architecture (Pets) as a successful implementation of GPA, with the intention of learning the innate hybrid fluctuation enhancement assemblages within temporal data, by adaptively disentangling the dynamic superposition states of multiple period-fluctuation patterns. Its core concept is to use a fixed spectrum distribution to guide the decoupling procedure, thereby establishing a residual-guided hybrid output paradigm by guaranteeing that the preponderant portion of frequency information and fluctuation energy is concentrated in particular patterns.

% the 1D observational sequence is disentangled and projected into the 2D time-frequency spectrum space. Thereafter, 
Concretely, an Amplitude Margin Interval filter (AMI) is contrived to effectuate an adaptive partitioning of the spectral intervals, ensuring that the sequence projections within each interval exhibit a fixed spectrum distribution. This approach is designated as the Spectrum Decomposition and Amplitude Quantization strategy (SDAQ). Based on obtaining decoupled multiple fluctuation patterns, we propose the Fluctuation Pattern Assisted (FPA) approach consisting of three components: (1) Periodic Prompt Adapter (PPA), (2) Multi-fluctuation Patterns Rendering (MPR), and (3) Multi-fluctuation Patterns Mixing (MPM) for the manipulation of the mixed periodic fluctuation modalities.
Motivated by the periodic pattern interference mechanism~\cite{Lin2024CycleNetET}, the PPA models the interaction factors among diverse fluctuations to uncover comprehensive temporal patterns. The MPR hierarchically aggregates these patterns and facilitates the backbone blocks in apprehending the hidden representations specific to certain patterns. The MPM promotes the renewal of compound fluctuation patterns and the capture of more profound hidden representations. 
In the generative scenario, we propose a context-guided mixture of predictors (MoP), which arranges the hidden representations of fluctuations in descending order of energy proportion. 
% Specifically, within the prediction task, the shallow predictor engenders the long-wave patterns of the future sequence, which subsequently functions as a conditional variable to guide the deep predictor in generating the short-wave patterns, thereby progressively constructing the predicted sequence. 
Contributions can be summarized as:
% \vspace{-5pt}
\begin{itemize}
    \item We propose Pets, a novel ``one-for-many'' model architecture that decouples and extracts generalizable fluctuation patterns in time series from an energy-guided temporal-spectral comprehensive perspective for the first time. This design serves as a versatile backbone for a wide range of downstream time series tasks.    
    % Sequence often exhibit a superimposed state of various fluctuation patterns, which makes time series analysis challenging. Surpassing the existing multi-period decoupling paradigms, we introduces a novel perspective based on energy distribution within the temporal-spectrum space and establishes a General Fluctuation \textbf{P}attern \textbf{E}hanced \textbf{T}emporal Architecture (\textbf{Pets}).
    % Surpassing prior approaches, we propose Pets, a novel ``one-for-many'' architecture that disentangles and extracts the generalizable fluctuation pattern in time series from the energy-guided temporal-spectral comprehensive perspective, enabling robust and interpretable time series decomposition for multiple tasks.
    % \vspace{-2pt}
    \item Pets incorporates three key components: PPA, MPR and MPM, to model the dependencies among patterns and aggregate them effectively. These components enable the hierarchical learning of task-relevant features, as well as the refinement of mixed patterns.
    % Through amplitude margin filtering, SDAQ adaptively segregates the sequence into fluctuation patterns with a stabilized energy distribution. Addiationally, PPA captures the dependencies among various patterns, while MPR and MoP execute adaptive aggregations on the mixed patterns according to the energy ranking order, enabling the patterns highly relevant to the task to dominate the output results.
    % Pets incorporates three key components: PPA, MPR and MPM, to model the dependencies among patterns and aggregate them effectively. These components enable the hierarchical learning of task-relevant features, and the refinement of mixed patterns, addressing the complexity of multi-pattern.
    % \vspace{-2pt}
    \item Pets achieves state-of-the-art performances in all 8 mainstream time series analysis tasks across 60 benchmarks, as shown in Fig.~\ref{fig:intro_1} (a). Furthermore, Pets can consistently enhance the performance of disparate model architectures on extensive datasets and tasks.
\end{itemize}

\section{Related Work} \label{section:relatedwork}
The core of time series investigation lies in meticulously designed architectures. The initial models employed RNN~\cite{Aseeri2023EffectiveRF,Shi2015ConvolutionalLN} and TCN~\cite{Luo2024ModernTCNAM,wang2023micn,wu2023timesnet} to capture the temporal dependency. Recently, models based on the Transformer~\cite{liu2023itransformer,Nie2023PatchTST,wu2021autoformer,zhou2021informer} have utilized the self-attention mechanism to establish global dependency among the observed sequences. Meanwhile, the lightweight MLP-based models~\cite{Campos2023LightTSLT,Xu2023FITSMT,Zeng2022DLinear} have demonstrated outstanding potential in both the performance and complexity. 
Moreover, the methods most closely related to our \myformer can be divided into two aspects.

\subsection{Multi-periodic or Multi-frequency Decomposition}
Beyond the research on modeling architectures, the enhancements based on frequency analysis and multi-periodic decomposition have also demonstrated remarkable potential~\cite{Finder2022WaveletFM,Yi2024FilterNetHF,zhou2022fedformer,Wang2024FCVAE}. 
Specifically, TimeMixer~\cite{Wang2024TimeMixerDM} and TimeMixer++~\cite{Wang2025TimeMixerAG} generate multi-scale sequences via downsampling only in the time domain and establishes connections between different-scale representations using linear layers.
N-BEATS~\cite{Oreshkin2019nbeats} and N-HITS~\cite{Challu2022NHiTSNH} impose constraints on expansion coefficients as prior information to guide layers to learn specific time-series features (e.g., seasonality), enabling interpretable sequence decomposition. 
ROSE~\cite{Wang2024ROSERA} separates the coupled semantic information in time series data through multiple frequency domain masks and reconstructions, subsequently extracting unified representations across diverse domains.
FreTS~\cite{yi2023FreTS} and FITS~\cite{Xu2023FITSMT} advocate the utilization of MLP to capture the complete perspective and global dependency of the observed sequence within the frequency domain.
% Rose~\cite{Wang2024ROSERA} and FreTS~\cite{yi2023FreTS} are designed as a task-specific enhancer for forecasting, improving the performance of a given backbone by introducing frequency decomposition. 

However, the multi-scale design in TimeMixer++ stems from globally fixed downsampling ratios, similarly, ROSE calculates a global frequency composition, which may be suboptimal for non-stationary time series.
SDAQ adaptively partitions the spectrum into sub-bands based on local energy distribution. Its core advantages, locality and energy guidance, enable SDAQ to capture the temporal evolution of frequency components. 
Furthermore, the energy-guided partitioning ensures that, within any local time window, dominant energy components are accurately isolated, enabling finer-grained capture and separation of transient and non-stationary fluctuation components.
This intrinsic property grants Pets a natural advantage when dealing with non-stationary data.

\subsection{Multi-task Models}
Recently, some researchers have begun exploring multi-task models, i.e., using a single model to perform multiple tasks. Representative works include TimesNet~\cite{wu2023timesnet}, OneFitsAll~\cite{Zhou2023OneFA}, ModernTCN~\cite{luo2024moderntcn} and TimeMixer++~\cite{Wang2025TimeMixerAG}. 
These methods typically employ a shared backbone architecture to extract temporal features and design task-specific heads for downstream tasks.

Nevertheless, existing works have not explicitly addressed a foundational question: \textbf{``How do fluctuation patterns of different frequencies within time series serve distinct analytical tasks?''}
Benefiting from the explicit decoupling of distinct patterns via SDAQ, we can explicitly investigate the sensitivity of various downstream tasks to different periodic–frequency patterns. 
Extensive representational visualization analyses in Figures~\ref{fig:representation_intra} and \ref{fig:representation_inter} demonstrate that low-frequency, long-period fluctuation patterns are crucial for generative tasks (forecasting and imputation), whereas decisions in discriminative tasks primarily depend on the presence of high-frequency perturbations and abrupt signals within the observation window, which is the key contribution of our work.

%%6. Fig.2. the overall architecture
\begin{figure*}[t]
\begin{center}
\centerline{\includegraphics[width=2.0\columnwidth]{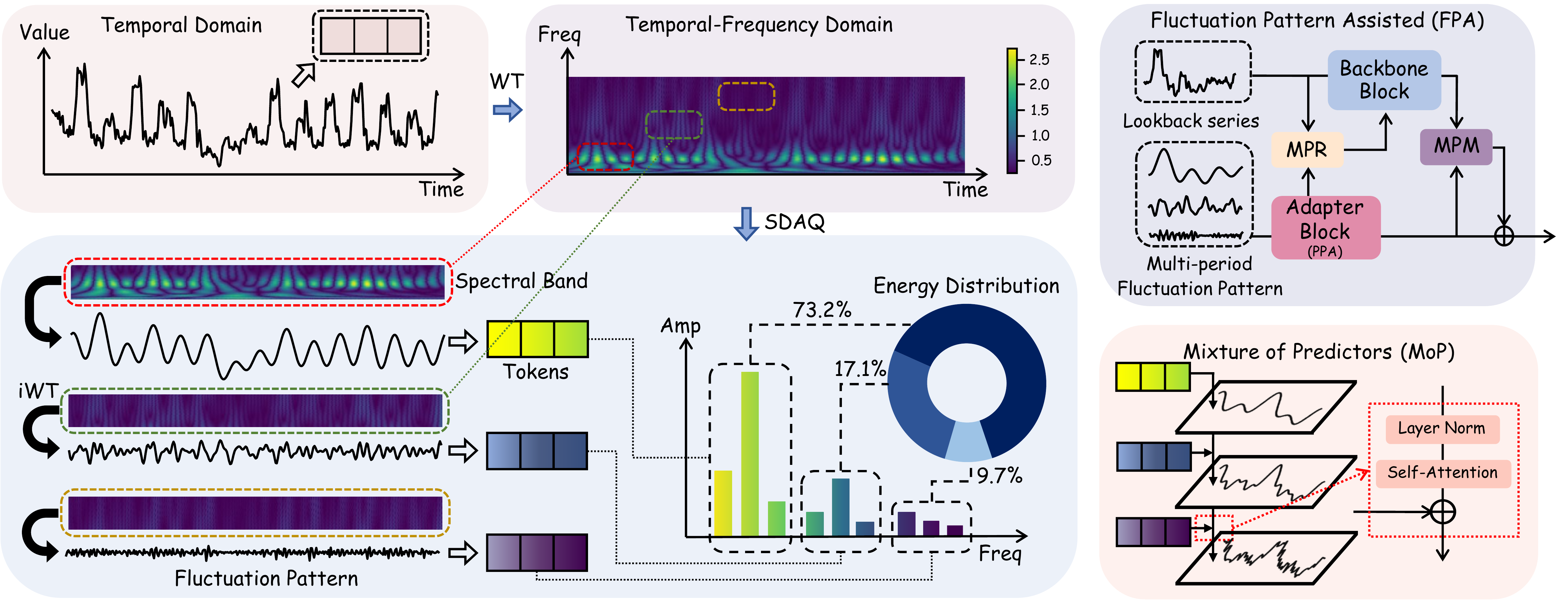}}
\vspace{-10pt}
\caption{
The overall architecture of Pets, which is based on the innovative temporal-spectral decomposition and amplitude quantization paradigm. Pets consists of features Fluctuation Pattern Assisted module (FPA) and context-guided Mixture of Predictors (MoP), which respectively capture and aggregate the universal fluctuation patterns to enhance diverse tasks.
}\label{fig:intro_2}
\end{center}
\vspace{-15pt}
\end{figure*}

\section{\myformer Architecture} \label{section:pets}

In the long-term forecasting task, \begin{small}$X_{-L+1:0}\in\mathbb{R}^{B\times d\times L}$\end{small} and \begin{small}$X_{1:H}\in\mathbb{R}^{B\times d\times H}$\end{small} are utilized to represent the observed series and the future series, respectively, where $B$ denotes the batch size, $d$ denotes the number of channels of the multivariate time series, $L$ and $H$ denote the lookback window and forecast horizon. Based on the channel independent design, the input observation sequence \begin{small}$X_{-L+1:0}\in\mathbb{R}^{B\times d\times L}$\end{small} is first processed as \begin{small}$X_{-L+1:0}\in\mathbb{R}^{Bd\times L}$\end{small}.

\vspace{-5pt}
\subsection{Decomposition and Quantization Paradigm}
\subsubsection{Temporal-Spectral Decomposition.}
Initially, through the application of the continuous wavelet transform (CWT)~\cite{Lee2023ForecastingWI}, the composite multi-periodic patterns inherent in the observed sequence \begin{small}$X_{-L+1:0}\in\mathbb{R}^{Bd\times L}$\end{small} are transmuted from the temporal space into biaxial space, yielding the temporal-frequency spectrum image \begin{small}$A\in\mathbb{R}^{Bd\times L\times\lambda}$\end{small}, wherein $\lambda$ designates the spectral width within the time-frequency spectrum space. Specifically, we calculate the wave amplitude $A_{i,j}$ of the original sequence at time $t_i$ and frequency $f_j$, where \begin{small}$i\in[1,L],j\in[1,\lambda]$\end{small}. This wave amplitude $A_{i,j}$ represents the magnitude of the observed sequence projected onto frequency band $f_j$ in the vicinity of time $t_i$, which is construed as the energy density of the observed sequence within the relevant fluctuation pattern, as shown in Fig.~\ref{fig:intro_2}.
Nevertheless, the spectral density distributions of time series from diverse domains vary significantly, which causes the direct modeling on the image to limit the cross-domain generalization capacity of the model, and leads to substantial computational complexity. To address this issue, we propose the temporal-spectral decomposition and amplitude quantization strategy.

\subsubsection{Amplitude Quantization Phase.}
In the realm of prediction, the preponderant portion of high-energy density regions within the temporal-frequency spectrum is concentrated within the long-wave and low-frequency bands. This has inspired us to instigate the conception of the amplitude-margin interval (\textit{AMI}). 
This paper aims to identify a relatively diminutive frequency domain interval (i.e., a frequency band) so that the preponderant amount of energy within the time-frequency domain space is involved within this band. The mathematical formulation of the \textit{AMI} function is:
% \begin{small}
\begin{equation}
  \begin{split}
    \sum_{j\in[1,\textit{AMI}(\mu)]}\sum_{i\in[1,L]}A_{i,j}=\mu\cdot\sum_{i}\sum_{j}A_{i,j},\mu\in(0,1),
  \end{split}
\end{equation}
% \end{small}
where \begin{small}$A_{:,j}$\end{small} represents cumulative energy density of the spectrum \begin{small}$f_j$\end{small} over all time points. 
The function $\textit{AMI}(\mu)$ demarcates spectral interval \begin{small}$\{f_j\}_{1\leq j<\textit{AMI}(\mu),j\in\mathbb{Z}}$\end{small} such that a specific proportion of energy is located within this band. 
A set of fixed proportions \begin{small}$\{\mu_1,\mu_2,..,\mu_{K\textit{-}1}\}$\end{small} can adaptively determine the partitioning of spectral sub-bands:
% \begin{small}
\begin{equation}
  \begin{split}
    sub_1=\{f_j\}_{1\leq j<\textit{AMI}(\mu_1)}, \\
    sub_2=\{f_j\}_{\textit{AMI}(\mu_1)\leq j<\textit{AMI}(\mu_2)}, \\
    sub_K\textit{=}\{f_{j}\}_{\textit{AMI}(\mu_{K\textit{-}1})\leq j<\lambda}.
  \end{split}
\end{equation}
% \end{small}
Subsequently, \begin{small}$A\in\mathbb{R}^{Bd\times L\times\lambda}$\end{small} is deconstructed into \begin{small}$K$\end{small} time-frequency sub-images \begin{small}$A^{1,2,..,K}\in\mathbb{R}^{Bd\times L\times\lambda}$\end{small}. Here, \begin{small}$A^{k}$\end{small} is generated from \begin{small}$A$\end{small} by mask operation:
% \begin{small}
\begin{equation}
  \begin{split}
    {A_{i,j}^{k}=A_{i,j}~if~f_j\in sub_k~else~0},~~k=1,2,..,K.
  \end{split}
\end{equation}
% \end{small}
Ultimately, the inverse wavelet transform (iWT) is employed to project the three time-frequency spectrum images back into the time domain space, thereby realizing the spectrum decomposition and amplitude quantization process of the composite fluctuation pattern. And the decoupled sequences are defined:
% \begin{small}
\begin{equation}
  \begin{split}
    X_{-L+1:0}^{k}=iWT(A^{k})\in\mathbb{R}^{Bd\times L},~~k=1,2,..,K.
  \end{split}
\end{equation}
% \end{small}
The auxiliary quantization strategy adaptively segregates the complex and variable continuous spectrum into multiple spectrum intervals. We guarantee the scalability and generalization of the modeling by fixing the energy density distribution of the spectrum intervals. 
Moreover, each spectrum interval is inversely converted into a time-domain sequence with specific periodic characteristics, which significantly reduces the complexity and memory overhead.
In our experiments, we keep the \begin{small}$(K\textit{=}3,\lambda\textit{=}50,\mu_1\textit{=}0.7,\mu_2\textit{=}0.9)$\end{small} setting, and we discuss the diverse hyperparameters 
in the Section IX-C.
% in the Section~\ref{sec:hyperparameter}.

\vspace{-5pt}
\subsection{Patch Embedding}
Following the patching instance strategy~\cite{Nie2023PatchTST}, each sequence of length \begin{small}$L$\end{small} is rerepresented as \begin{small}$P_{L}\times p$\end{small}, where $p$ as the patch length, \begin{small}$P_{L}$\end{small} is the number of tokens, and satisfies \begin{small}$P_L\times p=L$\end{small}. 
Based on this, the observation sequence \begin{small}$X_{-L+1:0}\in\mathbb{R}^{Bd\times L}$\end{small} and decoupled multi-periodic sequences \begin{small}$\{X_{-L+1:0}^{k}\in\mathbb{R}^{Bd\times L}\}_{k\in [1,K]}$\end{small} are converted as embedding \begin{small}$X_{-L+1:0}\in\mathbb{R}^{Bd\times P_L\times p}$\end{small} and \begin{small}$\{X_{-L+1:0}^{k}\in\mathbb{R}^{Bd\times P_{L}\times p}\}_{k\in [1,K]}$\end{small}, where $P_{d}$ represents the dimension of tokens. Subsequently, observation tokens \begin{small}$E^0\in\mathbb{R}^{Bd\times P_L\times P_d}$\end{small} and multi-periodic tokens \begin{small}$\{E_k^0\in\mathbb{R}^{Bd\times P_L\times P_d}\}_{k\in [1,K]}$\end{small} are calculated by the conv1d embedding layer.

\begin{figure*}[t]
\begin{center}
\centerline{\includegraphics[width=2.0\columnwidth]{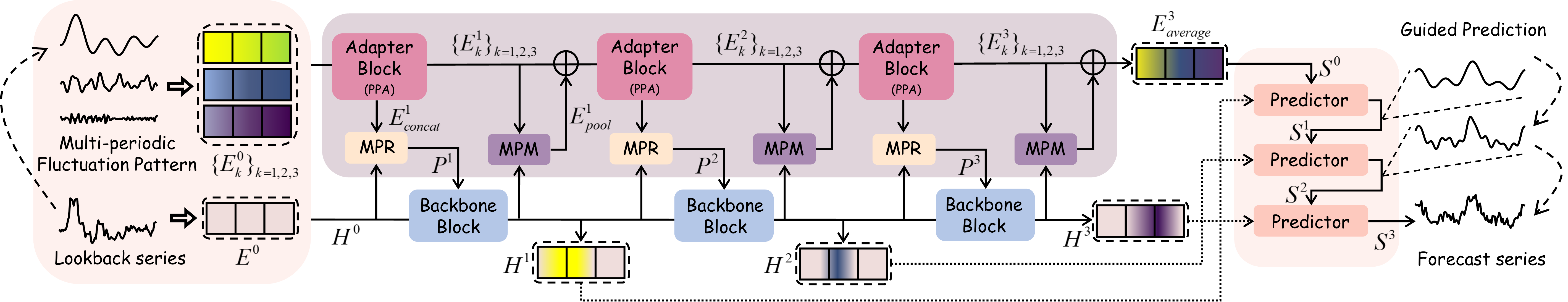}}
\vspace{-5pt}
\caption{
The overall architecture of Pets. Pets is a two-branch architecture consisting of three components: (a) a plain Transformer or MLP (Backbone Block), which is evenly divided into diverse stages for feature interaction. (b) FPA employs the proposed PPA (Adapter Block), MPR and MPM, to capture the dependencies between different fluctuation patterns, and (c) MoP adaptively aggregates the mixed fluctuation patterns according to the energy ranking order, and the fluctuation patterns with high task relevance can dominate the output. See Fig.~\ref{fig:method_2} for detailed designs of PPA, MPR and MPM.
}\label{fig:method_1}
\end{center}
\vspace{-15pt}
\end{figure*}

\subsection{Periodic Pattern Interference Mechanism}
The overall architecture of Pets is illustrated in Fig.~\ref{fig:intro_2} and comprises three components: (a) the common Transformer or MLP backbone. (b) the stacked FPA, which assimilates the proposed PPA, MPR, and MPM. (c) Mixture of Predictors (MoP). 
During the forward pass, the PPA and the backbone block respectively receive the composite fluctuation pattern embedding and the observed sequence embedding as inputs respectively. The PPA is employed to fortify the fluctuation pattern information. Subsequently, the MPR and MPM are devised to interact and fuse the bidirectional features between the two principal line. Finally, the MoP adaptively aggregates the fluctuation patterns, permitting the task highly relevant fluctuation patterns to dominate the model's output. Moreover, architecture details are entirety presented in the pseudocode~\ref{alg:training_part}.
The proposed components can be facilely integrated into any advanced models (such as DLinear~\cite{Zeng2022DLinear}, PatchTST~\cite{Nie2023PatchTST} and TimeMixer~\cite{Wang2024TimeMixerDM}) and consistently demonstrate state-of-the-art performance across diverse downstream tasks. 

\subsubsection{Periodic Prompt Adapter Blocks.}
PPA introduces composite fluctuation pattern information through bidirectional interaction to enhance the generalization capacity and semantic representation of the deep model while remaining adaptable to any fundamental architecture. Specifically, the adapter block at the $n$-th layer (\begin{scriptsize}$n\in[1,N]$\end{scriptsize}) receives the representations of diverse patterns \begin{small}$\{E_{k}^{n-1}\in\mathbb{R}^{B\cdot d\times P_{L}\times P_{d}}\}_{k\in [1,K]}$\end{small} as inputs, and fully exploits the inductive bias of local convolution modeling to distill the deep characterizations under each periodic pattern from the input multi-periodic pattern embedding tokens:
% \begin{small}
\begin{equation}
  \begin{split}
    E_k^{n-1}\textit{=}Conv1D(Linear(E_k^{n-1})^\top)^\top, \\
    E_k^{n-1}\textit{=}Dropout(Linear(Act(E_k^{n-1}))).
  \end{split}
\end{equation}
% \end{small}
Subsequently, a self-attention layer is meticulously contrived with the intention of capturing the dependencies among disparate representations of various periodic fluctuation patterns, thereby empowering the model with an enhanced capacity to capture the composite temporal information. We commence by splicing \begin{small}$\{E_k^{n-1}\in\mathbb{R}^{B\cdot d\times P_L\times P_d}\}_{k\in [1,K]}$\end{small} along the dimension of the number of tokens to obtain \begin{small}$E_{concat}^{n-1}\in\mathbb{R}^{B\cdot d\times KP_{L}\times P_{d}}$\end{small}, which as the concatenated mixed fluctuation pattern embedding. 
% The self-attention mechanism is formulated as, \begin{small}$E_{concat}^{n-1}\textit{+=}SelfAttn(E_{concat}^{n-1})$\end{small}, and then \begin{small}$E_{concat}^{n-1}\textit{+=}Conv1D(E_{concat}^{n-1})$\end{small}.
Finally, the mixed pattern representation independently executes three one-dimensional convolutions to decouple into embeddings \begin{small}$\{E_k^n\}_{k\in [1,K]}$\end{small} corresponding to multiple different fluctuation patterns, as:
% \begin{small}
\begin{equation}
  \begin{split}
    E_{k}^{n}\textit{=}Conv1D_{k}(E_{concat}^{n-1})_{k\in [1,K]}.
  \end{split}
\end{equation}
% \end{small}

\subsubsection{Multi-fluctuation Patterns Rendering.}
MPR augments the fluctuation patterns captured by the adapter block as conditional context to guide the backbone block to focus on modeling that particular fluctuation pattern. 
In the the prediction, the shallow backbone block accentuate the extraction of the low-frequency and long-wave pattern from the multi-period fluctuation patterns, which encompasses the trend and long-period fluctuation particular within the observed sequence (the yellow-tinted tokens in Fig.~\ref{fig:intro_2}). These proclivities will be reflected within the hidden representations outputted by each layer of the backbone block. 
Concretely, the MPR module takes in two input tensors: the decoupled fluctuation pattern group \begin{small}$\{E_{k}^{n}\in\mathbb{R}^{B\cdot d\times P_{L}\times P_{d}}\}_{k\in [1,K]}$\end{small} output by the current layer's adapter block and the hidden representation \begin{small}$H^{n-1}\in\mathbb{R}^{B\cdot d\times P_L\times P_d}$\end{small} generated by the previous layer's backbone block (for the first MPR module, it is the embedding \begin{small}$E^0\in\mathbb{R}^{B\cdot d\times P_L\times P_d}$\end{small} of the observed sequence). 
During the forward progression, the distinctive zero-convolution design ensures that MPRs are inclined to learn inductive biases from diverse fluctuation patterns. Specifically, within the shallow MPR block, the high-frequency pattern \begin{small}$\{E_k^n\}_{k\in [2,K]}$\end{small} undergoes a \begin{small}$1\times 1$\end{small} convolution with both weights and biases initialized to zero, and then is concatenated with the low-frequency pattern \begin{small}$\{E_k^n\}_{k=1}$\end{small} into the mixed fluctuation pattern tokens \begin{small}$E_{concat}^n$\end{small}. 
Subsequently, similarly through the self-attention mechanism, and finally through the pooling layer, it is combined with the hidden representation \begin{small}$H^{n-1}$\end{small} via element-wise addition as the prompt embedding \begin{small}$P^n\in\mathbb{R}^{B\cdot d\times P_L\times P_d}$\end{small} of the current MPR layer.
The mathematical formulation is expressed as: 
% \begin{small}
\begin{equation}
  \begin{split}
    E_{concat}^n=E_{concat}^n+SelfAttn(E_{concat}^n), \\
    E_{concat}^n=E_{concat}^n+Conv1D(E_{concat}^n), \\
    P^n=H^{n-1}+E_{concat}^nPool(E_{concat}^n).
  \end{split}
\end{equation}
% \end{small}

\begin{figure*}[t]
\begin{center}
\centerline{\includegraphics[width=2.0\columnwidth]{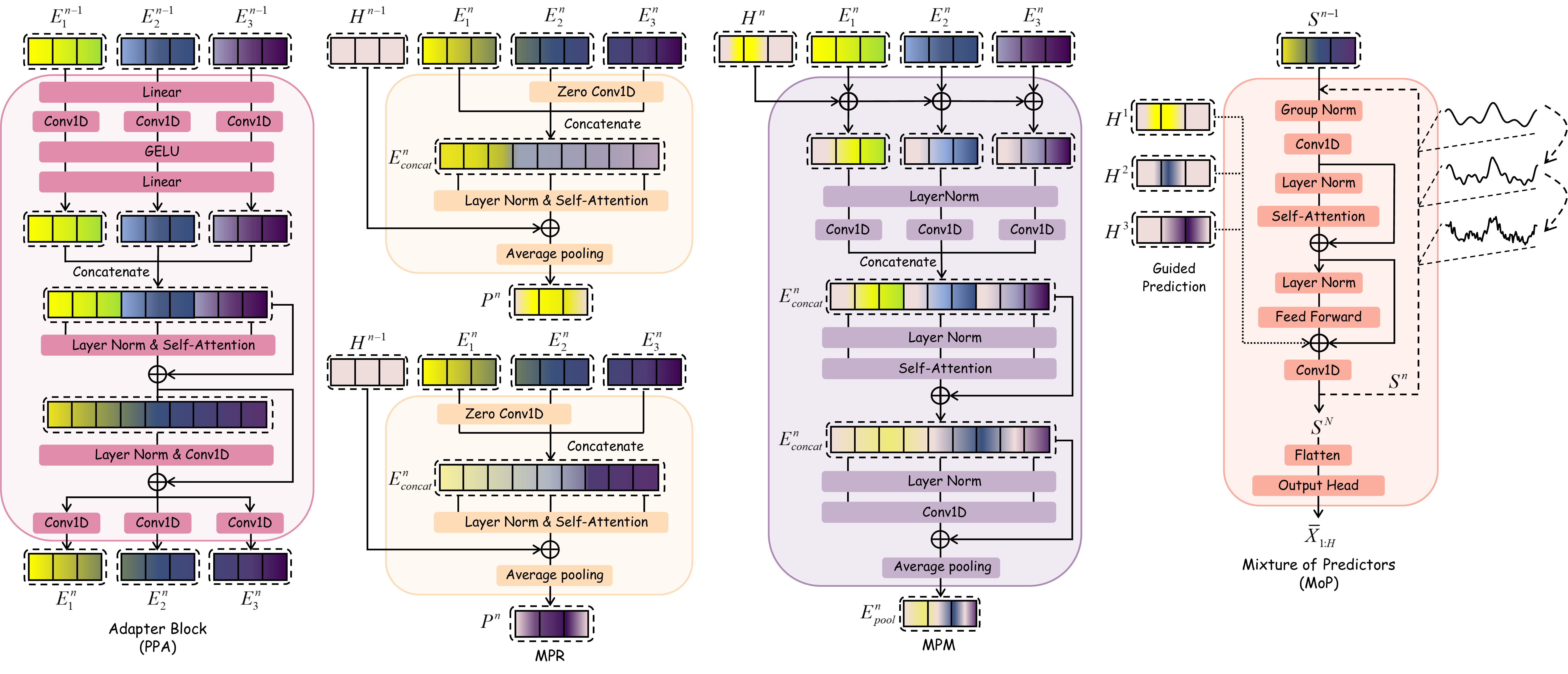}}
\vspace{-10pt}
\caption{
The detailed structure of the proposed model encompasses components: PPA, MPR, MPM and MoP.
}\label{fig:method_2}
\end{center}
\vspace{-20pt}
\end{figure*}

\subsubsection{Backbone Blocks.}
In alignment with the design of popular baselines, to ascertain that the proposed Pets serves as a universal and plug-and-play enhanced architecture. When PatchTST is selected as the Backbone, the hidden representation \begin{small}$H^{n-1}\in\mathbb{R}^{B\cdot d\times P_L\times P_d}$\end{small} generates by the previous layer's backbone block (for the first backbone block, it is the embedding \begin{small}$E^0\in\mathbb{R}^{B\cdot d\times P_L\times P_d}$\end{small} of the observed sequence) first passes through the transformer layer, and subsequently performs the element-wise addition operation with \begin{small}$P^n\in\mathbb{R}^{B\cdot d\times P_L\times P_d}$\end{small} calculated by the current layer's MPR, as follows:
% \begin{small}
\begin{equation}
  \begin{split}
    H^{n-1}=H^{n-1}+SelfAttn(H^{n-1}), \\
    H^n=P^n+H^{n-1}+FFN(H^{n-1}).
  \end{split}
\end{equation}
% \end{small}

\subsubsection{Multi-fluctuation Patterns Mixing.}
% The MPM module is meticulously contrived to ensure that the modeling results of the shallow backbone block regarding the prediction sequence are present within the hidden representations and are accessible to the deeper adapter block. 
The MPM module is purposed to establish connections among fluctuation patterns of diverse periods, precluding the model from conducting independent modeling on each periodic scale and, as a result, significantly enhancing the generalization and performance of Pets.

MPM accepts two segments of inputs, namely a set of fluctuation pattern \begin{small}$\{E_k^n\in\mathbb{R}^{B\cdot d\times P_L\times P_d}\}_{k\in [1,K]}$\end{small} generated by the current layer's adapter block and the hidden representation \begin{small}$H^n\in\mathbb{R}^{B\cdot d\times P_L\times P_d}$\end{small} generated by the current layer's backbone block, and obtains the input for the next layer's adapter block after forward progression. The hidden representation and a set of fluctuation patterns. The hidden representation contains the information of fluctuation patterns already learned by the model, which serves as conditional information to guide the model in capturing more profound fluctuation patterns. The MPM module first performs element-wise addition between the hidden representation and each fluctuation pattern representation, subsequently concatenating them together after independently passing through a convolution layer:
% \begin{small}
\begin{equation}
  \begin{split}
    E_k^n=Conv1D(E_k^n+H^n), \\
    E_{concat}^n=Concatenate(E_1^n,E_2^n,....E_K^n).
  \end{split}
\end{equation}
% \end{small}
Eventually, the attention is utilized to capture the dependencies among the mixed tokens of all multiple fluctuation patterns. 
After passing through the pooling layer, the updated pattern representation is obtained by:
% \begin{small}
\begin{equation}
  \begin{split}
    E_{pool}=Pool(E_{concat}^n+Conv1D(E_{concat}^n))),
  \end{split}
\end{equation}
% \end{small}
which is employed to guide the generation of the fluctuation pattern of the next layer, as \begin{small}$E_k^{n}\textit{=}E_k^{n}\textit{+}E_{pool}^{n}$\end{small}, where \begin{small}$k\in [1,K]$\end{small}.
Additionally, we present a novel hybrid predictor for gradually generating the fluctuation details of future sequences.

\subsubsection{Mixture of Predictors.}
% The amplitude quantization strategy guarantees the crucial fluctuation modalities within the observed sequence are all incorporated in the hidden representations captured by the shallow adapter block. Concretely, the long-wave pattern information (encompassing trend and long-period information) captured by the shallow adapter block contain abundant significant information that is sensitive to the prediction sequence. 
% By contrast, the short-wave pattern (high-frequency component) that the deep adapter block pays attention is more susceptible to noise interference. The importance of diverse fluctuation modalities for the prediction results is not entirely identical.

In the forward procedure, the output result \begin{small}$\{E_k^N\in\mathbb{R}^{B\cdot d\times P_L\times P_d}\}_{k\in [1,K]}$\end{small} of the last MPM module is obtained by element-wise addition to yield \begin{small}$E_{average}^N\in\mathbb{R}^{B\cdot d\times P_L\times P_d}$\end{small}. The intermediate result generated by the $n$-th backbone block is designated as the hidden representation group \begin{small}$H^n\in\mathbb{R}^{B\cdot d\times P_L\times P_d}$\end{small}, where \begin{small}$n=1,2,...,N$\end{small}. To fortify the robustness and predictive performance of the model, Pets devises a innovative stepwise prediction strategy and a hybrid predictor architecture. In the MoP consisting of $N$ weight-independent predictors, each predictor receives two inputs. The dotted and solid lines symbolize reference tokens and situational tokens respectively, as depicted at the rightmost side in Fig.~\ref{fig:intro_2}. 
The hidden representation $H^{n}$ serves as the reference tokens for $n$-th predictor, with the first predictor receiving \begin{small}$E_{average}^N$\end{small} as situational tokens. 
The output of the \begin{small}$\textit{n-th}$\end{small} predictor is adopted as the previous tokens \begin{small}$S^{n\textit{+}1}$\end{small} for the \begin{small}$\textit{(n+1)-th}$\end{small} predictor. 
Specifically, in the $n$-th predictor:
% \begin{small}
\begin{equation}
  \begin{split}
    S^{n}=S^{n}+Selfattn(Conv1D(S^{n})),
  \end{split}
\end{equation}
% \end{small}
and the Element-wise Addition is performed between the situational tokens and the reference tokens, as follows:
% \begin{small}
\begin{equation}
  \begin{split}
    S^{n}\text{=}S^{n}\text{+}H^n\text{+}FFN(S^{n}),~~S^{n+1}\text{=}Conv1D(S^{n}).
  \end{split}
\end{equation}
% \end{small}
The reference tokens \begin{small}$S^N\in\mathbb{R}^{B\cdot d\times P_L\times P_d}$\end{small} generated by the final predictor are flattened and then passed through a linear-based projection layer, thus obtaining the prediction result:
% \begin{small}
\begin{equation}
  \begin{split}
    \hat{X}_{1:H}~\textit{=}~OutputHead(Flatten(S^N)).
  \end{split}
\end{equation}
% \end{small}

\subsection{Algorithm for training of \myformer}
We provide the training procedure of \myformer in Algorithm~\ref{alg:training_part}. 
Where $p_\textit{emb}$ and $\left(\overline{h}_m,\overline{h}_c,\overline{h}_b,\overline{h}_a\right)$ together serve as the condition variable $c$, 
$\tilde{Y}_{1:H}^{t-1}\in\mathbb{R}^{Bd\times P_H\times P_d}$ and $\hat{Y}_{1:H}^{t-1}\in\mathbb{R}^{Bd\times P_H\times P_d}$ represent the conditional and unconditional outputs of the denoising network, respectively. Furthermore, $Y_{1:H}^{t-1}$ as the final result, and $\lambda$ is a parameter that controls the proportion of conditional and unconditional generation in the final result ($\lambda \textit{=7.5}$ for our experiment)

\begin{algorithm}[ht]
\caption{Training Procedure of Pets}\label{alg:training_part}
\label{alg:pets}
\begin{algorithmic}[1]
\REQUIRE Observation series $X_{-L+1:0} \in \mathbb{R}^{B \times d \times L}$, future series $X_{1:H} \in \mathbb{R}^{B \times d \times H}$, hyperparameters: model depth $N$, feature dimension $P_d$, spectral width $\lambda$, energy thresholds $(\mu_1, \mu_2)$, iteration count $N_{\text{iter}}$

\FOR{$i = 1$ to $N_{\text{iter}}$}
    \STATE // \textbf{Step 1: SDAQ – Spectral Decomposition \& Amplitude Quantization}
    \STATE Reshape input: $X_{-L+1:0} \in \mathbb{R}^{B \times d \times L} \to \mathbb{R}^{Bd \times L}$
    \STATE Apply CWT: $A \gets \text{CWT}(X_{-L+1:0}, \lambda) \in \mathbb{R}^{Bd \times L \times \lambda}$
    \STATE Partition spectrum by energy thresholds $(\mu_1, \mu_2)$ into $K=3$ sub-bands
    \STATE Mask and apply iWT: $X^k_{-L+1:0} \gets \text{iWT}(A^k)$ for $k = 1,\dots,K$

    \STATE // \textbf{Step 2: Patch Embedding}
    \STATE $E^{0} \gets \text{Embed}(X_{-L+1:0})$
    \STATE $E_k^{0} \gets \text{Embed}(X^k_{-L+1:0})$ for $k = 1,\dots,K$

    \STATE // \textbf{Step 3: Forward through $N$ FPA Layers}
    \STATE Initialize $H^{0} \gets E^{0}$
    \FOR{$n = 1$ to $N$}
        \STATE // FPA module (PPA,MPR,MPM, details omitted)
        \STATE $\{E_k^{n}\} \gets \text{PPA}(\{E_k^{n-1}\})$
        \STATE $P^n \gets \text{MPR}(\{E_k^{n}\}, H^{n-1})$
        \STATE $H^{n} \gets \text{BackboneBlock}(H^{n-1}, P^{n})$
        \STATE $\{E_k^{n}\} \gets \text{MPM}(\{E_k^{n}\}, H^{n})$
    \ENDFOR

    \STATE // \textbf{Step 4: Mixture of Predictors (MoP)}
    \STATE $E_{\text{avg}} \gets \frac{1}{K} \sum_{k=1}^K E_k^{N}$
    \STATE $S^{0} \gets E_{\text{avg}}$
    \FOR{$n = 1$ to $N$}
        \STATE $S^{n} \gets \text{Predictor}_n(S^{n-1}, H^{n-1})$
    \ENDFOR
    \STATE $\hat{X}_{1:H} \gets \text{OutputHead}(\text{Flatten}(S^{N}))$

    \STATE // \textbf{Step 5: Loss and Backpropagation}
    \STATE $\mathcal{L} \gets \text{MSE}(\hat{X}_{1:H}, X_{1:H})$
    \STATE Update model parameters via backpropagation
\ENDFOR
\end{algorithmic}
\end{algorithm}

\vspace{-5pt}
\section{Experiments} \label{section:experiments}
To validate the efficacy of \myformer functioning as a general fluctuation pattern assistance architecture, we conduct comprehensive experiments covering 8 popular temporal analytical scenarios. These encompass (1) long-term forecasting, (2) univariate and (3) multivariate short-term forecasting, (4) imputation, (5) classification, (6) anomaly detection, as well as (7) few-shot and (8) zero-shot forecasting. Collectively, as encapsulated in Figure~\ref{fig:intro_1}, \myformer invariably outperforms contemporary state-of-the-art across multiple tasks. 

\begin{table*}[htbp]
\centering
\small
\tabcolsep=0.15cm
\renewcommand\arraystretch{1.3}
\caption{
Comparison of the performance on \textbf{Full Shot Forecasting} task. We present the model's average performance across multiple prediction horizons for each dataset and boldface the best performance. 
The detailed results are presented in Table XVII.
% The detailed results are presented in Table~\ref{tab:forecasting_fullshot}.
}\label{tab:forecasting_fullshot_brief}
\vspace{-8pt}
    \resizebox{1.0\textwidth}{!}{
    \begin{tabular}{ccccccccccccccccccccccccc}
    \toprule
    \hline
    \multicolumn{1}{c}{\multirow{2}{*}{{Models}}} &
    \multicolumn{2}{c}{\textbf{Pets}} & 
    \multicolumn{2}{c}{DUET} & 
    \multicolumn{2}{c}{PDF} & 
    \multicolumn{2}{c}{iTransformer} & 
    \multicolumn{2}{c}{Pathformer} & 
    \multicolumn{2}{c}{FITS} & 
    \multicolumn{2}{c}{TimeMixer++} & 
    \multicolumn{2}{c}{TimeMixer} & 
    \multicolumn{2}{c}{PatchTST} & 
    \multicolumn{2}{c}{Crossformer} & 
    \multicolumn{2}{c}{TimesNet} & 
    \multicolumn{2}{c}{DLinear} \\
    
    \multicolumn{1}{c}{} & 
    \multicolumn{2}{c}{{(\textbf{Ours})}} & 
    \multicolumn{2}{c}{{\cite{qiu2025duet}}} & 
    \multicolumn{2}{c}{{\cite{dai2024pdf}}} & 
    \multicolumn{2}{c}{{\cite{liu2023itransformer}}} & 
    \multicolumn{2}{c}{{\cite{chen2024pathformer}}} & 
    \multicolumn{2}{c}{{\cite{xu2023fits}}} & 
    \multicolumn{2}{c}{{\cite{Wang2025TimeMixerAG}}} & 
    \multicolumn{2}{c}{{\cite{Wang2024TimeMixerDM}}} & 
    \multicolumn{2}{c}{{\cite{Nie2023PatchTST}}} & 
    \multicolumn{2}{c}{{\cite{zhang2023crossformer}}} & 
    \multicolumn{2}{c}{{\cite{wu2023timesnet}}} & 
    \multicolumn{2}{c}{{\cite{Zeng2022DLinear}}} \\
    
    \cline{2-25} 
    
    \multicolumn{1}{c}{{Metric}} & 
    MSE & MAE & MSE & MAE & MSE & MAE & MSE & MAE & MSE & MAE & 
    MSE & MAE & MSE & MAE & MSE & MAE & MSE & MAE & MSE & MAE & MSE & MAE & MSE & MAE \\
    
    \hline 
    
    \multicolumn{1}{c}{\multirow{1}{*}{\rotatebox{0}{ETTh1}}} &
    \textbf{0.395} & \textbf{0.407}
& 0.398 & 0.418 & 0.406 & 0.425 & 0.439 & 0.448 & 0.417 & 0.426 & 0.408 & 0.427 & 0.419 & 0.432
& 0.427 & 0.441 & 0.419 & 0.436 & 0.439 & 0.461 & 0.468 & 0.459 & 0.424 & 0.439 \\
    
    \multicolumn{1}{c}{\multirow{1}{*}{\rotatebox{0}{ETTh2}}} &
    0.340 & \textbf{0.378}
& \textbf{0.334} & 0.383 & 0.347 & 0.391 & 0.370 & 0.403 & 0.360 & 0.395 & 0.335 & 0.386 & 0.339 & 0.380
& 0.349 & 0.397 & 0.351 & 0.395 & 0.894 & 0.680 & 0.390 & 0.416 & 0.470 & 0.468 \\
    
    \multicolumn{1}{c}{\multirow{1}{*}{\rotatebox{0}{ETTm1}}} &
    \textbf{0.331} & \textbf{0.362}
& 0.338 & 0.369 & 0.342 & 0.376 & 0.361 & 0.390 & 0.357 & 0.374 & 0.357 & 0.377 & 0.369 & 0.378
& 0.355 & 0.380 & 0.349 & 0.381 & 0.464 & 0.455 & 0.407 & 0.415 & 0.356 & 0.378 \\
    
    \multicolumn{1}{c}{\multirow{1}{*}{\rotatebox{0}{ETTm2}}} &
    \textbf{0.245} & \textbf{0.299} 
& 0.247 & 0.307 & 0.250 & 0.313 & 0.268 & 0.327 & 0.253 & 0.308 & 0.254 & 0.313 & 0.269 & 0.320
& 0.257 & 0.318 & 0.256 & 0.314 & 0.501 & 0.505 & 0.292 & 0.331 & 0.259 & 0.324 \\
    
    \multicolumn{1}{c}{\multirow{1}{*}{\rotatebox{0}{Exchange}}} &
    0.291 & \textbf{0.362} 
& \textbf{0.280} & 0.364 & 0.350 & 0.397 & 0.360 & 0.403 & 0.384 & 0.414 & 0.349 & 0.396 & 0.357 & 0.391
& 0.381 & 0.416 & 0.322 & 0.385 & 0.389 & 0.423 & 0.406 & 0.437 & 0.292 & 0.391 \\
    
    \multicolumn{1}{c}{\multirow{1}{*}{\rotatebox{0}{Weather}}} &
    \textbf{0.209} & \textbf{0.245} 
& 0.218 & 0.252 & 0.227 & 0.263 & 0.232 & 0.270 & 0.225 & 0.258 & 0.243 & 0.280 & 0.226 & 0.262
& 0.226 & 0.264 & 0.223 & 0.261 & 0.233 & 0.292 & 0.255 & 0.282 & 0.242 & 0.293 \\
    
    \multicolumn{1}{c}{\multirow{1}{*}{\rotatebox{0}{Electricity}}} &
    \textbf{0.153} & \textbf{0.244} 
& 0.157 & 0.247 & 0.160 & 0.253 & 0.163 & 0.258 & 0.168 & 0.261 & 0.169 & 0.265 & 0.165 & 0.253
& 0.184 & 0.284 & 0.171 & 0.270 & 0.171 & 0.263 & 0.189 & 0.290 & 0.167 & 0.264 \\
     
    \multicolumn{1}{c}{\multirow{1}{*}{\rotatebox{0}{ILI}}} &
    1.679 & 0.839 
& \textbf{1.616} & \textbf{0.795} & 1.808 & 0.898 & 1.857 & 0.892 & 1.995 & 0.909 & 2.334 & 1.052 & 1.803 & 0.841
& 1.820 & 0.886 & 1.901 & 0.879 & 3.768 & 1.258 & 2.163 & 0.933 & 2.185 & 1.040 \\
    
    \multicolumn{1}{c}{\multirow{1}{*}{\rotatebox{0}{Solar}}} &
    \textbf{0.177} & \textbf{0.204} 
& 0.189 & 0.208 & 0.200 & 0.263 & 0.202 & 0.262 & 0.204 & 0.228 & 0.231 & 0.268 & 0.203 & 0.238
& 0.193 & 0.252 & 0.200 & 0.284 & 0.204 & 0.232 & 0.211 & 0.281 & 0.224 & 0.286 \\
    
    \multicolumn{1}{c}{\multirow{1}{*}{\rotatebox{0}{Traffic}}} &
    \textbf{0.383} & \textbf{0.239} 
& 0.393 & 0.256 & 0.395 & 0.270 & 0.397 & 0.281 & 0.416 & 0.264 & 0.429 & 0.302 & 0.416 & 0.264
& 0.409 & 0.279 & 0.397 & 0.275 & 0.521 & 0.282 & 0.617 & 0.327 & 0.418 & 0.287 \\

    \hline
    
    \multicolumn{1}{c}{$1^{\text{st}}$ Count} & 
    \multicolumn{2}{c}{\textbf{16}} & 
    \multicolumn{2}{c}{4} & 
    \multicolumn{2}{c}{0} & 
    \multicolumn{2}{c}{0} & 
    \multicolumn{2}{c}{0} & 
    \multicolumn{2}{c}{0} & 
    \multicolumn{2}{c}{0} & 
    \multicolumn{2}{c}{0} & 
    \multicolumn{2}{c}{0} & 
    \multicolumn{2}{c}{0} & 
    \multicolumn{2}{c}{0} & 
    \multicolumn{2}{c}{0} \\
    
    \hline
    \bottomrule
    \end{tabular}
    } 
\vspace{-10pt}
\end{table*}

\subsection{Experimental Setup}
\subsubsection{Datasets}
(1) For the full-shot and zero-shot forecasting, we use 10 well-acknowledged datasets from 6 different domains in TFB benchmark~\cite{qiu2024tfb} to comprehensively evaluate the performance. 
(2) For the shot-term forecasting,we utilize the M4 Competition dataset~\cite{Makridakis2018TheMC} and the PeMS dataset~\cite{Chen2001FreewayPM} as benchmarks. The M4 dataset comprises a hundred thousand marketing time steps, whereas the PeMS dataset incorporates four high-dimensional traffic network datasets.
(3) We conducted extensive experiments on the UEA Time Series Classification Archive~\cite{Bagnall2018TheUM} and five widely-used anomaly detection benchmarks~\cite{wu2023timesnet}, including SMD, SWaT, PSM, MSL and SMAP.
The above benchmarks ensure comprehensive evaluation across diverse scenarios.

\begin{table*}[htbp]
    \centering
    \small
    \tabcolsep=0.15cm
    \renewcommand\arraystretch{1.3}
    \caption{
    Short-term forecasting results in the M4 dataset with a single variate. All prediction lengths are in $\left[ 6,48 \right]$. A lower SMAPE, MASE or OWA indicates a better prediction. 
    The detailed results are presented in Table XX.
    % The detailed results are presented in Table~\ref{tab:short_m4}.
    }\label{tab:short_m4_brief}
    \vspace{-8pt}
    \resizebox{\textwidth}{!}{
    \begin{tabular}{cccccccccccccccccccccc}
    \toprule
    \hline
    
    \multicolumn{3}{c}{\multirow{2}{*}{Models}} & 
    \multicolumn{1}{c}{\rotatebox{0}{\scalebox{0.95}{\textbf{\myformer}}}}&
    \multicolumn{1}{c}{\rotatebox{0}{\scalebox{0.95}{{TimeMixer++}}}}&
    \multicolumn{1}{c}{\rotatebox{0}{\scalebox{0.95}{{TimeMixer}}}}&
    \multicolumn{1}{c}{\rotatebox{0}{\scalebox{0.95}{{iTransformer}}}}&
    \multicolumn{1}{c}{\rotatebox{0}{\scalebox{0.95}{{TiDE}}}}&
    \multicolumn{1}{c}{\rotatebox{0}{\scalebox{0.95}{TimesNet}}} &
    \multicolumn{1}{c}{\rotatebox{0}{\scalebox{0.95}{{N-HiTS}}}} &
    \multicolumn{1}{c}{\rotatebox{0}{\scalebox{0.95}{{N-BEATS}}}} &
    \multicolumn{1}{c}{\rotatebox{0}{\scalebox{0.95}{PatchTST}}} &
    \multicolumn{1}{c}{\rotatebox{0}{\scalebox{0.95}{MICN}}} &
    \multicolumn{1}{c}{\rotatebox{0}{\scalebox{0.95}{FiLM}}} &
    \multicolumn{1}{c}{\rotatebox{0}{\scalebox{0.95}{LightTS}}} &
    \multicolumn{1}{c}{\rotatebox{0}{\scalebox{0.95}{DLinear}}} &
    \multicolumn{1}{c}{\rotatebox{0}{\scalebox{0.95}{FED.}}} & 
    \multicolumn{1}{c}{\rotatebox{0}{\scalebox{0.95}{Stationary}}} & 
    \multicolumn{1}{c}{\rotatebox{0}{\scalebox{0.95}{Auto.}}} \\
    
    \multicolumn{1}{c}{}&\multicolumn{1}{c}{} &\multicolumn{1}{c}{} & 
    \multicolumn{1}{c}{\scalebox{0.95}{\textbf{(Ours)}}}& 
    \multicolumn{1}{c}{\scalebox{0.95}{\cite{Wang2025TimeMixerAG}}} &
    \multicolumn{1}{c}{\scalebox{0.95}{\cite{Wang2024TimeMixerDM}}} &
    \multicolumn{1}{c}{\scalebox{0.95}{\cite{liu2023itransformer}}} &
    \multicolumn{1}{c}{\scalebox{0.95}{\cite{Das2023LongtermTiDE}}} &
    \multicolumn{1}{c}{\scalebox{0.95}{\cite{wu2023timesnet}}} &
    \multicolumn{1}{c}{\scalebox{0.95}{\cite{Challu2022NHiTSNH}}} &
    \multicolumn{1}{c}{\scalebox{0.95}{\cite{Oreshkin2019nbeats}}} &
    \multicolumn{1}{c}{\scalebox{0.95}{\cite{Nie2023PatchTST}}} &
    \multicolumn{1}{c}{\scalebox{0.95}{\cite{wang2023micn}}} &
    \multicolumn{1}{c}{\scalebox{0.95}{\cite{zhou2022film}}} &
    \multicolumn{1}{c}{\scalebox{0.95}{\cite{Campos2023LightTSLT}}} &
    \multicolumn{1}{c}{\scalebox{0.95}{\cite{Zeng2022DLinear}}} &
    \multicolumn{1}{c}{\scalebox{0.95}{\cite{zhou2022fedformer}}} &
    \multicolumn{1}{c}{\scalebox{0.95}{\cite{Liu2022NonstationaryTR}}} &
    \multicolumn{1}{c}{\scalebox{0.95}{\cite{wu2021autoformer}}} \\
    
    \hline
    
    \multirow{3}{*}{\rotatebox{90}{\scalebox{0.9}{Weighted}}}& 
    \multirow{3}{*}{\rotatebox{90}{\scalebox{0.9}{Average}}} 
    &\scalebox{1.0}{SMAPE} & \textbf{\scalebox{1.0}{11.268}} & {\scalebox{1.0}{11.448}} &
    {\scalebox{1.0}{11.723}} &\scalebox{1.0}{12.684}&\scalebox{1.0}{13.950}&{\scalebox{1.0}{11.829}} &\scalebox{1.0}{11.927} &{\scalebox{1.0}{11.851}} &\scalebox{1.0}{13.152} &\scalebox{1.0}{19.638} &\scalebox{1.0}{14.863} &\scalebox{1.0}{13.525} &\scalebox{1.0}{13.639} &\scalebox{1.0}{12.840} &\scalebox{1.0}{12.780} &\scalebox{1.0}{12.909} \\
    
    & &  \scalebox{1.0}{MASE} & \textbf{\scalebox{1.0}{1.421}} & {\scalebox{1.0}{1.487}} &
    {\scalebox{1.0}{1.559}} &\scalebox{1.0}{1.764}&\scalebox{1.0}{1.940}&{\scalebox{1.0}{1.585}} &\scalebox{1.0}{1.613} &{\scalebox{1.0}{1.559}}  &\scalebox{1.0}{1.945}  &\scalebox{1.0}{5.947} &\scalebox{1.0}{2.207} &\scalebox{1.0}{2.111} &\scalebox{1.0}{2.095} &\scalebox{1.0}{1.701} &\scalebox{1.0}{1.756} &\scalebox{1.0}{1.771} \\
    
    & & \scalebox{1.0}{OWA} & \textbf{\scalebox{1.0}{0.784}} & {\scalebox{1.0}{0.821}} &
    {\scalebox{1.0}{0.840}} &\scalebox{1.0}{0.929}&\scalebox{1.0}{1.020}&{\scalebox{1.0}{0.851}} &\scalebox{1.0}{0.861} &{\scalebox{1.0}{0.855}} &\scalebox{1.0}{0.998} &\scalebox{1.0}{2.279} &\scalebox{1.0}{1.125}  &\scalebox{1.0}{1.051} &\scalebox{1.0}{1.051} &\scalebox{1.0}{0.918} &\scalebox{1.0}{0.930} &\scalebox{1.0}{0.939} \\
    
    \hline
    \bottomrule
    % \begin{tablenotes}
    % \footnotesize
    %     \item \large{$\ast$ The original paper of N-BEATS \cite{Oreshkin2019nbeats} adopts a special ensemble method to promote the performance. For fair comparisons, we remove the ensemble and only compare the pure forecasting models.}
    % \end{tablenotes}
    \end{tabular}
    } 
    \vspace{-10pt}
\end{table*}

%%%%%%%%%%%%%%%%%%%%%%%%%%%%%%%%%%%%%%%%%%%%%%%%%%%%%%%%%%%%%%%%%%%%%%%%%%%%%%%%%%%%%%%%%%%%%%%%%%%%%%%%

\begin{table*}[htbp]
    \centering
    \small
    \tabcolsep=0.15cm
    \renewcommand\arraystretch{1.3}
    \caption{
    Short-term forecasting results in the PEMS datasets with multiple variates. All input lengths are 96 and prediction lengths are 12. A lower MAE, MAPE or RMSE indicates a better prediction. 
    The detailed results are presented in Table XXI.
    % The detailed results are presented in Table~\ref{tab:short_pems}.
    }\label{tab:short_pems_brief}
    \vspace{-8pt}
    \resizebox{1.0\textwidth}{!}{
    \begin{tabular}{cccccccccccccccccccccc}
    \toprule
    \hline
    
    \multicolumn{2}{c}{\multirow{2}{*}{Models}} &
    \multicolumn{1}{c}{\rotatebox{0}{\scalebox{0.95}{\textbf{\myformer}}}}&
    \multicolumn{1}{c}{\rotatebox{0}{\scalebox{0.95}{TimeMixer++}}} &
    \multicolumn{1}{c}{\rotatebox{0}{\scalebox{0.95}{TimeMixer}}} &
    \multicolumn{1}{c}{\rotatebox{0}{\scalebox{0.95}{iTransformer}}} &
    \multicolumn{1}{c}{\rotatebox{0}{\scalebox{0.95}{TiDE}}} &
    \multicolumn{1}{c}{\rotatebox{0}{\scalebox{0.95}{SCINet}}} &
    \multicolumn{1}{c}{\rotatebox{0}{\scalebox{0.95}{Crossformer}}} &
    \multicolumn{1}{c}{\rotatebox{0}{\scalebox{0.95}{PatchTST}}} &
    \multicolumn{1}{c}{\rotatebox{0}{\scalebox{0.95}{TimesNet}}} &
    \multicolumn{1}{c}{\rotatebox{0}{\scalebox{0.95}{MICN}}} &
    \multicolumn{1}{c}{\rotatebox{0}{\scalebox{0.95}{DLinear}}} &
    \multicolumn{1}{c}{\rotatebox{0}{\scalebox{0.95}{FEDformer}}} & 
    \multicolumn{1}{c}{\rotatebox{0}{\scalebox{0.95}{Stationary}}} & 
    \multicolumn{1}{c}{\rotatebox{0}{\scalebox{0.95}{Autoformer}}} \\
    
    \multicolumn{2}{c}{}&\multicolumn{1}{c}{\scalebox{0.95}{\textbf{(Ours)}}} &
    \multicolumn{1}{c}{\scalebox{0.95}{\cite{Wang2025TimeMixerAG}}}&
    \multicolumn{1}{c}{\scalebox{0.95}{\cite{Wang2024TimeMixerDM}}}&
    \multicolumn{1}{c}{\scalebox{0.95}{\cite{liu2023itransformer}}}&
    \multicolumn{1}{c}{\scalebox{0.95}{\cite{Das2023LongtermTiDE}}}&
    \multicolumn{1}{c}{\scalebox{0.95}{\cite{Liu2021SCINetTS}}} &
    \multicolumn{1}{c}{\scalebox{0.95}{\cite{Du2021CrossDomainGD}}} &
    \multicolumn{1}{c}{\scalebox{0.95}{\cite{Nie2023PatchTST}}} &
    \multicolumn{1}{c}{\scalebox{0.95}{\cite{wu2023timesnet}}} &
    \multicolumn{1}{c}{\scalebox{0.95}{\cite{wang2023micn}}} &
    \multicolumn{1}{c}{\scalebox{0.95}{\cite{Zeng2022DLinear}}} &
    \multicolumn{1}{c}{\scalebox{0.95}{\cite{zhou2022fedformer}}} &
    \multicolumn{1}{c}{\scalebox{0.95}{\cite{Liu2022NonstationaryTR}}}&
    \multicolumn{1}{c}{\scalebox{0.95}{\cite{wu2021autoformer}}} \\

    \hline

    \multirow{3}{*}{\scalebox{1.0}{\rotatebox{0}{PEMS-avg$_{(03,04,07,08)}$}}} & 
    \scalebox{1.0}{MAE} & 
    \textbf{\scalebox{1.0}{15.53}} & 
    \scalebox{1.0}{15.91} & 
    \scalebox{1.0}{17.41} & 
    \scalebox{1.0}{19.87} & 
    \scalebox{1.0}{21.86} & 
    \scalebox{1.0}{19.12} & 
    \scalebox{1.0}{19.03} & 
    \scalebox{1.0}{23.01} & 
    \scalebox{1.0}{20.54} & 
    \scalebox{1.0}{19.34} & 
    \scalebox{1.0}{23.31} & 
    \scalebox{1.0}{23.50} & 
    \scalebox{1.0}{21.32} & 
    \scalebox{1.0}{22.62} \\
    
    & \scalebox{1.0}{MAPE} & 
    \textbf{\scalebox{1.0}{9.49}} & 
    \scalebox{1.0}{10.08} & 
    \scalebox{1.0}{10.59} & 
    \scalebox{1.0}{12.55} & 
    \scalebox{1.0}{13.80} & 
    \scalebox{1.0}{12.24} & 
    \scalebox{1.0}{12.22} & 
    \scalebox{1.0}{14.95} & 
    \scalebox{1.0}{12.69} & 
    \scalebox{1.0}{12.38} & 
    \scalebox{1.0}{14.68} & 
    \scalebox{1.0}{15.01} & 
    \scalebox{1.0}{14.09} & 
    \scalebox{1.0}{14.89} \\
    
    & \scalebox{1.0}{RMSE} & 
    \textbf{\scalebox{1.0}{26.11}} & 
    \scalebox{1.0}{27.06} & 
    \scalebox{1.0}{28.01} & 
    \scalebox{1.0}{31.29} & 
    \scalebox{1.0}{34.42} & 
    \scalebox{1.0}{30.12} & 
    \scalebox{1.0}{30.17} & 
    \scalebox{1.0}{36.05} & 
    \scalebox{1.0}{33.25} & 
    \scalebox{1.0}{30.40} & 
    \scalebox{1.0}{37.32} & 
    \scalebox{1.0}{36.78} & 
    \scalebox{1.0}{36.20} & 
    \scalebox{1.0}{34.49} \\
    
    \hline
    \bottomrule
    \end{tabular}
    } 
    \vspace{-10pt}
\end{table*}

\subsubsection{Baselines}
We choose the latest state-of-the-art proprietary methods to serve as baselines for full-shot forecasting, including CNN-based models (ModernTCN~\cite{luo2024moderntcn} and TimesNet~\cite{wu2023timesnet}), MLP-based models (FITS~\cite{xu2023fits}, TimeMixer~\cite{Wang2024TimeMixerDM}, and DLinear~\cite{Zeng2022DLinear}), and Transformer-based models (PDF~\cite{dai2024pdf}, Pathformer~\cite{chen2024pathformer}, iTransformer~\cite{liu2023itransformer}, PatchTST~\cite{Nie2023PatchTST} and Crossformer~\cite{zhang2023crossformer}). 
In zero-shot forecasting scenarios, we select well-established foundation models with strong generalization capabilities as baselines, including Decoder-only models (TimesFM~\cite{timesfm}), Encoder-only models (Moirai~\cite{woo2024moirai} and Moment~\cite{goswami2024moment}), and Encoder-Decoder models (Chronos~\cite{ansari2024chronos}).

\subsubsection{Key Settings for Fair Comparison}
We strictly adhere to the configuration of FoundTS~\cite{li2024foundts} and refrain from the ``Drop Last'' trick to ensure result fairness. 
Using the TFB codebase~\cite{qiu2024tfb}, we evaluate \myformer three times and report the averaged results, while the baseline results are sourced from the DUET~\cite{qiu2025duet} papers.
We strictly adhere to the configuration of FoundTS~\cite{li2024foundts} and refrain from the ``Drop Last'' trick to ensure result fairness. 
Furthermore, for the anomaly detection and classification experiments, we re-divided the dataset to avoid the problem of ``test leakage issue''~\cite{talukder2024totem}.

\subsubsection{Implementation Details}
To ensure consistency with existing studies~\cite{qiu2025duet,zhou2021informer,woo2024moirai}, we set four prediction lengths for the ILI dataset: 24, 36, 48 and 60; for the other nine datasets, we adopt four prediction lengths: 96, 192, 336 and 720.
To keep consistent with previous works~\cite{qiu2025duet,li2024foundts}, we repeatedly conduct experiments across multiple horizons for all baselines and report the best results across all horizons. For ILI, the optional horizons are 36, 104, For other datasets, the optional horizons are 96, 336, 512. 
All experiments are conducted via Python 3.10 and PyTorch~\cite{paszke2019pytorch} 2.6.0 on NVIDIA V100 GPUs. We utilize L2 loss function and Adam optimizer, with the batch size from 16 to 64.
The details are provided in Section VII of appendix.
% The details are provided in Section~\ref{sec:detail} of appendix. 

\begin{figure}[t]
\begin{center}
\centerline{\includegraphics[width=1.0\columnwidth]{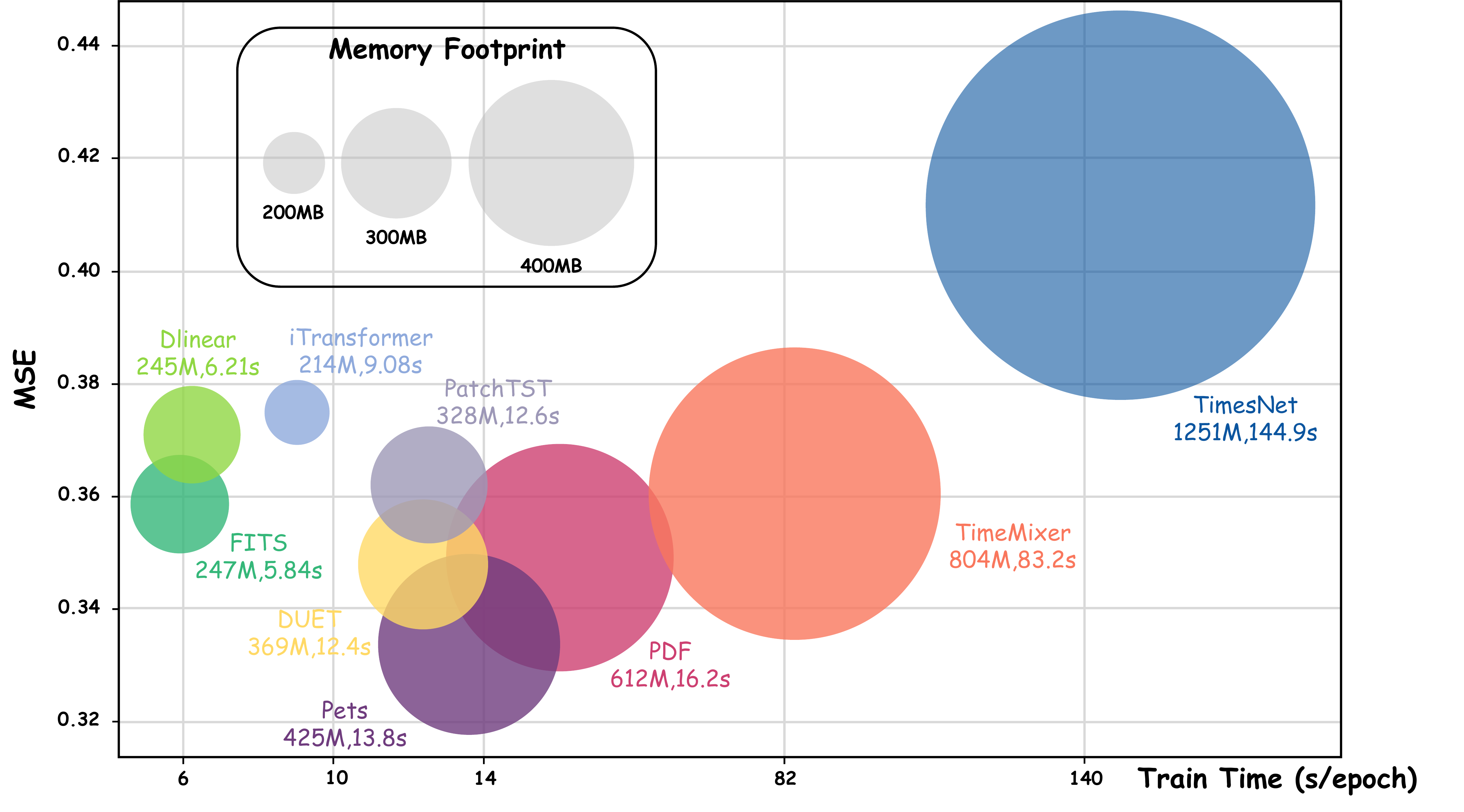}}
\vspace{-10pt}
\caption{
Comparison of the efficiency of \myformer and other baselines on ETTm2.
}\label{fig:efficiency}
\end{center}
\vspace{-25pt}
\end{figure}

\subsection{Long-term Forecasting}
\subsubsection{Setups} Long-term forecasting emerges as a cornerstone for strategic scheming in arenas like traffic governance and energy exploitation. To conduct a thoroughgoing evaluation of our model, a battery of experiments is executed on 8 ubiquitously-employed real-world datasets, incorporating those related to Weather, Solar-Energy, Electricity, and Traffic. These tasks conform to the precedent benchmarks established by ~\cite{haoyietal-informer-2021,wu2021autoformer,Liu2021SCINetTS}, thus guaranteeing comparability and relevance throughout our research undertakings.

\subsubsection{Results} Table~\ref{tab:forecasting_fullshot_brief} shows \myformer outperforms other models in long-term forecasting across various datasets. The MSE of \myformer is reduced by \textbf{8.7\%} and \textbf{15.1\%} compared to the DUET and iTransformer, respectively. For ETT (Avg), \myformer achieves \textbf{7.4\%} lower MSE than PDF. Specifically, it outperforms the runner-up model by a margin of \textbf{13.4\%} and \textbf{19.7\%} on the challenging Solar-Energy, which demonstrates the potential of \myformer as a general enhanced architecture.

\begin{table*}[htbp]
    \centering
    \small
    \tabcolsep=0.15cm
    \renewcommand\arraystretch{1.3}
    \caption{Comparison of the performance on \textbf{Zero-shot Transfer} task, including two sets of baselines, time series proprietary model and foundation model. We present the average performance across four prediction lengths.}\label{tab:forecasting_zeroshot_brief}
    \vspace{-8pt}
    \resizebox{1.0\textwidth}{!}{
    \begin{tabular}{ccccccccccccccc|ccccccccc}
    \toprule
    \hline

    &
    \multicolumn{14}{c|}{Time Series Proprietary Model} &
    &
    \multicolumn{6}{c}{Time Series Foundation Model} \\
    
    \multicolumn{1}{c}{\multirow{2}{*}{Models}} &
    \multicolumn{2}{c}{\textbf{\myformer}} & 
    \multicolumn{2}{c}{TimeMixer++} & 
    \multicolumn{2}{c}{LLMTime} &
    \multicolumn{2}{c}{GPT4TS} & 
    \multicolumn{2}{c}{PatchTST} & 
    \multicolumn{2}{c}{iTransformer} & 
    \multicolumn{2}{c|}{TimesNet} & 
    \multicolumn{1}{c}{\multirow{2}{*}{Models}} &
    \multicolumn{2}{c}{TimesFM} &
    \multicolumn{2}{c}{Moment} &
    \multicolumn{2}{c}{Chronos(L)} \\
    
    \multicolumn{1}{c}{} & 
    \multicolumn{2}{c}{{(\textbf{Ours})}} & 
    \multicolumn{2}{c}{{\cite{Wang2025TimeMixerAG}}} &
    \multicolumn{2}{c}{{\cite{llmtime}}} &
    \multicolumn{2}{c}{{\cite{Zhou2023OneFA}}} & 
    \multicolumn{2}{c}{{\cite{Nie2023PatchTST}}} &
    \multicolumn{2}{c}{{\cite{liu2023itransformer}}} &
    \multicolumn{2}{c|}{{\cite{wu2023timesnet}}} &
    &
    \multicolumn{2}{c}{{\cite{Das2024TimesFM}}} &
    \multicolumn{2}{c}{{\cite{goswami2024moment}}} &
    \multicolumn{2}{c}{{\cite{ansari2024chronos}}} \\

    \hline
    
    \multicolumn{1}{c}{Metric} & 
    MSE & MAE & MSE & MAE & MSE & MAE & MSE & MAE & MSE & MAE & MSE & MAE & MSE & MAE & Metric &
    MSE & MAE & MSE & MAE & MSE & MAE \\
    
    \hline
    
    \multirow{1}{*}{\rotatebox{0}{ETT\scalebox{0.9}{h2$\rightarrow$h1}}} & 
    \textbf{0.452} & \textbf{0.463} & 
    0.511 & 0.498 & 1.961 & 0.981 & 0.757 & 0.578 & 0.565 & 0.513 & 0.552 & 0.511 & 0.865 & 0.621 &
    ETT\scalebox{0.9}{h1} &
    \textbf{0.473} & \textbf{0.443} & 0.683 & 0.566 & 0.588 & 0.466 \\

    \multirow{1}{*}{\rotatebox{0}{ETT\scalebox{0.9}{h1$\rightarrow$h2}}} & 
    \textbf{0.355} & 0.405 & 
    0.367 & \textbf{0.391} & 0.992 & 0.708 & 0.406 & 0.422 & 0.380 & 0.405 & 0.481 & 0.474 & 0.421 & 0.431 &
    ETT\scalebox{0.9}{h2} &
    0.392 & 0.406 & \textbf{0.361} & 0.409 & 0.455 & 0.427 \\

    \multirow{1}{*}{\rotatebox{0}{ETT\scalebox{0.9}{m2$\rightarrow$m1}}} & 
    \textbf{0.384} & \textbf{0.394} & 
    0.427 & 0.448 & 1.933 & 0.984 & 0.769 & 0.567 & 0.568 & 0.492 & 0.559 & 0.491 & 0.769 & 0.567 &
    ETT\scalebox{0.9}{m1} &
    0.433 & 0.418 & 0.670 & 0.536 & 0.555 & 0.465 \\

    \multirow{1}{*}{\rotatebox{0}{ETT\scalebox{0.9}{m1$\rightarrow$m2}}} & 
    \textbf{0.273} & \textbf{0.316} & 
    0.291 & 0.331 & 1.867 & 0.869 & 0.313 & 0.348 & 0.296 & 0.334 & 0.324 & \textbf{0.331} & 0.322 & 0.354 &
    ETT\scalebox{0.9}{m2} &
    0.328 & 0.346 & 0.316 & 0.365 & 0.295 & 0.338 \\

    \hline
    \bottomrule
    \end{tabular}
    } 
\vspace{-10pt}
\end{table*}

\begin{table*}[htbp]
    \centering
    \small
    \tabcolsep=0.15cm
    \renewcommand\arraystretch{1.3}
    \caption{Comparison of the complete performance with diverse prediction length on few-data (10\% data) long-term forecasting task. \textbf{Avg.} is averaged from all four prediction lengths, that $\{96, 192, 336, 720\} $. We boldface the best performance. 
    The detailed results are presented in Table XVIII.
    % The detailed results are presented in Table~\ref{tab:forecasting_fewshot}.
    }\label{tab:new_few_shot}
    \vspace{-8pt}
    \resizebox{\textwidth}{!}{
    \begin{tabular}{ccccccccccccccccccccccccc}
    \toprule
    \hline
    \multicolumn{1}{c}{\multirow{2}{*}{{Models}}} &
    \multicolumn{2}{c}{{\textbf{\myformer}}} &
    \multicolumn{2}{c}{{TimeMixer++}} &
    \multicolumn{2}{c}{{iTransformer}} &
    \multicolumn{2}{c}{{TiDE}} &
    \multicolumn{2}{c}{{DLinear}} &
    \multicolumn{2}{c}{{PatchTST}} &
    \multicolumn{2}{c}{{TimesNet}} &
    \multicolumn{2}{c}{{FEDformer}} &
    \multicolumn{2}{c}{{Autoformer}} &
    \multicolumn{2}{c}{{Stationary}} &
    \multicolumn{2}{c}{{ETSformer}} &
    \multicolumn{2}{c}{{LightTS}} \\

    \multicolumn{1}{c}{} & 
    \multicolumn{2}{c}{{(\textbf{Ours})}} & 
    \multicolumn{2}{c}{{\cite{Wang2025TimeMixerAG}}} &
    \multicolumn{2}{c}{{\cite{liu2023itransformer}}}& 
    \multicolumn{2}{c}{{\cite{Das2023LongtermTiDE}}} &
    \multicolumn{2}{c}{{\cite{Zeng2022DLinear}}} &
    \multicolumn{2}{c}{{\cite{Nie2023PatchTST}}} & 
    \multicolumn{2}{c}{{\cite{wu2023timesnet}}} & 
    \multicolumn{2}{c}{{\cite{zhou2022fedformer}}} & 
    \multicolumn{2}{c}{{\cite{wu2021autoformer}}} & 
    \multicolumn{2}{c}{{\cite{Liu2022NonstationaryTR}}} &  
    \multicolumn{2}{c}{{\cite{woo2022etsformer}}} & 
    \multicolumn{2}{c}{{\cite{Campos2023LightTSLT}}} \\

    \hline 

    \multicolumn{1}{c}{{Metric}} & 
    MSE & MAE & MSE & MAE & MSE & MAE & MSE & MAE & MSE & MAE & MSE & MAE & MSE & MAE &
    MSE & MAE & MSE & MAE & MSE & MAE & MSE & MAE & MSE & MAE \\
    
    \hline 

    \multicolumn{1}{c}{\multirow{1}{*}{\rotatebox{0}{Weather}}} & 
    \textbf{0.229} & \textbf{0.265} &
    {0.241} & {0.271} &
    {0.291} & {0.331} &
    {0.249} & {0.291} &
    {0.241} & {0.283} &
    {0.242} & {0.279} &
    {0.279} & {0.301} &
    {0.284} & {0.324} &
    {0.300} & {0.342} &
    {0.318} & {0.323} &
    {0.318} & {0.360} &
    {0.289} & {0.322} \\
    
    \multicolumn{1}{c}{\multirow{1}{*}{\rotatebox{0}{Electricity}}} & 
    \textbf{0.162} & \textbf{0.256} &
    {0.168} & {0.271} &
    {0.241} & {0.337} &
    {0.196} & {0.289} &
    {0.180} & {0.280} &
    {0.180} & {0.273} &
    {0.323} & {0.392} &
    {0.346} & {0.427} &
    {0.431} & {0.478} &
    {0.444} & {0.480} &
    {0.660} & {0.617} &
    {0.441} & {0.489} \\
    
    \multicolumn{1}{c}{\multirow{1}{*}{\rotatebox{0}{Traffic}}} & 
    \textbf{0.406} & \textbf{0.277} &
    {0.469} & {0.299} &
    {0.604} & {0.365} &
    {0.491} & {0.313} &
    {0.447} & {0.313} &
    {0.430} & {0.305} &
    {0.951} & {0.535} &
    {0.663} & {0.425} &
    {0.749} & {0.446} &
    {1.453} & {0.815} &
    {1.914} & {0.936} &
    {1.248} & {0.684} \\
    
    \multicolumn{1}{c}{\multirow{1}{*}{\rotatebox{0}{ETTh1}}} & 
    \textbf{0.446} & \textbf{0.447} &
    {0.613} & {0.520} &
    {0.510} & {0.597} &
    {0.589} & {0.535} &
    {0.691} & {0.600} &
    {0.633} & {0.542} &
    {0.869} & {0.628} &
    {0.639} & {0.561} &
    {0.702} & {0.596} &
    {0.915} & {0.639} &
    {1.180} & {0.834} &
    {1.375} & {0.877} \\
    
    \multicolumn{1}{c}{\multirow{1}{*}{\rotatebox{0}{ETTh2}}} & 
    \textbf{0.357} & \textbf{0.398} &
    {0.402} & {0.433} &
    {0.455} & {0.461} &
    {0.395} & {0.412} &
    {0.605} & {0.538} &
    {0.415} & {0.431} &
    {0.479} & {0.465} &
    {0.466} & {0.475} &
    {0.488} & {0.499} &
    {0.462} & {0.455} &
    {0.894} & {0.713} &
    {2.655} & {1.160} \\
    
    \multicolumn{1}{c}{\multirow{1}{*}{\rotatebox{0}{ETTm1}}} & 
    \textbf{0.354} & \textbf{0.383} &
    {0.487} & {0.461} &
    {0.491} & {0.516} &
    {0.425} & {0.458} &
    {0.411} & {0.429} &
    {0.501} & {0.466} &
    {0.677} & {0.537} &
    {0.722} & {0.605} &
    {0.802} & {0.628} &
    {0.797} & {0.578} &
    {0.980} & {0.714} &
    {0.971} & {0.705} \\
    
    \multicolumn{1}{c}{\multirow{1}{*}{\rotatebox{0}{ETTm2}}} & 
    \textbf{0.261} & \textbf{0.321} &
    {0.311} & {0.367} &
    {0.375} & {0.412} &
    {0.317} & {0.371} &
    {0.316} & {0.368} &
    {0.296} & {0.343} &
    {0.320} & {0.353} &
    {0.463} & {0.488} &
    {1.342} & {0.930} &
    {0.332} & {0.366} &
    {0.447} & {0.487} &
    {0.987} & {0.756} \\

    \hline
    
    \multicolumn{1}{c}{$1^{\text{st}}$ Count} & 
    \multicolumn{2}{c}{\textbf{69}} & 
    \multicolumn{2}{c}{0} & 
    \multicolumn{2}{c}{0} & 
    \multicolumn{2}{c}{0} & 
    \multicolumn{2}{c}{0} & 
    \multicolumn{2}{c}{0} & 
    \multicolumn{2}{c}{0} & 
    \multicolumn{2}{c}{0} & 
    \multicolumn{2}{c}{0} & 
    \multicolumn{2}{c}{0} & 
    \multicolumn{2}{c}{0} & 
    \multicolumn{2}{c}{0} \\
    \hline
    \bottomrule
    \end{tabular}
    } 
    \vspace{-10pt}
\end{table*}

\subsection{Short-term Forecasting}
\subsubsection{Setups} Short-term forecasting assumes a critical function within the purview of financial decision-making and market risk evaluation tasks. To appraise the short-term forecasting capabilities under both univariate and multivariate configurations, we utilize the M4 Competition dataset~\cite{Makridakis2018TheMC} and the PeMS dataset~\cite{Chen2001FreewayPM} as benchmarks. In where, the M4 dataset comprises a hundred thousand marketing time steps, whereas the PeMS dataset incorporates four high-dimensional traffic network datasets. These comprehensive benchmarks furnish a formidable evaluating platform for an exhaustive assessment of the model's efficacy and robustness.

\subsubsection{Results} The results enumerated in Table~\ref{tab:short_m4_brief} and \ref{tab:short_pems_brief} manifest the state-of-the-art performance of \myformer within all short-term forecasting benchmarks. In the M4 dataset, compared to the challenging TimeMixer and TimesNet, \myformer accomplishes a reduction of \textbf{5.6\%} and \textbf{7.2\%} in MASE, respectively. Within the PEMS, in comparison to the leading TimeMixer and iTransformer, \myformer reduces MAPE by \textbf{5.7\%} and \textbf{20.3\%}. Remarkably, in comparison with the conventional PatchTST, \myformer exhibits performance augmentations on M4 and PEMS that exceed \textbf{24.4\%} and \textbf{33.1\%}, which highlights the potential of \myformer as the novel benchmark.

\vspace{-15pt}
\subsection{Imputation}The endeavor of imputing missing values profound significance for time series analysis. For instance, its pivotal role in rectifying the absent values extant in wearable sensors. 
To appraise the imputation capabilities of our model, Table~\ref{tab:new_imputation} demonstrates the performance of \myformer in interpolating missing values across 6 datasets. Notably, \myformer achieves uniform state-of-the-art performance, especially, compared to the existing TimesNet and TimeMixer, \myformer effects an average diminution of \textbf{26.8\%} and \textbf{44.3\%} in MSE.
\begin{table}[htbp]
    \centering
    \small
    \tabcolsep=0.6mm
    \renewcommand\arraystretch{1.3}
    \vspace{-5pt}
    \caption{Comparison of the performance on \textbf{Imputation} task. We present the average result across multiple mask ratios.}\label{tab:new_imputation}
    \vspace{-8pt}
    \resizebox{0.475\textwidth}{!}{
    \begin{tabular}{ccccccccccccccc}
    \toprule
    \hline
    
    \multicolumn{1}{c}{\multirow{2}{*}{Models}} & 
    \multicolumn{2}{c}{\rotatebox{0}{{\textbf{\myformer}}}} &
    \multicolumn{2}{c}{\rotatebox{0}{{TimeMixer++}}} &
    \multicolumn{2}{c}{\rotatebox{0}{{iTrans.}}} &
    \multicolumn{2}{c}{\rotatebox{0}{{PatchTST}}} &
    \multicolumn{2}{c}{\rotatebox{0}{{FEDformer}}} & 
    \multicolumn{2}{c}{\rotatebox{0}{{TIDE}}} & 
    \multicolumn{2}{c}{\rotatebox{0}{{TimesNet}}} \\
    
    \multicolumn{1}{c}{} & 
    \multicolumn{2}{c}{{(\textbf{Ours})}} & 
    \multicolumn{2}{c}{{\cite{Wang2025TimeMixerAG}}} &
    \multicolumn{2}{c}{{\cite{liu2023itransformer}}} &
    \multicolumn{2}{c}{{\cite{Nie2023PatchTST}}} & 
    \multicolumn{2}{c}{{\cite{zhou2022fedformer}}} & 
    \multicolumn{2}{c}{{\cite{Das2023LongtermTiDE}}} & 
    \multicolumn{2}{c}{{\cite{wu2023timesnet}}} \\

    \cline{1-15}
    
    \multicolumn{1}{c}{{Metric}} & 
    {MSE} &  {MAE} &  {MSE} &  {MAE} &  {MSE} &  {MAE} &  {MSE} &
    {MAE} &  {MSE} &  {MAE} &  {MSE} &  {MAE} &  {MSE} &  {MAE} \\

    \hline
    
    {ETTm1} &
    \textbf{{0.037}} & \textbf{{0.121}} & 
    {0.052} & {0.178} & {0.075} & {0.177} & {0.097} & {0.194} & {0.048} & {0.152} & 
    {0.090} & {0.210} & {0.049} & {0.147} \\
    
    {ETTm2} &
    \textbf{{0.034}} & {{0.145}} & 
    {0.051} & {0.166} & {0.055} & {0.169} & {0.080} & {0.183} & {0.087} & {0.198} & 
    {0.169} & {0.263} & {0.035} & \textbf{0.124} \\
    
    {ETTh1} &
    \textbf{{0.078}} & \textbf{{0.179}} & 
    {0.092} & {0.212} & {0.130} & {0.213} & {0.178} & {0.231} & {0.099} & {0.225} & 
    {0.289} & {0.395} & {0.142} & {0.258} \\
    
    {ETTh2} &
    \textbf{{0.058}} & \textbf{{0.143}} & 
    {0.066} & {0.194} & {0.125} & {0.259} & {0.124} & {0.293} & {0.262} & {0.344} & 
    {0.709} & {0.596} & {0.088} & {0.198} \\

    {ECL} &
    {\textbf{{0.099}}} & \textbf{{0.185}} &  
    {0.109} & {0.197} & {0.140} & {0.223} & {0.129} & {0.198} & {0.181} & {0.314} & 
    {0.182} & {0.202} & {0.135} & {0.255} \\
    
    {Weather} &
    \textbf{{0.044}} & \textbf{{0.073}} & 
    {0.049} & {0.078} & {0.095} & {0.102} & {0.082} & {0.149} & {0.064} & {0.139} &
    {0.063} & {0.131} & {0.061} & {0.098} \\

    \hline
    \bottomrule
    \end{tabular}
    } 
\vspace{-10pt}
\end{table}

\subsection{Few-shot Forecasting}
\subsubsection{Setups} Models with sufficient robustness are necessitated to efficiently capture temporal patterns in challenging environments, which play an irreplaceable role in real-world prediction tasks within limited data. To evaluate the adaptability of \myformer to sparse data and its competence in discerning universal fluctuation information in recognition tasks, we trained all baselines utilizing merely $10\%$ of the time steps across 7 datasets. 

\subsubsection{Results} Table~\ref{tab:new_few_shot} presents the state-of-the-art performance achieved by \myformer in few-shot learning. In comparison with TimeMixer and iTransformer, \myformer exhibits an average performance enhancement of \textbf{16.3\%} and \textbf{25.6\%} in MSE. These results validate that spectral augmentation strategy can effectively capture patterns under the challenge of limited data.

\begin{figure}[t]
\begin{center}
\centerline{\includegraphics[width=1.0\columnwidth]{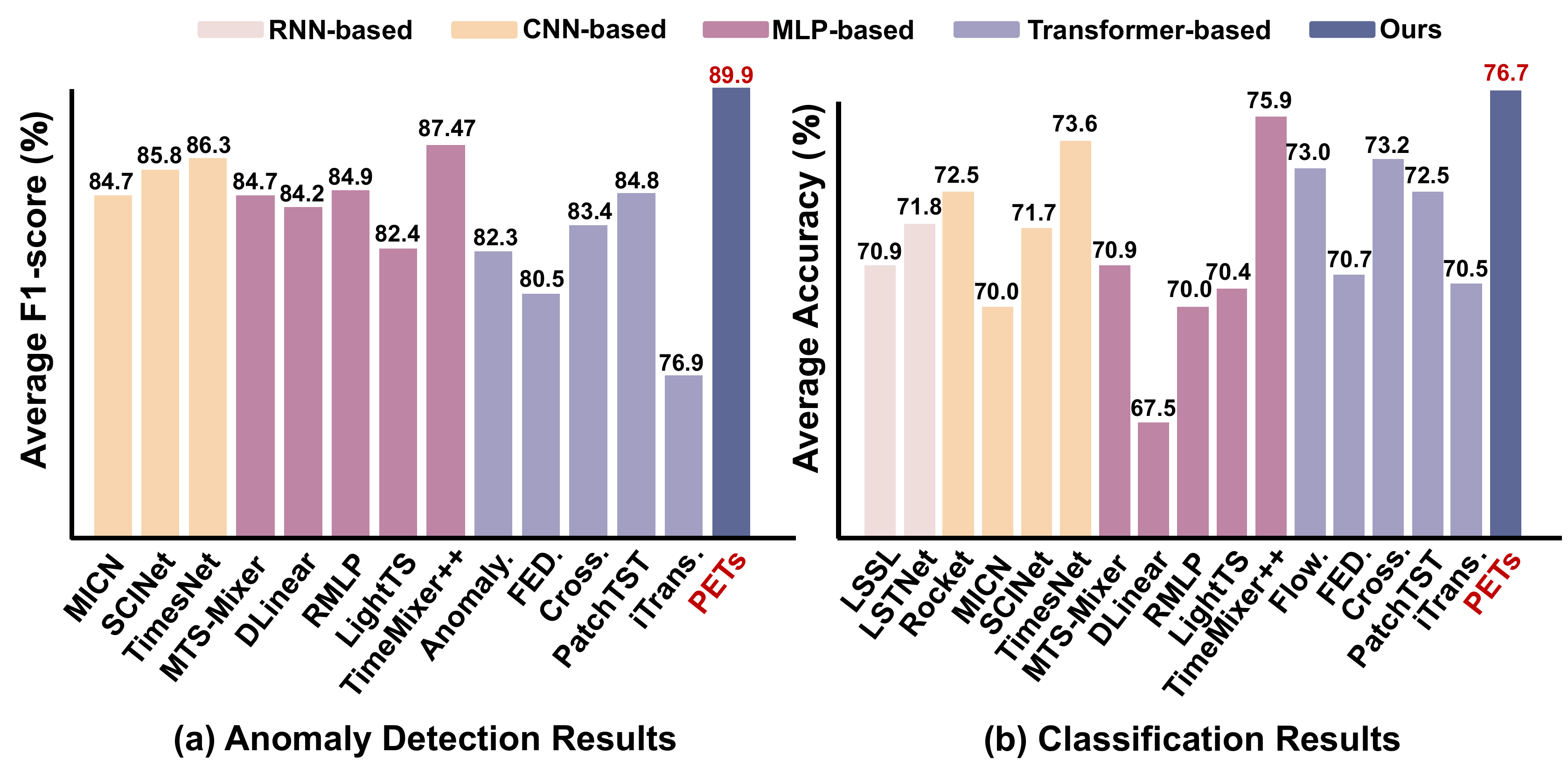}}
\vspace{-4mm}
\caption{
Results of classification and anomaly detection. The results are averaged from several datasets. Higher accuracy and F1 score indicate better performance. 
The detailed results are presented in Table XXII and XXIII.
% The detailed results are presented in Table~\ref{tab:anomaly} and ~\ref{tab:classification}.
}\label{fig:anomaly_and_classification}
\end{center}
\vspace{-15pt}
\end{figure}

\subsection{Zero-shot and Transfer Forecasting} \label{sec:zero_shot}
\subsubsection{Setups} Following the popular zero-shot transfer setup in GPT4TS~\cite{Zhou2023OneFA}, we evaluated models' ability to generalize across different contexts. As shown in the left of Table~\ref{tab:forecasting_zeroshot_brief}, models trained on dataset $D_a$ are evaluated on unseen dataset $D_b$ without further training. This direct transfer ($D\scalebox{0.8}{a$\rightarrow$b}$) tests models' adaptability and predictive robustness across disparate datasets. 

\subsubsection{Results} The right of Table~\ref{tab:forecasting_zeroshot_brief} presents the zero-shot forecasting of the time series foundation model, which is directly evaluated on the unseen dataset after pre-training.
In this paper, zero-shot inference is harnessed to appraise the robustness and cross-domain generalization capabilities of the model. In Table~\ref{tab:forecasting_zeroshot_brief}, \myformer exhibits consistent state-of-the-art performance across all transfer scenarios. In comparison with the existing TimeMixer and iTransformer, \myformer achieves a significant reduction of \textbf{16.3\%} and \textbf{12.0\%} in the average MSE across all transfer scenarios. Notably, the remarkable cross-domain generalization ability demonstrated by \myformer in zero-shot learning validates its potential to become a large-scale time series foundation model.

\begin{figure*}[t]
\begin{center}
\vspace{-5pt}
\centerline{\includegraphics[width=2.0\columnwidth]{images/representation_1.pdf}}
\vspace{-10pt}
\caption{Analysis of the dependency relationships within each single fluctuation pattern. 
More representation visualization and analysis is in the appendix XI-C.
% More representation visualization and analysis is in the appendix~\ref{sec:inter_pattern}.
}\label{fig:representation_intra}
\end{center}
\vspace{-20pt}
\end{figure*}

\subsection{Anomaly Detection and Classification}
\subsubsection{Setup} Anomaly detection and classification are fundamental tasks that evaluate the ability to capture subtle mutation information from temporal fluctuations, playing a critical role in applications such as fault diagnosis and predictive maintenance. Due to the differing inductive biases of discriminative and generative tasks, developing a universal model remains a significant challenge, as these tasks require balancing pattern recognition and data distribution modeling. To address this, we conducted extensive experiments on the UEA Time Series Classification Archive~\cite{Bagnall2018TheUM} and five widely-used anomaly detection benchmarks~\cite{wu2023timesnet}, ensuring a comprehensive evaluation across diverse scenarios.

\subsubsection{Results} On the benchmarks of classification and anomaly detection, \myformer has achieved consistent state-of-the-art performance in Figure~\ref{fig:anomaly_and_classification}, with the accuracy rate and F1-score reaching \textbf{74.7\%} and \textbf{89.9\%} respectively. Notably, compared with popular transformer models such as PatchTST and iTransformer, \myformer achieves respective enhancements of \textbf{5.9\%} and \textbf{16.9\%} in terms of accuracy and f1-score. It should be noted that the attention architecture augmented by a fluctuation pattern, for the first time, simultaneously attained optimal performance in both generative and discriminative tasks, which shows \myformer can capture universal patterns from diverse scenarios.

\begin{table}[htbp]
\caption{We conducted 4 ablation experiments on all 5 benchmarks. The results evaluate the effectiveness of components.}
\vspace{-8pt}
 \label{tab:ablation}
\vskip 0.02in
\centering
\resizebox{1.0\columnwidth}{!}{
\begin{tabular}{lccccccc}
        \toprule
            &
            Weather &
            Solar &
            Electricity &
            Traffic &
            % ETTh1 &
            ETTh2 &
            % ETTm1 &
            % ETTm2 &
            Average &
            Promotion \\
        \midrule
           \rowcolor{blue!10}
            \myformer &
            \textbf{0.219} &
            \textbf{0.187} &
            \textbf{0.173} &
            \textbf{0.403} &
            % \textbf{0.415} &
            \textbf{0.360} &
            % \textbf{0.351} &
            % \textbf{0.255} &
            \textbf{0.268} &
            \textbf{-} \\
            w/o FPA &
            0.249 &
            0.242 &
            0.196 &
            0.448 &
            0.398 &
            0.307 &
            14.6\% \\
            w/o PPA &
            0.229 &
            0.199 &
            0.183 &
            0.427 &
            % 0.436 &
            0.383 &
            % 0.368 &
            % 0.269 &
            0.284 &
            5.97\% \\
            w/o MPR &
            0.238 &
            0.202 &
            0.189 &
            0.437 &
            % 0.449 &
            0.392 &
            % 0.380 &
            % 0.275 &
            0.291 &
            8.58\% \\
            w/o MPM &
            0.232 &
            0.197 &
            0.181 &
            0.423 &
            % 0.438 &
            0.376 &
            % 0.370 &
            % 0.271 &
            0.282 &
            5.22\% \\
            w/o MoP &
            0.235 &
            0.208 &
            0.184 &
            0.433 &
            % 0.446 &
            0.383 &
            % 0.371 &
            % 0.274 &
            0.289 &
            7.84\% \\
        \bottomrule
    \end{tabular}
   }
\vspace{-10pt}
\end{table}

\subsection{Model Analysis}
\subsubsection{Ablation Study} 
To elucidate which components \myformer benefits from, we conducted a comprehensive ablation study on the model architecture. As presented in Table~\ref{tab:ablation}, we found that removing (w/o) the FPA led to a notable decline in performance. These results may be attributed to the fact that the proposed fluctuation pattern
assisted (FPA) module utilizes the captured fluctuation patterns as conditional context, guiding the backbone blocks to focus on modeling specific fluctuation patterns. Specifically, in the challenging multivariate benchmarks such as Solar, Electricity, and Traffic, the proposed fluctuation pattern assisted approach enhanced performance by \textbf{29.4\%}, \textbf{13.3\%}, and \textbf{11.2\%} respectively. Similar results were obtained on other datasets, demonstrating the superiority of this design. In addition, we have also conducted comprehensive ablations on each basic component.

\begin{table*}[htbp]
    \centering
    \small
    \tabcolsep=0.15cm
    \renewcommand\arraystretch{1.3}
    \caption{Evaluate the generality of the \myformer, \textbf{Orig.} as original baseline, \textbf{+Pets.} as pattern-enhanced version by proposed SDAQ and FPA architecture, and \textbf{Imp.} demonstrates the improvement. 
    The detailed results are presented in Table XXIV.
    % The detailed results are presented in Table~\ref{tab:forecasting_enhanced}.
    }\label{tab:forecasting_enhanced_brief}
    \vspace{-8pt}
    \resizebox{1.0\textwidth}{!}{
    \begin{tabular}{c|c|cccccccccccccccc}
    \toprule
    \hline
    \multicolumn{2}{c}{\multirow{1}{*}{Datasets}} &
    \multicolumn{2}{c}{Weather} & 
    \multicolumn{2}{c}{Solar} & 
    \multicolumn{2}{c}{Electricity} & 
    \multicolumn{2}{c}{Traffic} & 
    \multicolumn{2}{c}{ETTh1} &
    \multicolumn{2}{c}{ETTh2} &
    \multicolumn{2}{c}{ETTm1} &
    \multicolumn{2}{c}{ETTm2} \\
    
    \cline{3-18} 
    
    \multicolumn{2}{c}{\multirow{1}{*}{Metrics}} & MSE & MAE & MSE & MAE & MSE & MAE &
    MSE & MAE & MSE & MAE & MSE & MAE & MSE & MAE & MSE & MAE \\
    
    \hline
    
    \multicolumn{1}{c|}{\multirow{3}{*}{\rotatebox{0}{\shortstack{PatchTST\\ \cite{Nie2023PatchTST}}}}} 
    & Orig. & 
    {0.265} & {0.285} &
    {0.287} & {0.333} &
    {0.216} & {0.318} &
    {0.529} & {0.341} &
    {0.516} & {0.484} &
    {0.391} & {0.411} &
    {0.406} & {0.407} &
    {0.290} & {0.334} \\
    & \textbf{+Pets.} & 
    {0.219} & {0.255} &
    {0.187} & {0.244} &
    {0.178} & {0.269} &
    {0.478} & {0.299} &
    {0.415} & {0.426} &
    {0.360} & {0.399} &
    {0.351} & {0.382} &
    {0.255} & {0.310} \\
    & \cellcolor{blue!10}\textbf{Imp.} & 
    \cellcolor{blue!10}{\textbf{17.4\%}} & \cellcolor{blue!10}{\textbf{10.5\%}} & 
    \cellcolor{blue!10}{\textbf{34.8\%}} & \cellcolor{blue!10}{\textbf{26.7\%}} & 
    \cellcolor{blue!10}{\textbf{17.6\%}} & \cellcolor{blue!10}{\textbf{15.4\%}} & 
    \cellcolor{blue!10}{\textbf{9.6\%}} & \cellcolor{blue!10}{\textbf{12.3\%}} & 
    \cellcolor{blue!10}{\textbf{19.6\%}} & \cellcolor{blue!10}{\textbf{12.0\%}} & 
    \cellcolor{blue!10}{\textbf{7.9\%}} & \cellcolor{blue!10}{\textbf{2.9\%}} & 
    \cellcolor{blue!10}{\textbf{13.5\%}} & \cellcolor{blue!10}{\textbf{6.1\%}} & 
    \cellcolor{blue!10}{\textbf{12.1\%}} & \cellcolor{blue!10}{\textbf{7.2\%}} \\
    
    \hline 
    
    \multicolumn{1}{c|}{\multirow{3}{*}{\rotatebox{0}{\shortstack{TimeMixer\\ \cite{Wang2024TimeMixerDM}}}}} 
    & Orig. & 
    {0.240} & {0.271} &
    {0.216} & {0.280} &
    {0.182} & {0.272} &
    {0.484} & {0.297} &
    {0.447} & {0.440} &
    {0.364} & {0.395} &
    {0.381} & {0.395} &
    {0.275} & {0.323} \\
    & \textbf{+Pets.} & 
    {0.224} & {0.264} &
    {0.188} & {0.251} &
    {0.161} & {0.254} &
    {0.414} & {0.278} &
    {0.411} & {0.421} &
    {0.357} & {0.398} &
    {0.352} & {0.384} &
    {0.263} & {0.319} \\
    & \cellcolor{blue!10}\textbf{Imp.} & 
    \cellcolor{blue!10}{\textbf{6.7\%}} & \cellcolor{blue!10}{\textbf{2.6\%}} & 
    \cellcolor{blue!10}{\textbf{13.0\%}} & \cellcolor{blue!10}{\textbf{10.4\%}} & 
    \cellcolor{blue!10}{\textbf{11.5\%}} & \cellcolor{blue!10}{\textbf{6.6\%}} & 
    \cellcolor{blue!10}{\textbf{14.5\%}} & \cellcolor{blue!10}{\textbf{6.4\%}} & 
    \cellcolor{blue!10}{\textbf{8.1\%}} & \cellcolor{blue!10}{\textbf{4.3\%}} & 
    \cellcolor{blue!10}{\textbf{1.9\%}} & \textbf{-0.8\%} & 
    \cellcolor{blue!10}{\textbf{7.6\%}} & \cellcolor{blue!10}{\textbf{2.8\%}} & 
    \cellcolor{blue!10}{\textbf{4.4\%}} & \cellcolor{blue!10}{\textbf{1.3\%}} \\

    \hline 
    
    \multicolumn{1}{c|}{\multirow{3}{*}{\rotatebox{0}{\shortstack{DLinear\\ \cite{Zeng2022DLinear}}}}} 
    & Orig. & 
    {0.265} & {0.315} &
    {0.330} & {0.401} &
    {0.225} & {0.319} &
    {0.625} & {0.383} &
    {0.461} & {0.457} &
    {0.563} & {0.519} &
    {0.404} & {0.408} &
    {0.354} & {0.402} \\
    & \textbf{+Pets.} & 
    {0.245} & {0.277} &
    {0.259} & {0.276} &
    {0.189} & {0.283} &
    {0.504} & {0.323} &
    {0.409} & {0.422} &
    {0.376} & {0.383} &
    {0.354} & {0.373} &
    {0.255} & {0.313} \\
    & \cellcolor{blue!10}\textbf{Imp.} & 
    \cellcolor{blue!10}{\textbf{8.3\%}} & \cellcolor{blue!10}{\textbf{12.1\%}} & 
    \cellcolor{blue!10}{\textbf{32.9\%}} & \cellcolor{blue!10}{\textbf{31.2\%}} & 
    \cellcolor{blue!10}{\textbf{16.0\%}} & \cellcolor{blue!10}{\textbf{11.3\%}} & 
    \cellcolor{blue!10}{\textbf{19.4\%}} & \cellcolor{blue!10}{\textbf{15.7\%}} & 
    \cellcolor{blue!10}{\textbf{11.6\%}} & \cellcolor{blue!10}{\textbf{7.7\%}} & 
    \cellcolor{blue!10}{\textbf{33.2\%}} & \cellcolor{blue!10}{\textbf{26.2\%}} & 
    \cellcolor{blue!10}{\textbf{13.1\%}} & \cellcolor{blue!10}{\textbf{8.6\%}} & 
    \cellcolor{blue!10}{\textbf{36.0\%}} & \cellcolor{blue!10}{\textbf{22.1\%}} \\

    \hline 
    
    \multicolumn{1}{c|}{\multirow{3}{*}{\rotatebox{0}{\shortstack{TimesNet\\ \cite{wu2023timesnet}}}}} 
    & Orig. & 
    {0.251} & {0.294} &
    {0.403} & {0.374} &
    {0.193} & {0.304} &
    {0.620} & {0.336} &
    {0.495} & {0.450} &
    {0.414} & {0.427} &
    {0.401} & {0.406} &
    {0.291} & {0.333} \\
    & \textbf{+Pets.} & 
    {0.241} & {0.280} &
    {0.374} & {0.340} &
    {0.172} & {0.270} &
    {0.578} & {0.314} &
    {0.459} & {0.420} &
    {0.392} & {0.397} &
    {0.373} & {0.377} &
    {0.265} & {0.297} \\
    & \cellcolor{blue!10}\textbf{Imp.} & 
    \cellcolor{blue!10}{\textbf{4.0\%}} & \cellcolor{blue!10} {\textbf{4.7\%}} & 
    \cellcolor{blue!10}{\textbf{7.3\%}} & \cellcolor{blue!10}{\textbf{9.1\%}} & 
    \cellcolor{blue!10}{\textbf{10.7\%}} & \cellcolor{blue!10} {\textbf{11.1\%}} & 
    \cellcolor{blue!10}{\textbf{6.7\%}} & \cellcolor{blue!10} {\textbf{6.6\%}} & 
    \cellcolor{blue!10}{\textbf{7.2\%}} & \cellcolor{blue!10}{\textbf{6.7\%}} & 
    \cellcolor{blue!10}{\textbf{5.4\%}} & \cellcolor{blue!10}{\textbf{7.1\%}} & 
    \cellcolor{blue!10}{\textbf{6.7\%}} & \cellcolor{blue!10}{\textbf{7.2\%}} & 
    \cellcolor{blue!10}{\textbf{8.8\%}} & \cellcolor{blue!10}  {\textbf{10.8\%}} \\
    
    \hline
    \bottomrule
    \end{tabular}
    } 
\vspace{-10pt}
\end{table*}

\subsubsection{Applicability Study of the Augmentation Strategy} 
Notably, the proposed Spectrum Decomposition and Amplitude Quantization (SDAQ) and Fluctuation Pattern Assisted (FPA) strategy can be seamlessly integrated into various deep models, in a plug-and-play fashion. This prompts us to investigate its generality by inserting FPA into diverse types of structures. We selected representative baselines composed of divergent underlying architectures. The results in Table~\ref{tab:forecasting_enhanced_brief} validate that the augmentation strategy can significantly elevate model predictive capability. Concretely, the attention-based PatchTST witnessed performance improvements of \textbf{34.8\%} and \textbf{17.6\%} on Solar and ECL respectively. The Linear-based TimeMixer and DLinear achieved enhancements of \textbf{14.5\%} and \textbf{19.4\%} on Traffic. The CNN-based TimesNet showed an improvement of \textbf{8.8\%} on ETTm2. These findings authenticate that proposed augmentation strategy is applicable to various architectures.

\begin{table}[htpb]
  \vspace{-5pt}
  \caption{To verify the sensitivity of \myformer to hyperparameter selection, we show the performance of \myformer with different parameter scales on multiple benchmarks, with a fixed forecasting window of 96, and a performance metric of MSE.}\label{tab:exp_sensitivity_brief}
  \vspace{-8pt}
  \centering
  \resizebox{1.\columnwidth}{!}{
  \begin{small}
  \renewcommand{\multirowsetup}{\centering}
  \tabcolsep=0.15cm
  \renewcommand\arraystretch{1.3}
  \begin{tabular}{ccccccccc}
    \toprule
    \hline
    
    \multicolumn{2}{c}{\multirow{1}{*}{\scalebox{1.0}{Hyperparameter}}} & 
    \multicolumn{1}{c}{\rotatebox{0}{\scalebox{1.0}{ETTh1}}} & 
    \multicolumn{1}{c}{\rotatebox{0}{\scalebox{1.0}{ETTh2}}} & 
    \multicolumn{1}{c}{\rotatebox{0}{\scalebox{1.0}{ETTm1}}} & 
    \multicolumn{1}{c}{\rotatebox{0}{\scalebox{1.0}{ETTm2}}} & 
    \multicolumn{1}{c}{\rotatebox{0}{\scalebox{1.0}{Weather}}} & 
    \multicolumn{1}{c}{\rotatebox{0}{\scalebox{1.0}{ECL}}} & 
    \multicolumn{1}{c}{\rotatebox{0}{\scalebox{1.0}{Traffic}}} \\
    \hline

    \multicolumn{1}{c}{\multirow{3}{*}{\rotatebox{0}{$\mu_1,\mu_2$=0.3,0.6}}} &
    \multirow{1}{*}{\scalebox{1.0}{\shortstack{L=3,D=16}}} &
    0.398 & 0.327 & 0.304 & 0.170 & 0.155 & 0.139 & 0.394 \\
    \cline{2-9}

    & \multirow{1}{*}{\scalebox{1.0}{\shortstack{L=4,D=32}}} &
    0.389 & 0.322 & 0.289 & 0.177 & 0.152 & 0.144 & 0.412 \\
    \cline{2-9}
    
    & \multirow{1}{*}{\scalebox{1.0}{\shortstack{L=6,D=128}}} &
    0.395 & 0.331 & 0.295 & 0.173 & 0.159 & 0.148 & 0.415 \\
    \hline

    \multicolumn{1}{c}{\multirow{3}{*}{\rotatebox{0}{$\mu_1,\mu_2$=0.4,0.8}}} &
    \multirow{1}{*}{\scalebox{1.0}{\shortstack{L=3,D=16}}} &
    0.371 & \textbf{0.292} & 0.291 & 0.167 & 0.146 & 0.140 & \textbf{0.375} \\
    \cline{2-9}
    
    & \multirow{1}{*}{\scalebox{1.0}{\shortstack{L=4,D=32}}} &
    0.378 & 0.304 & 0.279 & \textbf{0.163} & 0.145 & 0.135 & 0.381 \\
    \cline{2-9}

    & \multirow{1}{*}{\scalebox{1.0}{\shortstack{L=6,D=128}}} &
    0.383 & 0.315 & 0.292 & 0.170 & 0.152 & 0.141 & 0.392 \\
    \hline
    
    \multicolumn{1}{c}{\multirow{3}{*}{\rotatebox{0}{$\mu_1,\mu_2$=0.7,0.9}}} &
    \multirow{1}{*}{\scalebox{1.0}{\shortstack{L=3,D=16}}} &
    \textbf{0.364} & 0.298 & \textbf{0.274} & 0.164 & \textbf{0.142} & \textbf{0.132} & 0.378 \\
    \cline{2-9}

    & \cellcolor{blue!10}{\multirow{1}{*}{\scalebox{1.0}{\shortstack{L=4,D=32}}}} &
    \cellcolor{blue!10}{0.385} & \cellcolor{blue!10}{0.334} & \cellcolor{blue!10}{0.289} & \cellcolor{blue!10}{0.164} & \cellcolor{blue!10}{0.154} & \cellcolor{blue!10}{0.145} & \cellcolor{blue!10}{0.388} \\
    \cline{2-9}

    & \multirow{1}{*}{\scalebox{1.0}{\shortstack{L=6,D=128}}} &
    0.389 & 0.331 & 0.282 & 0.173 & 0.159 & 0.143 & 0.395 \\
    \hline
    
    \bottomrule
  \end{tabular}
  \end{small}
}
\vspace{-5pt}
\end{table}

\subsubsection{Model Hyperparameter Sensitivity Analysis} 
In the design of SDAQ, the energy in time series is mainly concentrated in the low-frequency bands. Additionally, when the number of divided frequency bands is excessive, limited information falling within the high-frequency range will be further split. This leads to the decoupled series containing a large amount of noise and a small amount of information, thereby reducing the predictive performance of the model. Based on these considerations, we fix the number of decoupled fluctuation patterns as $K\textit{=}3$, and set the energy ratio as $(\mu_1,\mu_2)$. 

In the design of component FPA and MoP, the model consists of $N$ consecutive composite layers. Each layer contains single adapter block (PPA), single backbone block, single MPM, and single MPR. Correspondingly, the hybrid predictor is also composed of $N$ predictors with the same structure. Additionally, both the original sequence and the decoupled sequence yield $P_L$ tokens after the patch-based embedding operation, where the dimension of each token is $P_d$.

To verify the sensitivity of \myformer to hyperparameter selection, the Table~\ref{tab:exp_sensitivity_brief} shows the performance of \myformer with different parameter scales on multiple benchmarks. Where the model parameter combinations include \begin{small}$(N,P_d) \textit{=} (3,16), (4,32), (6,128)$\end{small} and \begin{small}$(\mu_1,\mu_2) \textit{=} (0.3,0.6), (0.4,0.8), (0.7,0.9)$\end{small}, and a performance metric of MSE. By default, the hyperparameters of \myformer are fixed to $N\textit{=}4$, $P_d\textit{=}32$ and $(\mu_1,\mu_2)\textit{=}(0.7,0.9)$, all experimental results presented follow this setting.

\subsubsection{Efficiency comparison}
Figure~\ref{fig:efficiency} presents a comprehensive comparison between \myformer and diverse baselines regarding memory consumption, training time, and performance on the ETTm2 dataset with the forecasting length of 720. 
Notably, \myformer achieves an optimal balance between performance and efficiency through adaptive scaling of the prototype waves.

\subsubsection{Comprehensive complexity analysis}
In this section, we provide a detailed analysis of the computational complexity of Pets’ three main components (SDAQ, FPA, and MoP) and address the specific concerns regarding wavelet transform, spectral quantization, and multi-stage predictors. Pets consists of three main components: SDAQ, FPA, and MoP. 
Details in the Section IX-D of appendix.
% Details in the Section~\ref{sec:complexity_analysis} of appendix.

The computational complexity of SDAQ involves three steps: (1) transforming an input sequence of length $L$ into a time-frequency spectrogram with $\lambda$ scales using an FFT-optimized Continuous Wavelet Transform (CWT), which reduces convolution operations to $O(\lambda\cdot L\log L)$; (2) partitioning the spectrogram into three subgraphs based on energy distribution, an operation with negligible complexity; and (3) reconstructing each subgraph into the time domain via inverse wavelet transform (iWT), with the same $O(\lambda\cdot L\log L)$ complexity. Together, SDAQ operates with an overall complexity $O(\lambda\cdot L\log L)$. Used in SDAQ, the wavelet transform (CWT and iWT) operates with a combined complexity of $O(\lambda\cdot L\log L)$, introducing computational overhead but maintaining scalability through FFT optimization.

The FPA and MoP modules consist of $N$ composite layers, each containing lightweight operations such as 1D convolutions, self-attention, and linear layers. After patching, the tensor shape is $Bd\times P_L\times P_d$, where $P_L$ is the number of tokens and $P_d$ is the token dimension. 
These modules have an overall complexity of $O(L+LP_d)$, ensuring efficient processing while capturing complex patterns. Regarding the specific processes mentioned:

\begin{figure*}[t]
\begin{center}
\vspace{-5pt}
\centerline{\includegraphics[width=2.0\columnwidth]{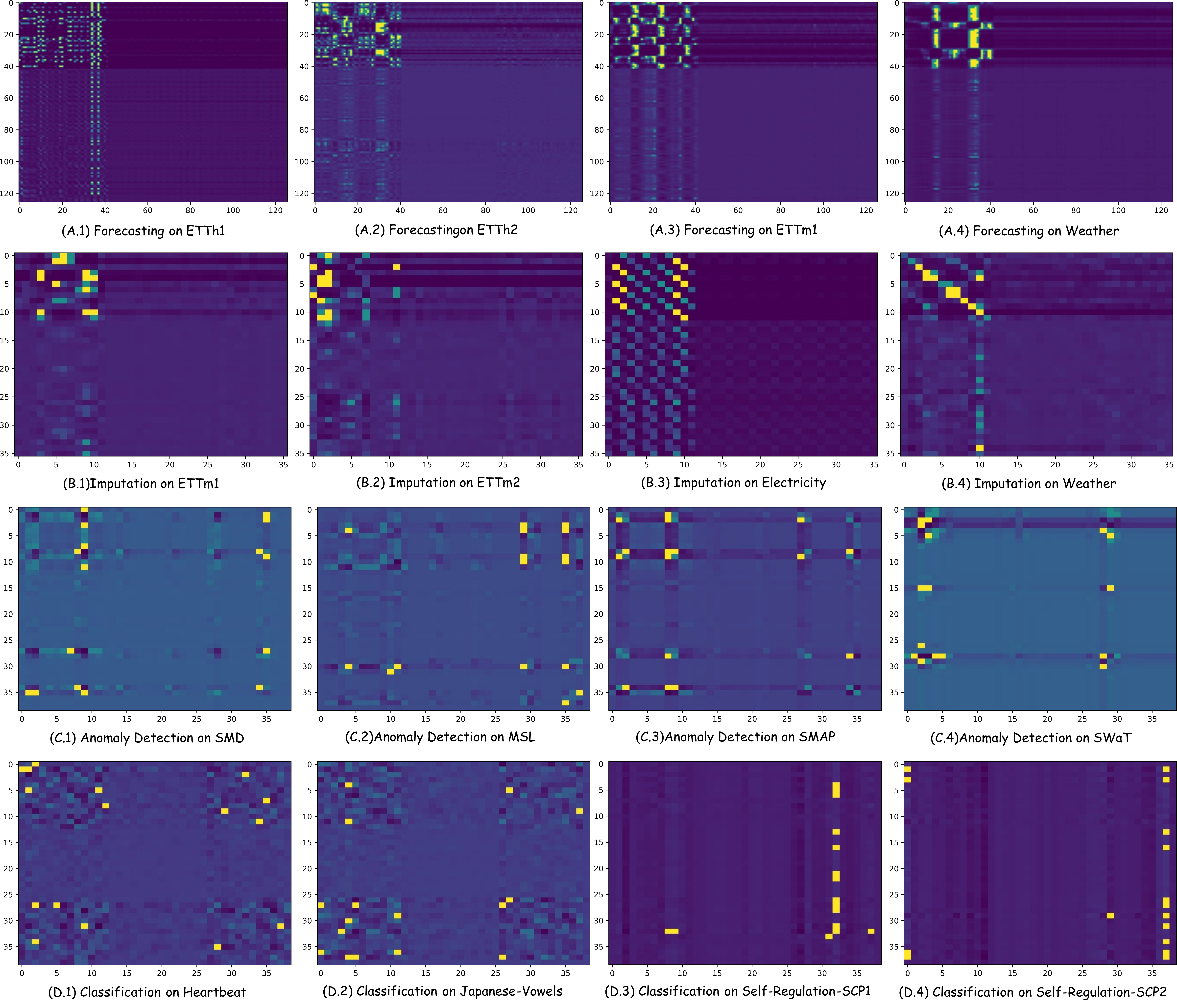}}
\vspace{-5pt}
\caption{Analysis of the dependency relationships within each single fluctuation pattern. 
More representation visualization and analysis is in the appendix XI-C.
% More representation visualization and analysis is in the appendix~\ref{sec:inter_pattern}.
}\label{fig:representation_inter}
\end{center}
\vspace{-20pt}
\end{figure*}

\subsection{Representation Analysis and Visualization} 
\subsubsection{Intra-pattern Dependencies}
In Fig.~\ref{fig:representation_intra}, the original time series samples are decoupled into three independent fluctuation modes, and the fluctuation modes 1 to 3 exhibit the pattern characteristics from long-to-short period and low-to-high frequency respectively. We then compute the global dependencies between all time points in each fluctuation pattern, and the top plot shows the attention score visualization for original sequence. 
Specifically, in the original sequence, the dependencies between time points are chaotic, which is caused by the mutual coupling between multiple periodic patterns in the original sequence. In contrast, the decoupled fluctuation patterns show distinct periodic variation characteristics. Concretely, in the fluctuation pattern 1, the bright yellow blob shows a sparse distribution characteristic, which means that the attention mechanism learns the dependence between long periods from it, which we call the long wave pattern. However, in the fluctuation mode 3, the bright yellow blobs show the characteristics of clustered distribution, and the attention learns the dependence between short periods from it, which we call the short-wave mode.

\subsubsection{Inter-pattern Dependencies}
In Fig.~\ref{fig:representation_inter}, we present the visualization of the importance of different fluctuation patterns for specific downstream tasks. 
We first decouple the observed sequence into three fluctuating sequences. Subsequently, each of these fluctuating sequences undergoes a patch-based embedding operation to obtain three groups of pattern tokens, where each set contains 40 tokens. Finally, all 120 tokens are concatenated in the order from low-frequency pattern to high-frequency pattern, and then fed into the transformer, where the attention scores between these tokens in different tasks are calculated.

Specifically, for prediction (Fig.~\ref{fig:representation_inter} A.1-A.4) and imputation (Fig.~\ref{fig:representation_inter} B.1-B.4) tasks, the model favors learning dependencies among low-frequency fluctuation tokens (\begin{small}$x\in [0,40]$\end{small}), and the intra-pattern score distribution shows periodical variation. This is also in line with cognition, since long-term trends are important for prediction tasks, and high-frequency oscillations in them are likely due to noise. In contrast, in the anomaly detection (Fig.~\ref{fig:representation_inter} C.1-C.4) and classification (Fig.~\ref{fig:representation_inter} D.1-D.4) tasks, the yellow bright spot distribution will appear in the small right corner and the bottom left corner, which means that the high-frequency perturbation is critical for the classification and anomaly detection tasks. It can be understood that abrupt signals are often caused by reasons such as mechanical failures, and therefore these high-frequency patterns are very important for discriminative tasks.

\begin{table}[htpb]
  \vspace{-10pt}
  \caption{We compared the trade-off between performance and efficiency of Pets built with CWT versus FFT on ETTm2 (prediction horizon 720).}\label{tab:cwt_fft}
  \vspace{-8pt}
  \centering
  \resizebox{1.\columnwidth}{!}{
  \begin{small}
  \renewcommand{\multirowsetup}{\centering}
  \tabcolsep=0.15cm
  \renewcommand\arraystretch{1.3}
  \begin{tabular}{cccc}
    \toprule

    & MSE & Memory (MiB) & Train Time (s/epoch) \\

    \hline
    
    Pets(CWT) & 0.344 & 425 & 13.8 \\
    Pets(FFT) & 0.351 & 349 & 9.5 \\
    
    \bottomrule
  \end{tabular}
  \end{small}
}
\vspace{-15pt}
\end{table}

\subsection{Limitation on Time Cost and FFT-enhanced Solution}
Although \myformer effectively decouples different frequency components, computing the spectrogram via CWT in SDAQ is time-consuming. To address this trade-off, we propose replacing CWT with Fast Fourier Transform (FFT) in SDAQ. The enhanced Pets can still decouple the original sequence into sub-sequences with distinct fluctuation patterns, while significantly reducing SDAQ's computational cost to match that of popular models, as shown in Table~\ref{tab:cwt_fft}.

\begin{table}[htpb]
  \vspace{-10pt}
  \caption{We compared the performance of different wavelet functions on both generative and discriminative tasks.}\label{tab:wavelet}
  \vspace{-8pt}
  \centering
  \resizebox{1.\columnwidth}{!}{
  \begin{small}
  \renewcommand{\multirowsetup}{\centering}
  \tabcolsep=0.15cm
  \renewcommand\arraystretch{1.3}
  \begin{tabular}{cccc}
    \toprule
	
    & ETTh1 (MSE) & SMD (F1-Score) & Heartbeat (Accuracy) \\

    \hline
    
    Meyer & 0.412 & 88.91 & 69.9 \\
    Morlet & 0.401 & 89.75 & 71.5 \\
    \rowcolor{blue!10}
    Haar (Ours) & 0.395 & 90.21 & 76.5 \\
    
    \bottomrule
  \end{tabular}
  \end{small}
}
\vspace{-15pt}
\end{table}

\subsection{Experiment on Elimination of Wave Functions}
We choose the Haar wavelet~\cite{xu2023haar} function in \myformer. As a compactly supported orthogonal wavelet function, Haar wavelets only consider local regions during decomposition, leading to high computational efficiency that helps reduce Pets' overall complexity. 
Additionally, Haar wavelets excel in decomposing high-frequency components, enhancing Pets' generalization capability for handling diverse fluctuation patterns across frequencies. 

It is necessary to investigate the performance of various wavelet functions across different downstream tasks. 
To this end, we select representative wavelets such as Morlet~\cite{lin2000morlet} and Meyer~\cite{meyer1992meyer} as alternatives. 
We evaluate the model's sensitivity to different wavelet functions in both generative and discriminative tasks. 
The results in Table~\ref{tab:wavelet} demonstrate that smoother wavelets (e.g., Meyer) may over-smooth sharp transitions and local details in time series, leading to performance degradation. 
Moreover, the Haar wavelet achieves the highest computational efficiency, demonstrating its overall advantage in terms of performance, robustness, and efficiency.

\section{Conclusion}\label{section:conclusion}
We propose Pets, a universal Fluctuation Pattern-Assisted Temporal Architecture. It maximizes the modeling and generalization power of existing methods across different data domains and tasks. Pets identifies the underlying hybrid fluctuation patterns in time series by adaptively separating multi-periodic patterns. Meanwhile, PPA and MoP aggregate these patterns based on energy ranking, ensuring task-relevant patterns drive the result.

% \clearpage
\bibliographystyle{IEEEtrans}
\normalem
\bibliography{references_master}

%%%%%%%%%%%%%%%%%%%%%%%%%%%%%%%%%%%%%%%%%%%%%%%%%%%%%%%%%%%%%%%%%%%%%%%%%%%%%%%%%
% \clearpage
\vspace{-1.0cm}
\begin{IEEEbiography}[{\includegraphics[width=1in,height=1.25in,clip,keepaspectratio]{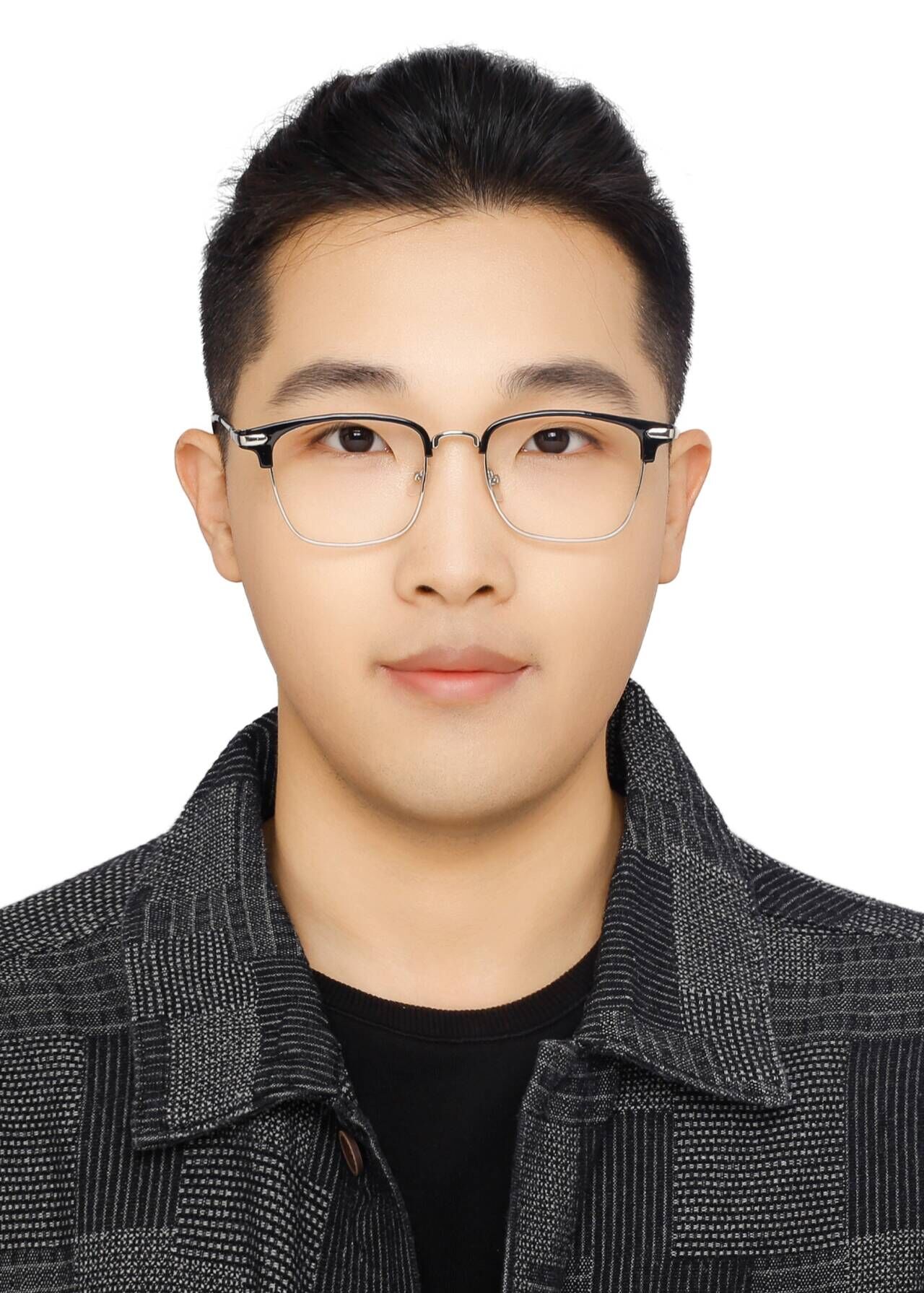}}]
  {Xiangkai Ma} is currently pursuing his Ph.D. at the School of Computer Science, Nanjing University. He earned his bachelor’s degree from the University of Electronic Science and Technology of China in 2022. His research interests include time series analysis, spatiotemporal data mining, large multimodal models, and embodied intelligence. He has authored multiple papers as the first author in top-tier journals and conferences such as ICDE and TKDD, etc.
\end{IEEEbiography}
\vspace{-1.0cm}
\begin{IEEEbiography}[{\includegraphics[width=1in,height=1.25in,clip,keepaspectratio]{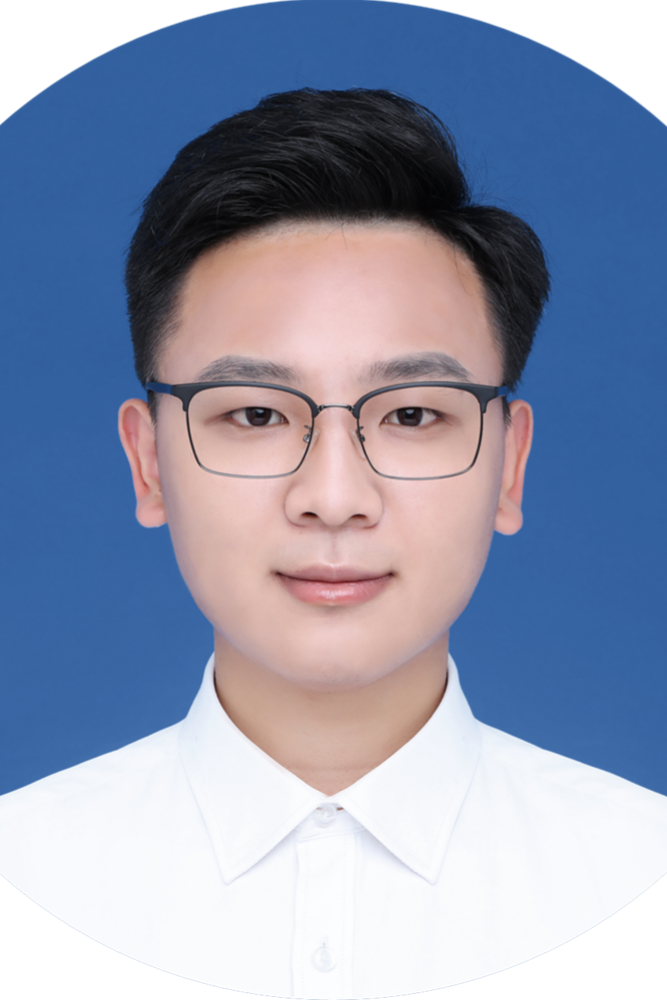}}]
  {Xiaobin Hong} received the MSc degree from Nanjing University of Science and Technology (211), Nanjing, China, and he is a current Ph.D student at the Department of Computer Science at Nanjing University (985/211) from 2022. He was the recipient of the Outstanding Master's Thesis award of the Jiangsu Computer Society. He has published about 20+ journal and conference papers, including JAS, Information Fusion, TKDD, AAAI, ICDE, ECCV, ICLR, ICRA, etc. He has served on the review committee of several journals and conferences, including TKDE, TOMM, AAAI, ICLR, CVPR, etc. His current research interests include Data Mining, Graph Learning, Time Series Analysis, LLM Reasoning, and AI4Science.
\end{IEEEbiography}
\vspace{-1.0cm}
\begin{IEEEbiography}[{\includegraphics[width=1in,height=1.25in,clip,keepaspectratio]{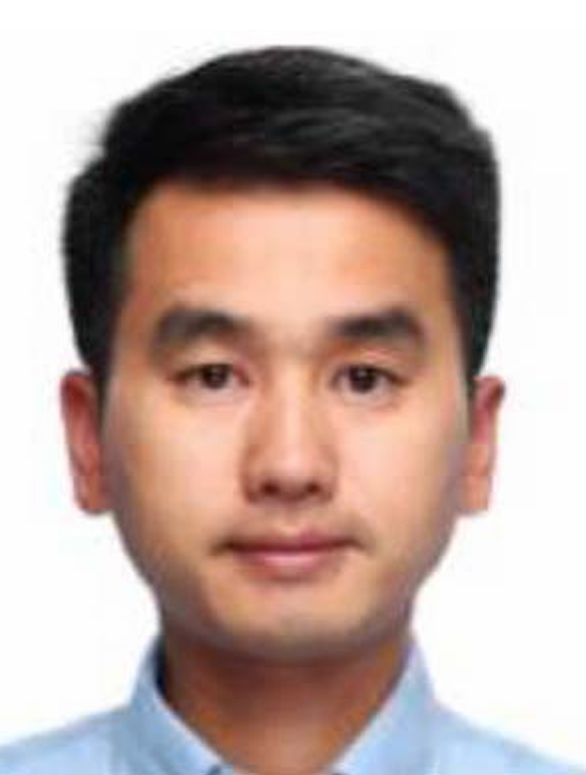}}]
  {Wenzhong Li} (Member, IEEE) receives the BS and PhD degree in computer science from Nanjing University, China. He was an Alexander von Humboldt Scholar fellow with the University of Goettingen, Germany. He is now a professor with the Department of Computer Science, Nanjing University. His research interests include distributed computing, Big Data mining and social networks. He has published more than 150 peer-review papers at international conferences and journals, which include INFOCOM, UBICOMP, AAAI, IJCAI, ACM Multimedia, CVPR, IEEE Communications Magazine, IEEE/ACM ToN, IEEE JSAC, IEEE TKDE, IEEE TPDS, etc. He served as program co-chair of MobiArch 2013 and Registration chair of ICNP 2013. He was the TPC member of several international conferences and the reviewer of many journals. He is the principle investigator of four fundings from NSFC, and the co-principle investigator of a China-Europe international research staff exchange program. He is a member of ACM, and China Computer Federation (CCF). He was featured on Elsevier’s Most Cited Chinese Researchers in 2022-2023. 
\end{IEEEbiography}
\vspace{-1.0cm}
\begin{IEEEbiography}[{\includegraphics[width=1in,height=1.25in,clip,keepaspectratio]{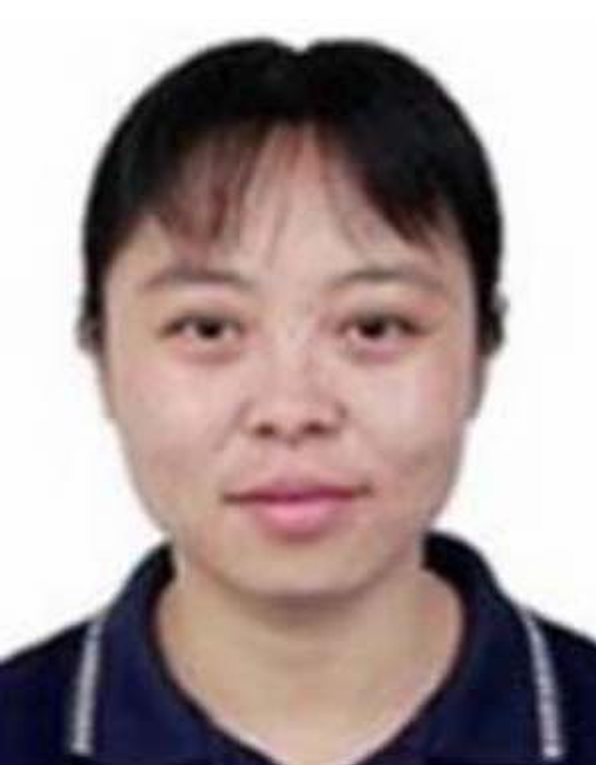}}]
  {Sanglu Lu} (Member, IEEE) received the BS, MS, and PhD degrees in computer science from Nanjing University in 1992, 1995, and 1997, respectively. She is currently a professor with the Department of Computer Science and Technology and the deputy Director of State Key Laboratory for Novel Software Technology.Her research interests include distributed computing, pervasive computing, and wireless networks. She has published more than 100 papers in referred journals and conferences in the above areas. She is a member of ACM.
\end{IEEEbiography}
\vspace{-1.0cm}
%%%%%%%%%%%%%%%%%%%%%%%%%%%%%%%%%%%%%%%%%%%%%%%%%%%%%%%%%%%%%%%%%%%%%%%%%%%%%%%%%

\clearpage
\section{Further Related Work} \label{section:further_relatedwork}
\subsection{Time Series Prediction Models}
Prior to the advent of deep learning, the ARIMA model~\cite{Zhang2003ARIMA}, founded on mathematical theory, was extensively employed in time series prediction tasks. Nevertheless, in the face of time series data possessing nonlinear trends or seasonality, ARIMA encounters difficulties in precisely apprehending the intrinsic characteristics~\cite{Kontopoulou2023ARO}. 
During the nascent stage of deep learning's development, models based on Recurrent Neural Networks (RNN)~\cite{Shi2015ConvolutionalLN,Aseeri2023EffectiveRF} were devised to capture the dependency patterns among time points within time series data.
Subsequently, the Temporal Convolutional Network (TCN)~\cite{wu2023timesnet,Luo2024ModernTCNAM} attracted extensive attention by virtue of its remarkable local modeling capabilities. 
Nonetheless, the Markovian-based RNN models are inherently flawed in their inability to efficaciously model long-term dependencies~\cite{zhou2021informer,wu2021autoformer}. Meanwhile, the one-dimensional TCN models, circumscribed by the limitations of their receptive fields, fail to capture global dependency~\cite{Tang2020OmniScaleCA}. 

During this phase, a profusion of research endeavors burgeoned, with the aim of augmenting the receptive fields of convolutional architectures. MICN~\cite{wang2023micn} conducts modeling from a local-global perspective, initially distilling the local traits of the observed sequence and subsequently derives the correlations among all local features to constitute the global characteristics. Meanwhile, TimesNet~\cite{wu2023timesnet} exploits the periodic pattern information to convert the original series into a two-dimensional temporal image and employs two-dimensional convolution to simultaneously model the time series patterns across multiple periods, serving as a supplementation to the global perspective.
Moreover, the Multilayer Perceptron (MLP)~\cite{Zeng2022DLinear,Xu2023FITSMT} founded on pointwise projection has also evinced substantial potential in the time series prediction tasks.
Subsequently, Approaches~\cite{zhou2021informer,wu2021autoformer,zhou2022fedformer,liu2021pyraformer} based on the attention mechanism, which directly learned long-term dependency relations and extract global information, have been extensively utilized in long-term prediction tasks. The remarkable methodologies PatchTST~\cite{Nie2023PatchTST} and iTransformer~\cite{liu2023itransformer} respectively introduced the embedding strategies of "Patch as Token" and "Series as Token" to ameliorate the traditional "Point as Token", and have presently emerged as paradigmatic works in the application of the attention mechanism within the time series. 

Certain recent investigations~\cite{Wang2024TimeMixerDM,Wang2025TimeMixerAG,Lin2024CycleNetET}, have discerned the inherent challenges of time series data, namely, the superposition and coupling of multiple fluctuation patterns, which present a formidable challenge for time series.

\subsection{Frequency Assisted Analysis}
Frequency domain enhancement strategies have been astutely implemented in \textbf{diverse foundational architectures} and have attracted extensive attention in the purview of time series analysis. 
Specifically, FEDformer~\cite{zhou2022fedformer} has proposed a mixture of expert’s decomposition module and a frequency enhancement mechanism to ameliorate the computational complexity of the attention mechanism. 
FreTS~\cite{yi2023FreTS} advocates the utilization of MLP to capture the complete perspective and global dependency of the observed sequence within the frequency domain.
FITS~\cite{Xu2023FITSMT} has developed low-pass filters and linear transformations in the complex frequency domain to directly model the mapping relationships between various frequency components of the observed sequence and the predicted sequence in the frequency domain. 
FilterNet~\cite{Yi2024FilterNetHF} introduces learnable frequency filters to extract pivotal temporal patterns, thereby facilitating the efficient handling of high-frequency noise and the utilization of the entire frequency spectrum for prediction. 
FCVAE~\cite{Wang2024FCVAE} introduces frequency domain information as guiding information and utilizes both global and local frequency domain representations to capture heterogeneous periodic patterns and detailed trend patterns, aiming to enhance the accuracy of anomaly detection algorithms. 
ROSE~\cite{Wang2024ROSERA} separates the coupled semantic information in time series data through multiple frequency domain masks and reconstructions, subsequently extracting unified representations across diverse domains.

As expounded above, given that decoupling the mixed fluctuation patterns directly within the temporal domain poses challenges. At the same time, spectrum analysis, by virtue of its inherent advantages of possessing a global perspective and energy compression, has been extensively deployed in prevalent methodologies to fortify the model's capacity to identify and generalize intricate fluctuation patterns from observed sequences. It is natural to put forward a conjecture:

\begin{tcolorbox}[notitle, rounded corners, colframe=gray, colback=white, boxrule=2pt, boxsep=0pt, left=0.15cm, right=0.17cm, enhanced, toprule=2pt, before skip=0.65em, after skip=0.75em]
\emph{{\centering 
  {\fontsize{8.5pt}{13.2pt}\selectfont 
  Would it be effective to transform the observed sequence into the frequency spectrum and subsequently disentangle disparate fluctuation patterns by virtue of distinct frequency magnitudes? 
  }\\
}}
\end{tcolorbox}

\subsection{Multi-scale and Multi-period Decomposition}
Beyond the research route focusing on modeling architectures, the enhancement strategies based on trend-season and multi-scale decomposition have also demonstrated remarkable potential.

In the context of multi-scale decomposition algorithms, Pyraformer~\cite{liu2021pyraformer} incorporates a pyramid attention module, within which the scale architecture simultaneously aggregates features at disparate resolutions and time correlations across diverse scopes. 
Subsequently, SCINet~\cite{Liu2021SCINetTS} designs a binary tree structure to extract time series dependencies at various scales from the observed sequence and employs additional channels to compensate for the long-term dependency information deficient in TCN. 
The recent TimeMixer~\cite{Wang2024TimeMixerDM} proposes a multi-scale mixed architecture to disentangle the complex time series patterns in time series prediction and utilizes multi-scale complementarity to enhance the prediction capabilities.

With regard to multi-period decomposition, recent models incorporating multi-periodic designs, such as MSGNet~\cite{Cai2023MSGNetLM}, which employs frequency domain analysis and adaptive graph convolution to model the correlations among disparate sequences across multiple time scales, thereby capturing prominent periodic patterns. 
Moreover, MTST~\cite{Zhang2023MTST} concurrently models variegated time patterns at diverse resolutions and utilizes patches of multiple scales to identify different periodic patterns. 

Conventional periodic decomposition and multi-scale augmentation algorithms fall short of effectively disentangling disparate fluctuation patterns, as the high-frequency pattern exists only in the seasonal and fine-grained segments and remains superposed and intertwined with other patterns. These approaches mainly focus on prediction and lack the capacity to capture universal temporal patterns applicable to diverse tasks. 

Moreover, time series originating from disparate domains typically exhibit markedly heterogeneous compound periodic patterns, which precludes the commonplace multi-period decomposition from accurately partitioning the various periodic patterns. For instance, traffic data is constituted by the superposition of periodic fluctuations at the daily, monthly, and yearly scales, whereas medical physiological signals are more concerned with fine-grained periodic patterns such as those at the second, minute, and hour scales. Even in the presence of ample prior information, the endeavor to model diverse periodic-fluctuation patterns via shared channels remains arduous~\cite{Lin2024CycleNetET}.

\subsection{Enhancement Strategy}
Enhancement strategies represent a potent modality for eliciting the complete potential of deep learning within complex datasets and have witnessed extensive application across vision and natural language.
VitComer~\cite{Xia2024ViTCoMerVT} has devised a feature enhancement architecture that is plain, devoid of pre-training, and encompasses a bidirectional fusion and interaction module of CNN-Transformer. This architecture facilitates multi-scale amalgamation of cross-level features, thereby mitigating the problem of restricted local information interchange. VitAdapter~\cite{chen2022vitadapter} introduces inductive biases via a supplementary augmentation architecture to compensate for the prior information regarding images that ViT lacks. This paper further investigates the enhancement of fluctuation patterns that are applicable across diverse tasks within time series.

Distinct from prior designs, Pets adaptively converts time series data into fluctuation patterns possessing a fixed energy distribution through amplitude margin filtering and performs adaptive aggregation of the mixed fluctuation patterns in accordance with the energy ranking order, guaranteeing that the fluctuation patterns with high task relevance can dominate the output results of the model.

\section{Implementation Details}\label{sec:detail}

\paragraph{Datasets Details.}
We evaluate the performance of different models for long-term forecasting on 10 well-established datasets, including Weather, Traffic, Electricity, Exchange, Solar, ILI, and ETT datasets (ETTh1, ETTh2, ETTm1, ETTm2). Furthermore, we adopt PeMS and M4 datasets for short-term forecasting. We detail the descriptions of the dataset in Table \ref{tab:dataset}.To comprehensively evaluate the model's performance in time series analysis tasks, we further introduced datasets for classification and anomaly detection. The classification task is designed to test the model's ability to capture coarse-grained patterns in time series data, while anomaly detection focuses on the recognition of fine-grained patterns. Specifically, we used 10 multivariate datasets from the UEA Time Series Classification Archive (2018) for the evaluation of classification tasks. For anomaly detection, we selected datasets such as SMD (2019), SWaT (2016), PSM (2021), MSL, and SMAP (2018). We detail the descriptions of the datasets for classification and anomaly detection in Table \ref{tab:uea_datasets} and Table \ref{tab:anomaly_detection_datasets}
\begin{table*}[thbp]
  % \vspace{-10pt}
  \caption{Dataset detailed descriptions. The dataset size is organized in (Train, Validation, Test).}\label{tab:dataset}
  \vskip 0.1in
  \centering
   \resizebox{2.0\columnwidth}{!}{
  \begin{threeparttable}
  \begin{small}
  \renewcommand{\multirowsetup}{\centering}
  \setlength{\tabcolsep}{3.8pt}
  \begin{tabular}{c|l|c|c|c|c|c|c}
    \toprule
    Tasks & Dataset & Dim & Series Length & Dataset Size &Frequency &Forecastability$\ast$ & {Information} \\
    \toprule
     & ETTm1 & 7 &  {\{96, 192, 336, 720\}} & (34465, 11521, 11521)  & 15min &0.46 & {Temperature}\\
    \cmidrule{2-8}
    & ETTm2 & 7 &  {\{96, 192, 336, 720\}} & (34465, 11521, 11521)  & 15min &0.55 & {Temperature}\\
    \cmidrule{2-8}
     & ETTh1 & 7 &  {\{96, 192, 336, 720\}} & (8545, 2881, 2881) & 15 min &0.38 & {Temperature} \\
    \cmidrule{2-8}
     &ETTh2 & 7 &  {\{96, 192, 336, 720\}} & (8545, 2881, 2881) & 15 min &0.45 & {Temperature} \\
    \cmidrule{2-8}
    Long-term & Electricity & 321 &  {\{96, 192, 336, 720\}} & (18317, 2633, 5261) & Hourly &0.77 &  {Electricity} \\
    \cmidrule{2-8}
    Forecasting & Traffic & 862 &  {\{96, 192, 336, 720\}} & (12185, 1757, 3509) & Hourly &0.68 &  {Transportation} \\
     \cmidrule{2-8}
     & Exchange & 8 &  {\{96, 192, 336, 720\}} & (5120, 665, 1422)  &Daily &0.41  & {Weather} \\
    \cmidrule{2-8}
     & Weather & 21 &  {\{96, 192, 336, 720\}} & (36792, 5271, 10540)  &10 min &0.75  & {Weather} \\
    \cmidrule{2-8}
    & Solar-Energy & 137  &  {\{96, 192, 336, 720\}}  & (36601, 5161, 10417)& 10min &0.33 &  {Electricity} \\
    \midrule
    & PEMS03 & 358 & 12 & (15617,5135,5135) & 5min &0.65 &  {Transportation}\\
    \cmidrule{2-8}
    & PEMS04 & 307 & 12 & (10172,3375,3375) & 5min &0.45 &  {Transportation}\\
    \cmidrule{2-8}
    & PEMS07 & 883 & 12 & (16911,5622,5622) & 5min &0.58 &  {Transportation}\\
    \cmidrule{2-8}
    Short-term & PEMS08 & 170 & 12 & (10690,3548,265) & 5min &0.52 &  {Transportation}\\
    \cmidrule{2-8}
    Forecasting & M4-Yearly & 1 & 6 & (23000, 0, 23000) &Yearly &0.43 & {Demographic} \\
    \cmidrule{2-8}
     & M4-Quarterly & 1 & 8 & (24000, 0, 24000) &Quarterly &0.47 &  {Finance} \\
    \cmidrule{2-8}
    & M4-Monthly & 1 & 18 & (48000, 0, 48000) & Monthly &0.44 &  {Industry} \\
    \cmidrule{2-8}
    & M4-Weakly & 1 & 13 & (359, 0, 359) & Weakly &0.43 &  {Macro} \\
    \cmidrule{2-8}
     & M4-Daily & 1 & 14 & (4227, 0, 4227) &Daily &0.44 &  {Micro} \\
    \cmidrule{2-8}
     & M4-Hourly & 1 &48 & (414, 0, 414) & Hourly &0.46 &  {Other} \\
    \bottomrule
    \end{tabular}
     \begin{tablenotes}
        \item $\ast$ The forecastability is calculated by one minus the entropy of Fourier decomposition of time series \cite{Goerg2013ForecastableCA}. A larger value indicates better predictability.
    \end{tablenotes}
    \end{small}
  \end{threeparttable}
  }
  % \vspace{-5pt}
\end{table*}

\begin{table*}[thbp]
    \centering
    \caption{Datasets and mapping details of UEA dataset \cite{Bagnall2018TheUM}.}
      \vskip 0.1in
    \label{tab:uea_datasets}
    \begin{tabular}{l|cc|c}
        \hline
        Dataset & Sample Numbers(train set,test set) & Variable Number & Series Length \\
        \hline
        EthanolConcentration & (261, 263) & 3 & 1751 \\
        FaceDetection & (5890, 3524) & 144 & 62 \\
        Handwriting & (150, 850) & 3 & 152 \\
        Heartbeat & (204, 205) & 61 & 405 \\
        JapaneseVowels & (270, 370) & 12 & 29 \\
        PEMSSF & (267, 173) & 963 & 144 \\
        SelfRegulationSCP1 & (268, 293) & 6 & 896 \\
        SelfRegulationSCP2 & (200, 180) & 7 & 1152 \\
        SpokenArabicDigits & (6599, 2199) & 13 & 93 \\
        UWaveGestureLibrary & (120, 320) & 3 & 315 \\
        \hline
    \end{tabular}
\end{table*}

\begin{table*}[thbp]
    \centering
    \caption{Datasets and mapping details of anomaly detection dataset.}
    \vskip 0.05in
    \label{tab:anomaly_detection_datasets}
    \begin{tabular}{l|ccc}
        \hline
        Dataset & Dataset sizes(train set,val set, test set) & Variable Number & Sliding Window Length \\
        \hline
        SMD & (566724, 141681, 708420) & 38 & 100 \\
        MSL & (44653, 11664, 73729) & 55 & 100 \\
        SMAP & (108146, 27037, 427617) & 25 & 100 \\
        SWaT & (396000, 99000, 449919) & 51 & 100 \\
        PSM & (105984, 26497, 87841) & 25 & 100 \\
        \hline
    \end{tabular}
\end{table*}

\paragraph{Baseline Details.} 
To assess the effectiveness of our method across various tasks, we select 27 advanced baseline models spanning a wide range of architectures. Specifically, we utilize CNN-based models: MICN~\cite{wang2023micn}, SCINet~\cite{Liu2021SCINetTS}, and TimesNet~\cite{wu2023timesnet}; MLP-based models: TimeMixer~\cite{Wang2024TimeMixerDM}, LightTS~\cite{Campos2023LightTSLT}, and DLinear~\cite{Zeng2022DLinear}; 
RMLP\&RLinear~\cite{Li2023RevisitingLT}
and Transformer-based models: iTransformer~\cite{liu2023itransformer}, PatchTST~\cite{Nie2023PatchTST}, Crossformer~\cite{Du2021CrossDomainGD}, FEDformer~\cite{zhou2022fedformer}, Stationary~\cite{Liu2022NonstationaryTR}, Autoformer~\cite{wu2021autoformer}, and Informer~\cite{zhou2021informer}. 
These models have demonstrated superior capabilities in temporal modeling and provide a robust framework for comparative analysis.
For specific tasks, TiDE~\cite{Das2023LongtermTiDE}, FiLM~\cite{zhou2022film}, N-HiTS~\cite{Challu2022NHiTSNH}, and N-BEATS~\cite{Oreshkin2019nbeats} address long- or short-term forecasting; 
Anomaly Transformer~\cite{xu2021anomaly} and MTS-Mixers~\cite{Li2023MTSMixersMT} target anomaly detection; 
while Rocket~\cite{Li2023RevisitingLT}, LSTNet~\cite{Lai2017ModelingLA}, LSSL~\cite{Gu2021EfficientlyML}, and Flowformer~\cite{Wu2022FlowformerLT} are utilized for classification. 
Few/zero-shot forecasting tasks employ ETSformer~\cite{woo2022etsformer}, Reformer~\cite{kitaev2020reformer}, and LLMTime~\cite{llmtime}.

\paragraph{Metric Details.}
Regarding metrics, we utilize the mean square error (MSE) and mean absolute error (MAE) for long-term forecasting. In the case of short-term forecasting, we follow the metrics of SCINet \cite{Liu2021SCINetTS} on the PeMS datasets, including mean absolute error (MAE), mean absolute percentage error (MAPE), root mean squared error (RMSE). As for the M4 datasets, we follow the methodology of N-BEATS \cite{Oreshkin2019nbeats} and implement the symmetric mean absolute percentage error (SMAPE), mean absolute scaled error (MASE), and overall weighted average (OWA) as metrics. It is worth noting that OWA is a specific metric utilized in the M4 competition. The calculations of these metrics are:
\begin{small}
\begin{align*} \label{equ:metrics}
    \text{RMSE} &= (\sum_{i=1}^F (\mathbf{X}_{i} - \widehat{\mathbf{X}}_{i})^2)^{\frac{1}{2}}, \text{MAE} = \sum_{i=1}^F|\mathbf{X}_{i} - \widehat{\mathbf{X}}_{i}|,\\
    \text{SMAPE} &= \frac{200}{F} \sum_{i=1}^F \frac{|\mathbf{X}_{i} - \widehat{\mathbf{X}}_{i}|}{|\mathbf{X}_{i}| + |\widehat{\mathbf{X}}_{i}|}, \text{MAPE} = \frac{100}{F} \sum_{i=1}^F \frac{|\mathbf{X}_{i} - \widehat{\mathbf{X}}_{i}|}{|\mathbf{X}_{i}|}, \\
    \text{MASE} &= \frac{1}{F} \sum_{i=1}^F \frac{|\mathbf{X}_{i} - \widehat{\mathbf{X}}_{i}|}{\frac{1}{F-s}\sum_{j=s+1}^{F}|\mathbf{X}_j - \mathbf{X}_{j-s}|}, \\
    \text{OWA} &= \frac{1}{2} \left[ \frac{\text{SMAPE}}{\text{SMAPE}_{\textrm{Naïve2}}}  + \frac{\text{MASE}}{\text{MASE}_{\textrm{Naïve2}}}  \right],
\end{align*}
\end{small}
where $s$ is the periodicity of the data. $\mathbf{X},\widehat{\mathbf{X}}\in\mathbb{R}^{F\times C}$ are the ground truth and prediction results of the future with $F$ time pints and $C$ dimensions. $\mathbf{X}_{i}$ means the $i$-th future time point.

\paragraph{Experiment Details.}
To valudate the reproducibility of all experimental results, each experiment was executed three times and the average results were presented. The code was implemented on the basis of PyTorch \cite{Paszke2019PyTorchAI} and executed on multiple NVIDIA V100 40GB GPUs. The initial learning rate was configured as either $10^{-3}$ or $10^{-4}$, and the model was optimized by means of the ADAM optimizer \cite{Kingma2014AdamAM} with L2 loss. The batch size was designated as 128. By default, \myformer encompasses all 4 layers, with each layer incorporating PPA (adapter block), MPR, MPM, and backbone block.  Moreover, each dataset inherently supports 3 fluctuation patterns and an additional trend component. To ensure the impartiality and efficacy of the experimental results, we replicated all the experimental configurations of TimeMixer~\cite{Wang2024TimeMixerDM} and employed the experimental results of certain baselines presented in its original publication.

\section{Further Discussion for \myformer}
\subsection{Advantages of \myformer over multi-resolution-based models for diverse tasks} 

In the existing multi-resolution approaches, TimeMixer generates multi-scale sequences via downsampling only in the time domain and establishes connections between different-scale representations using linear layers. N-BEATS and N-HITS imposes constraints on expansion coefficients as prior information to guide layers to learn specific time-series features (e.g., seasonality), enabling interpretable sequence decomposition. However, these methods inherently rely on traditional seasonal-trend decomposition and multi-scale decomposition, which fail to achieve complete decoupling—for example, composite high-frequency components remain coupled in seasonal or fine-grained parts.

In contrast, our proposed method introduces a novel frequency-domain-based decoupling algorithm (SDAQ), which transforms input sequences into the frequency domain and decouples patterns explicitly in the frequency domain, allowing for more precise reconstruction of multi-period fluctuations. This ensures lossless and independent decomposition, resolving the coupling of high-frequency components often present in traditional time-domain approaches.

Concretely, SDAQ transforms input sequences into the frequency domain and explicitly separates them into three frequency components (bands), ensuring lossless and non-overlapping distribution of information from observed sequences into three decoupled sequences. Then SDAQ inverse-transforms each group back to the time domain, thus resolving the information coupling between different frequency bands.

\subsection{The Distinction between zero-shot transfer and zeros-shot forecasting} 
In Section~\ref{sec:zero_shot}, we presented the detailed results of the proposed approach under the zero-shot scenario. However, unlike existing time series foundation models that directly conduct inference on the target data domain after pre-training. In our designed experiments, \myformer and some recent small-scale models (including PatchTST~\cite{Nie2023PatchTST}, TimesNet~\cite{wu2023timesnet}, and TimeMixer~\cite{Wang2024TimeMixerDM}) are first trained from scratch in source data domain and then predict in the unseen target data domain. 

To avoid ambiguity, we refer to these two forecasting paradigms as \textbf{Zero-shot Forecasting} and \textbf{Zero-shot Transfer}. In the past, zero-shot and few-shot tasks were only regarded as benchmarks for evaluating large-scale models (such as large language models and time series foundation models). However, some recent works~\cite{Zhou2023OneFA,Jin2023TimeLLMTS,llmtime,Wang2025TimeMixerAG} commonly choose zero-shot and few-shot forecasting as one of the tasks for evaluating model performance. These approaches believe that cross-domain generalization and robustness are significant capabilities of purpose general model, which are crucial for addressing out-of-distribution (OOD) issues and establishing more powerful general time series models.

\subsection{The selection of the mother wavelet in the continuous wavelet transform (CWT)} 
In the open source code, we choose the Haar function as the mother wavelet function. As a compactly supported orthogonal wavelet function, Haar wavelets only consider local regions during decomposition, leading to high computational efficiency that helps reduce Pets’ overall complexity. 
Additionally, Haar wavelets excel in decomposing high-frequency components, enhancing \myformer’ generalization capability for handling diverse fluctuation patterns across frequencies. In \myformer, we selected Haar wavelets due to their extensive application in time-series analysis. Further exploring the potential advantages of other wavelet functions for decoupling composite fluctuation patterns will be part of our future work.

\section{Details of Model design}
\subsection{Fluctuation Pattern Assisted}
In this section, we present the detailed architecture of the proposed \myformer, encompassing Periodic Prompt Adapter (\textbf{PPA}), Multi-fluctuation Patterns Rendering (\textbf{MPR}), Multi-fluctuation Patterns Mixing (\textbf{MPM}) and mixture of predictors (\textbf{MoP}), as shown in Fig.~\ref{fig:method_2}. Specifically, within the adapter block, the input tokens of diverse fluctuation patterns are passed through successive linear, conv1d, and activation layers, after which they are concatenated to obtain tokens of a comprehensive pattern. These tokens then pass through a self-attention mechanism and a feed-forward structure connected by a residual connection, yielding a deep representation. Finally, this deep representation is processed by three independent conv1d layers to generate a single fluctuation pattern for the subsequent stage.

In the MPR, we devised a novel variable zero-convolution gate. This gate is designed to make MPR at different stages inclined to learn information from designated pattern tokens. For instance, in the MPR of the first layer, the zero-convolution layer is inserted into the short-wave and high-frequency pattern channels. This makes the current MPR tend to learn temporal representations from long-wave patterns. Meanwhile, all the weights of the zero-convolution layer are fixed at zero only during initialization and are permitted to participate in subsequent training. This ensures that the convolution gate has a distinct bias in the initial training phase, while in subsequent training, each layer of MPR can flexibly capture comprehensive temporal representations from all pattern tokens.

The MPM simultaneously receives two inputs: the hidden representation and a set of fluctuation patterns. The hidden representation contains the information of fluctuation patterns already learned by the model, which serves as conditional information to guide the model in capturing more profound fluctuation patterns. Additionally, we present a novel hybrid predictor capable of gradually generating the fluctuation details of future sequences.

\subsection{Multi-channel Analysis}
In real world scenarios of time series analysis, the vast majority of datasets consist of multiple time series, and there are evident dependencies among these channels. For instance, modern industrial equipment often incorporates multiple groups of sensors. Each group of sensors focuses on collecting signals from different parts. Meanwhile, for sensitive segments, sensors for different physical signals may be configured, such as photosensitive, current, and temperature sensors. Additionally, in medical diagnosis scenario, the condition of patient is jointly determined by multiple physiological signals, such as heart rate, blood pressure, and the surface temperature at the lesion site.

These complex and variable realm scenarios demand that a model must capture comprehensive temporal fluctuation information from all channels to make accurate predictions and judgments about samples. To address these challenges, supporting multi-channel data input has naturally become a latent requirement for general models. In this section, we elaborate on how the model models the inter-channel dependencies.

\begin{figure*}[t]
\begin{center}
\centerline{\includegraphics[width=2.0\columnwidth]{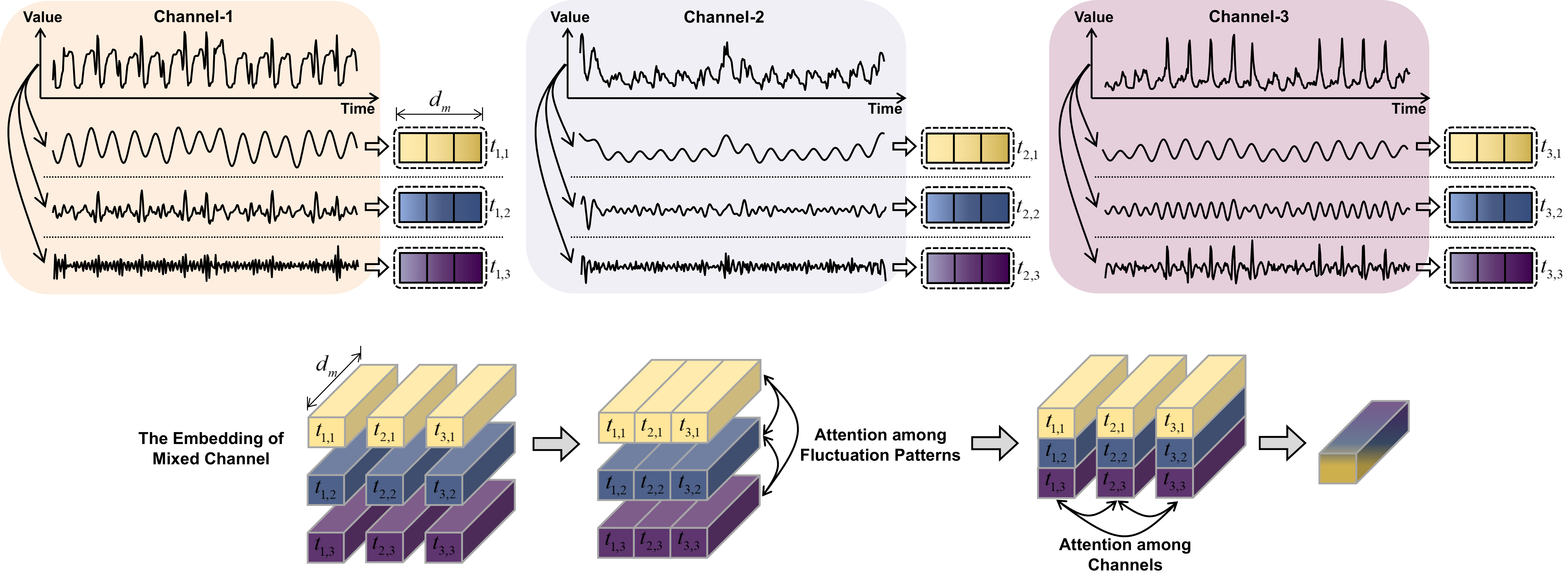}}
\vspace{-10pt}
\caption{
We present the design details of the proposed \myformer under the channel-mixing design. It incorporates two key configurations, the attention mechanism among fluctuation patterns and the attention mechanism among channels.
}\label{fig:method_3}
\end{center}
\end{figure*}

As depicted in Fig.~\ref{fig:method_3}, time series from different channels undergo independent decoupling and patch-based embedding operations, resulting in an identical set of long- and short-wave pattern tokens. These tokens are re-concatenated into two groups of representations according to the principles of the same fluctuation pattern and the same channel. Subsequently, we first calculate the attention scores among the fluctuation patterns and then among the channels.

This flexible computational strategy for the attention mechanism ensures that, for each token, both other fluctuation patterns from the same channel and the same fluctuation pattern from different channels are visible. Additionally, the spectrum-quantization decoupling operation guarantees that, compared with the coupled and superimposed observation sequences, each decoupled fluctuation pattern exhibits more distinct periodic characteristics. Meanwhile, the same fluctuation patterns from different channels also demonstrate explicit convergence, which significantly alleviates the potential heterogeneity among different channels. These advantages ensure that the proposed \myformer can gain an edge on datasets with a large number of channels, as exemplified by the experimental results of PEMS presented in Table.~\ref{tab:short_pems}.

\subsection{Model Hyperparameter Sensitivity Analysis} \label{sec:hyperparameter}
In the design of SDAQ, the energy in time series is mainly concentrated in the low-frequency bands, and the energy of the bands continuously decays as the frequency increases. This implies that by dividing only a limited number of frequency bands, the input series can be decoupled into multiple fluctuation patterns without overlapping. 
Additionally, when the number of divided frequency bands is excessive, limited information falling within the high-frequency range will be further split. This leads to the decoupled series containing a large amount of noise and a small amount of information, thereby reducing the predictive performance of the model. Based on these considerations, we fix the number of decoupled fluctuation patterns as $K\textit{=}3$, and set the energy ratio as $(\mu_1,\mu_2)$. 

In the design of component FPA and MoP, the model consists of $N$ consecutive composite layers. Each layer contains single adapter block (PPA), single backbone block, single MPM, and single MPR. Correspondingly, the hybrid predictor is also composed of $N$ predictors with the same structure. Additionally, both the original sequence and the decoupled sequence yield $P_L$ tokens after the patch-based embedding operation, where the dimension of each token is $P_d$.

In order to verify the sensitivity of \myformer to hyperparameter selection, the Table~\ref{tab:exp_sensitivity} shows the performance of \myformer with different parameter scales on multiple benchmarks. Where the model parameter combinations include \begin{small}$(N,P_d) \textit{=} (3,16), (4,32), (6,128)$\end{small} and \begin{small}$(\mu_1,\mu_2) \textit{=} (0.3,0.6), (0.4,0.8), (0.7,0.9)$\end{small}, and a performance metric of MSE. By default, the hyperparameters of \myformer are fixed to $N\textit{=}4$, $P_d\textit{=}32$ and $(\mu_1,\mu_2)\textit{=}(0.7,0.9)$, all experimental results presented follow this setting.

\begin{table*}[htpb]
  \vspace{-10pt}
  \caption{To verify the sensitivity of \myformer to hyperparameter selection, we show the performance of \myformer with different parameter scales on multiple benchmarks, with a fixed forecasting window of 96, and a performance metric of MSE.}\label{tab:exp_sensitivity}
  \centering
  \resizebox{1.7\columnwidth}{!}{
  \begin{small}
  \renewcommand{\multirowsetup}{\centering}
  \tabcolsep=0.4cm
  \renewcommand\arraystretch{1.3}
  \begin{tabular}{ccccccccc}
    \toprule
    \hline
    
    \multicolumn{2}{c}{\multirow{1}{*}{\scalebox{1.0}{Hyperparameter}}} & 
    \multicolumn{1}{c}{\rotatebox{0}{\scalebox{1.0}{ETTh1}}} & 
    \multicolumn{1}{c}{\rotatebox{0}{\scalebox{1.0}{ETTh2}}} & 
    \multicolumn{1}{c}{\rotatebox{0}{\scalebox{1.0}{ETTm1}}} & 
    \multicolumn{1}{c}{\rotatebox{0}{\scalebox{1.0}{ETTm2}}} & 
    \multicolumn{1}{c}{\rotatebox{0}{\scalebox{1.0}{Weather}}} & 
    \multicolumn{1}{c}{\rotatebox{0}{\scalebox{1.0}{ECL}}} & 
    \multicolumn{1}{c}{\rotatebox{0}{\scalebox{1.0}{Traffic}}} \\
    \hline

    \multicolumn{1}{c}{\multirow{3}{*}{\rotatebox{0}{$\mu_1,\mu_2$=0.3,0.6}}} &
    \multirow{1}{*}{\scalebox{1.0}{\shortstack{L=3,D=16}}} &
    0.398 & 0.327 & 0.304 & 0.170 & 0.155 & 0.139 & 0.394 \\
    \cline{2-9}

    & \multirow{1}{*}{\scalebox{1.0}{\shortstack{L=4,D=32}}} &
    0.389 & 0.322 & 0.289 & 0.177 & 0.152 & 0.144 & 0.412 \\
    \cline{2-9}
    
    & \multirow{1}{*}{\scalebox{1.0}{\shortstack{L=6,D=128}}} &
    0.395 & 0.331 & 0.295 & 0.173 & 0.159 & 0.148 & 0.415 \\
    \hline

    \multicolumn{1}{c}{\multirow{3}{*}{\rotatebox{0}{$\mu_1,\mu_2$=0.4,0.8}}} &
    \multirow{1}{*}{\scalebox{1.0}{\shortstack{L=3,D=16}}} &
    0.371 & \textbf{0.292} & 0.291 & 0.167 & 0.146 & 0.140 & \textbf{0.375} \\
    \cline{2-9}
    
    & \multirow{1}{*}{\scalebox{1.0}{\shortstack{L=4,D=32}}} &
    0.378 & 0.304 & 0.279 & \textbf{0.163} & 0.145 & 0.135 & 0.381 \\
    \cline{2-9}

    & \multirow{1}{*}{\scalebox{1.0}{\shortstack{L=6,D=128}}} &
    0.383 & 0.315 & 0.292 & 0.170 & 0.152 & 0.141 & 0.392 \\
    \hline
    
    \multicolumn{1}{c}{\multirow{3}{*}{\rotatebox{0}{$\mu_1,\mu_2$=0.7,0.9}}} &
    \multirow{1}{*}{\scalebox{1.0}{\shortstack{L=3,D=16}}} &
    \textbf{0.364} & 0.298 & \textbf{0.274} & 0.164 & \textbf{0.142} & \textbf{0.132} & 0.378 \\
    \cline{2-9}
    
    & \multirow{1}{*}{\scalebox{1.0}{\shortstack{L=4,D=32}}} &
    0.385 & 0.334 & 0.289 & 0.164 & 0.154 & 0.145 & 0.388 \\
    \cline{2-9}

    & \multirow{1}{*}{\scalebox{1.0}{\shortstack{L=6,D=128}}} &
    0.389 & 0.331 & 0.282 & 0.173 & 0.159 & 0.143 & 0.395 \\
    \hline
    
    \bottomrule
  \end{tabular}
  \end{small}
}
\end{table*}

\subsection{Comprehensive complexity analysis} \label{sec:complexity_analysis}
In this section, we provide a detailed analysis of the computational complexity of Pets’ three main components (SDAQ, FPA, and MoP) and address the specific concerns regarding wavelet transform, spectral quantization, and multi-stage predictors. Pets consists of three main components: SDAQ, FPA, and MoP. 

The computational complexity of SDAQ involves three steps: (1) transforming an input sequence of length $L$ into a time-frequency spectrogram with $\lambda$ scales using an FFT-optimized Continuous Wavelet Transform (CWT), which reduces convolution operations to $O(\lambda\cdot L\log L)$; (2) partitioning the spectrogram into three subgraphs based on energy distribution, an operation with negligible complexity; and (3) reconstructing each subgraph into the time domain via inverse wavelet transform (iWT), with the same $O(\lambda\cdot L\log L)$ complexity. Together, SDAQ operates with an overall complexity $O(\lambda\cdot L\log L)$. 

The FPA and MoP modules consist of $N$ composite layers, each containing lightweight operations such as 1D convolutions, self-attention, and linear layers. After patching, the tensor shape is $Bd\times P_L\times P_d$, where $P_L$ is the number of tokens and $P_d$ is the token dimension. 
Therefore, linear layer and one-dimensional convolution computation complexity is $O\left(\left(P_L\right)^2+\left(P_d\right)^2\right)$, and the computational complexity of attention is $O(P_{L}(P_{d})^{2})$. Note that $P_{L}P_{d}=L$, So there is $O\left(\left(P_L\right)^2+\left(P_d\right)^2\right)\geq O\left(2P_LP_d\right)=O\left(L\right)$ and $O\left(P_L\left(P_d\right)^2\right)=O\left(LP_d\right)$. These modules have an overall complexity of $O(L+LP_d)$, ensuring efficient processing while capturing complex patterns. Regarding the specific processes mentioned:

\begin{itemize}
    \item Wavelet Transform: Used in SDAQ, the wavelet transform (CWT and iWT) operates with a combined complexity of $O(\lambda\cdot L\log L)$, introducing computational overhead but maintaining scalability through FFT optimization.
    \item Spectral Quantization: This step partitions the spectrogram into three subgraphs based on frequency energy distribution. The complexity is negligible as it involves simple calculations and indexing.
    \item Multi-Stage Predictors: Incorporated in FPA and MoP, these predictors rely on 1D convolutions, self-attention, and linear layers, with complexity $O(L+LP_d)$. This design ensures a balance between computational cost and the ability to model diverse fluctuation patterns effectively.
\end{itemize}

% \input{tables/computational_cost.tex}
% Furthermore, to further demonstrate Pets’ real-world inference overhead, we present inference speed (ms/iter) and memory usage (GB) during training on the ETTm1 dataset for a forecast horizon of 720 in the Table~\ref{tab:computation}.

\subsection{Algorithm for training of \myformer} \label{sec:algorithm}
We provide the training procedure of \myformer in Algorithm~\ref{alg:training}, which are described in Section~\ref{section:pets}. 

Where $p_\textit{emb}$ and $\left(\overline{h}_m,\overline{h}_c,\overline{h}_b,\overline{h}_a\right)$ together serve as the condition variable $c$, 
% $z_\textit{emb}$ is an all-zero tensor sharing identical shape with $p_\textit{emb}$, and $\left(\overline{z}_m,\overline{z}_c,\overline{z}_b,\overline{z}_a\right)$ follows the same design. Under this setup, 
$\tilde{Y}_{1:H}^{t-1}\in\mathbb{R}^{Bd\times P_H\times P_d}$ and $\hat{Y}_{1:H}^{t-1}\in\mathbb{R}^{Bd\times P_H\times P_d}$ represent the conditional and unconditional outputs of the denoising network, respectively. Furthermore, $Y_{1:H}^{t-1}$ as the final result, and $\lambda$ is a parameter that controls the proportion of conditional and unconditional generation in the final result ($\lambda \textit{=7.5}$ for our experiment)

\begin{algorithm*}[htbp]
   \caption{Training of \myformer}
   \label{alg:training}
\begin{algorithmic}[1]
   \STATE {\bfseries Input:} The observation series $X_{-L+1:0}\in\mathbb{R}^{B\times d\times L}$ and prediction series $X_{1:H}\in\mathbb{R}^{B\times d\times H}$, the hyperparameters: the model depth $N$, the feature dimension $P_d$, the spectral width $\lambda$, along with the energy thresholds $(\mu_1,\mu_2)$ and the iteration $N_{\mathrm{iter}}$.
   \STATE {\bfseries Output:} Optimized \textit{Pets} consisting of $N \times$ FPAs and MoP.
   \FOR{ $i=1$ {\bfseries to} $N_{\mathrm{iter}}$}
   \STATE Channel Independent: \\
          \begin{small}{$X_{-L+1:0}\in\mathbb{R}^{B\times d\times L}\to X_{-L+1:0}\in\mathbb{R}^{Bd\times L}$}\end{small};\\
          Temporal-Spectral Decomposition by \begin{small}$CWT_{\lambda}$\end{small}: \\
          \begin{small}{$X_{-L+1:0}\in\mathbb{R}^{Bd\times L}\to A\in\mathbb{R}^{Bd\times L\times\lambda}$}\end{small}; \\
          Amplitude Quantization Phase: \\
          \begin{small}{$A\in\mathbb{R}^{Bd\times L\times\lambda}\to A^{1:K}\in\mathbb{R}^{Bd\times L\times\lambda}\to X_{-L+1:0}^{1:K}\textit{=}iWT(A^{1:K})\in\mathbb{R}^{Bd\times L}$}\end{small};\\
          Patching Instance and Embedding: \\
          \begin{small}{$X_{-L+1:0}\in\mathbb{R}^{Bd\times L}\to E^0\in\mathbb{R}^{Bd\times P_L\times P_d}$}\end{small}, \\
          \begin{small}$\{X_{-L+1:0}^{k}\in\mathbb{R}^{Bd\times L}\}_{k\in [1,K]}\to \{E_k^0\in\mathbb{R}^{Bd\times P_L\times P_d}\}_{k\in [1,K]}$\end{small}; \\
          % Patch Embedding: \\
          % \begin{small}{$$}\end{small};\\
          % Patch Embedding: \\
          % \begin{small}{$$}\end{small};\\
          % Patch Embedding: \\
          % \begin{small}{$$}\end{small};\\
          \FOR{ $n=1$ {\bfseries to} $N$}
                \STATE In the Periodic Prompt Adapter block: \\
                \begin{small}$E_k^{n-1}\textit{=}Conv1D(Linear(E_k^{n-1})^\top)^\top, E_k^{n-1}\textit{=}Dropout(Linear(Act(E_k^{n-1})))$\end{small}; \\
                \begin{small}$E_{concat}^{n-1}\textit{=}Concatenate(\{E_k^{n-1}\}_{k\in [1,K]})$\end{small}; \\
                \begin{small}$E_{concat}^{n-1}\textit{+=}SelfAttn(E_{concat}^{n-1}), E_{concat}^{n-1}\textit{+=}Conv1D(E_{concat}^{n-1})$\end{small}; \\
                \begin{small}$E_{k}^{n}\textit{=}Conv1D_{k}(E_{concat}^{n-1})_{k\in [1,K]}$\end{small}; \\
                \STATE In the Multi-fluctuation Patterns Rendering block: \\
                \begin{small}$\left\{E_k^n\right\}_{k\in[1,n-1]\cup[n+1,K]}=ZeroConv1d\left(\left\{E_k^n\right\}_{k\in[1,n-1]\cup[n+1,K]}\right)$\end{small}; \\
                \begin{small}$E_{concat}^{n}\textit{=}Concatenate(\{E_k^{n}\}_{k\in [1,K]})$\end{small}; \\
                \begin{small}$E_{concat}^n\textit{+=}SelfAttn(E_{concat}^n), E_{concat}^n\textit{+=}Conv1D(E_{concat}^n)$\end{small}; \\
                \begin{small}$P^n~\textit{=}~H^{n-1}\textit{+}Pool(E_{concat}^n),~where~H^0\textit{=}E^0 $\end{small}; \\
                \STATE In the Backbone block: \\
                \begin{small}$H^{n-1}\textit{+=}SelfAttn(H^{n-1}),~where~H^0\textit{=}E^0$\end{small}; \\
                \begin{small}$H^n\textit{=}P^n\textit{+}H^{n-1}\textit{+}FFN(H^{n-1})$\end{small}; \\
                \STATE In the Multi-fluctuation Patterns Mixing block: \\
                \begin{small}$E_k^n\textit{=}Conv1D(E_k^n\textit{+}H^n)_{k\in [1,K]}, E_{concat}^n\textit{=}Concatenate(\{E_k^n\}_{k\in [1,K]})$\end{small}; \\
                \begin{small}$E_{concat}^n\textit{+=}SelfAttn(E_{concat}^n$\end{small}; \\
                \begin{small}$E_{pool}\textit{=}Pool(E_{concat}^n\textit{+}Conv1D(E_{concat}^n))), \{E_k^{n}\textit{=}E_k^{n}\textit{+}E_{pool}^{n}\}_{k\in [1,K]}$\end{small}; \\
          \ENDFOR \\
          \begin{small}$E_{average}^N=\left(\sum_{k=1}^KE_k^N\right)/K$\end{small}; \\
          \FOR{ $n=1$ {\bfseries to} $N$}
                \STATE In the Mixture of Predictors: \\
                \begin{small}$S^{n}\textit{+=}Selfattn(Conv1D(S^{n})),~where~S^0\textit{=}E_{average}^N$\end{small}; \\
                \begin{small}$S^{n\textit{+}1}\textit{=}Conv1D(S^{n}\textit{+}H^n\textit{+}FFN(S^{n}))$\end{small}; \\
          \ENDFOR \\
          \STATE Obtain the prediction result through projection layer: \\
          \begin{small}$\overline{X}_{1:H}~\textit{=}~OutputHead(Flatten(S^N))$\end{small}, \\
          \STATE Update model parameters based on $\mathcal{L}_{mse}$ gradient.
          \begin{small}{$\mathcal{L}_{mse}\textit{=MSE}\left(X_{1:H},\overline{X}_{1:H}\right)$}\end{small}. \\
   \ENDFOR
\end{algorithmic}
\end{algorithm*}

\section{\myformer's limitations}
\vspace{-5pt}
\begin{table*}[htpb]
  \caption{Comparison of computational complexity between the proposed \myformer and existing methods.}\label{tab:computation}
  \centering
  \resizebox{1.8\columnwidth}{!}{
  \begin{small}
  \renewcommand{\multirowsetup}{\centering}
  \tabcolsep=0.2cm
  \renewcommand\arraystretch{1.3}
  \begin{tabular}{ccccccccc}
    \toprule
    \hline
    
    \multicolumn{1}{c}{\multirow{1}{*}{\scalebox{1.0}{Metric}}} & 
    \multicolumn{1}{c}{\rotatebox{0}{\scalebox{1.0}{TiDE}}} & 
    \multicolumn{1}{c}{\rotatebox{0}{\scalebox{1.0}{TimeMixer}}} & 
    \multicolumn{1}{c}{\rotatebox{0}{\scalebox{1.0}{PatchTST}}} & 
    \multicolumn{1}{c}{\rotatebox{0}{\scalebox{1.0}{Pets}}} & 
    \multicolumn{1}{c}{\rotatebox{0}{\scalebox{1.0}{iTransformer}}} & 
    \multicolumn{1}{c}{\rotatebox{0}{\scalebox{1.0}{SCINet}}} & 
    \multicolumn{1}{c}{\rotatebox{0}{\scalebox{1.0}{TimesNet}}} & 
    \multicolumn{1}{c}{\rotatebox{0}{\scalebox{1.0}{FEDformer}}} \\
    
    \hline
    \multirow{1}{*}{\scalebox{1.0}{Training Time (ms/iter)}} &
    263 & 271 & 319 & 376 & 435 & 454 & 477 & 722 \\

    \hline
    \multirow{1}{*}{\scalebox{1.0}{Memory Cost (GB)}} &
    2.92 & 3.14 & 4.95 & 6.17 & 7.92 & 5.91 & 6.72 & 8.44 \\

    \hline
    \bottomrule
  \end{tabular}
  \end{small}
}
\end{table*}
\vspace{-5pt}

\subsection{High arithmetic overhead and time cost.} 
In Table~\ref{tab:computation}, the computational cost of \myformer is higher than that of the popular TimeMixer and PatchTST. Therefore, the computational cost serves as the main limitation of \myformer. According to our analysis in Section~\ref{sec:complexity_analysis}, the computational cost of \myformer mainly stems from its frequency-domain-based composite pattern decoupling algorithm, SDAQ.

Although SDAQ thoroughly addresses the issue of different frequency components being coupled together, computing the spectrogram requires a substantial amount of time. In fact, the complexity of SDAQ is mainly caused by the Continuous Wavelet Transform (CWT) and the Inverse Continuous Wavelet Transform (iCWT). The advantage of introducing CWT is that the spectrogram has both frequency and time domains. This means that it can effectively represent the compositional structure of the frequency components in the original sequence as they change over time, which is crucial for the analysis of non-stationary time series. However, CWT leads to a huge computational complexity.

Considering these trade-offs, we propose replacing CWT with the Fast Fourier Transform (FFT) in SDAQ. The improved Pets can still achieve the lossless and non-overlapping decoupling of the original sequence into subsequences with different fluctuation patterns. At the same time, it can significantly reduce the computational cost of SDAQ, bringing the computational cost of Pets down to the same level as that of the popular TimeMixer.

\section{Representation Analysis}
\subsection{Linear Weights}\label{sec:linear_weights}
For intuitive illustration of the mapping relationship between the input sequence and the predicted sequence in the frequency-domain space across different tasks, we first transform the original sequence to the frequency domain via the Fourier transform. Subsequently, we employ the linear layer to map the frequency components of the observed sequence to those of the predicted sequence. Finally, the frequency components of the predicted sequence are transformed back to the time domain through the inverse Fourier transform, thereby obtaining the predicted sequence, as \begin{small}$Y\textit{=}iFT(Linear(FT(X)))$\end{small}

\begin{figure*}[htbp]
\begin{center}
	\centerline{\includegraphics[width=2.0\columnwidth]{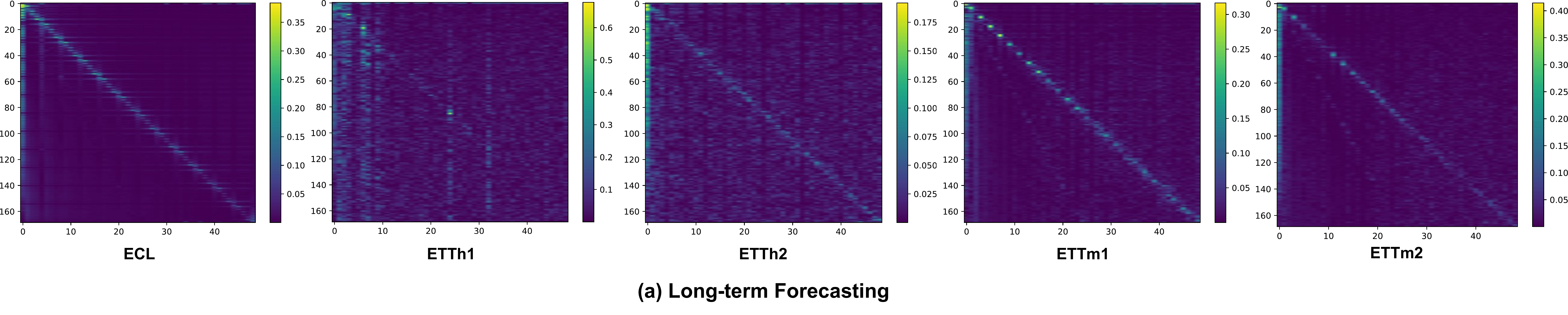}}
	\centerline{\includegraphics[width=2.0\columnwidth]{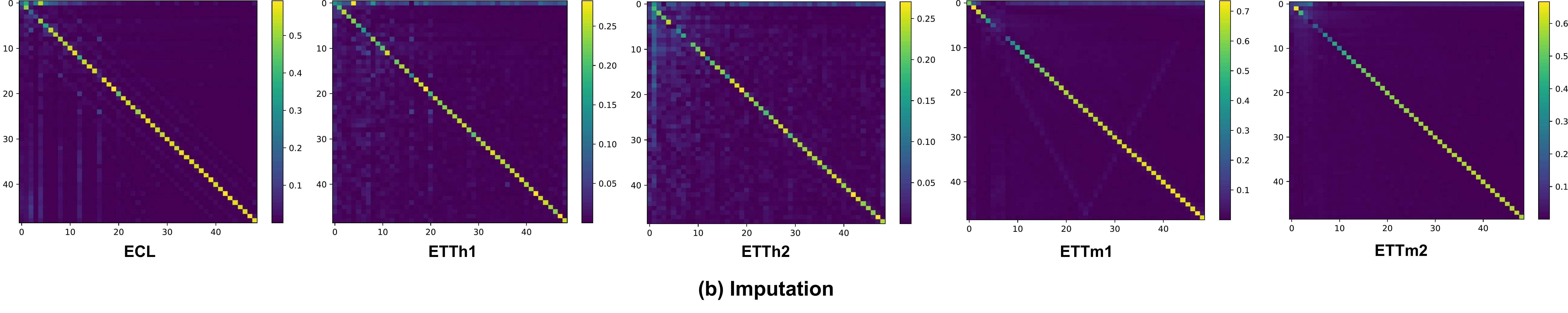}}
	\centerline{\includegraphics[width=2.0\columnwidth]{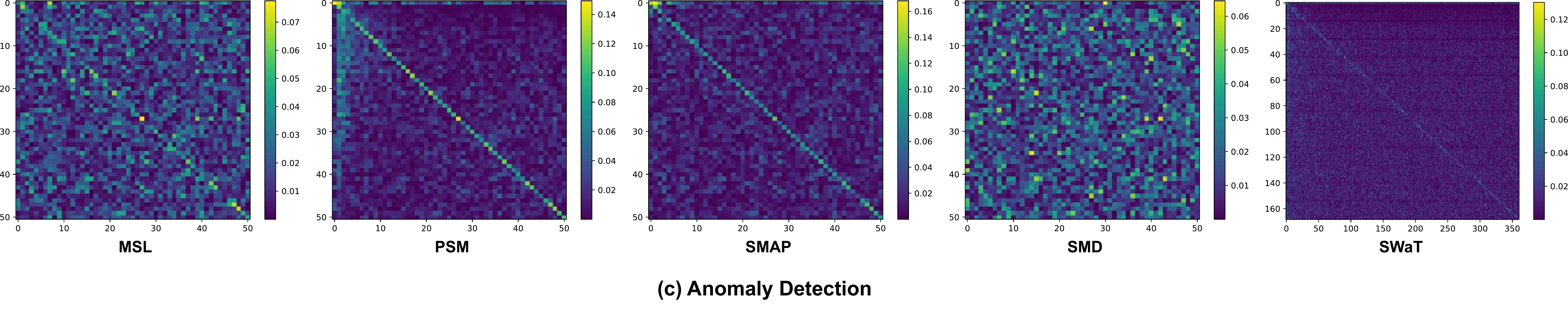}}
	\centerline{\includegraphics[width=2.0\columnwidth]{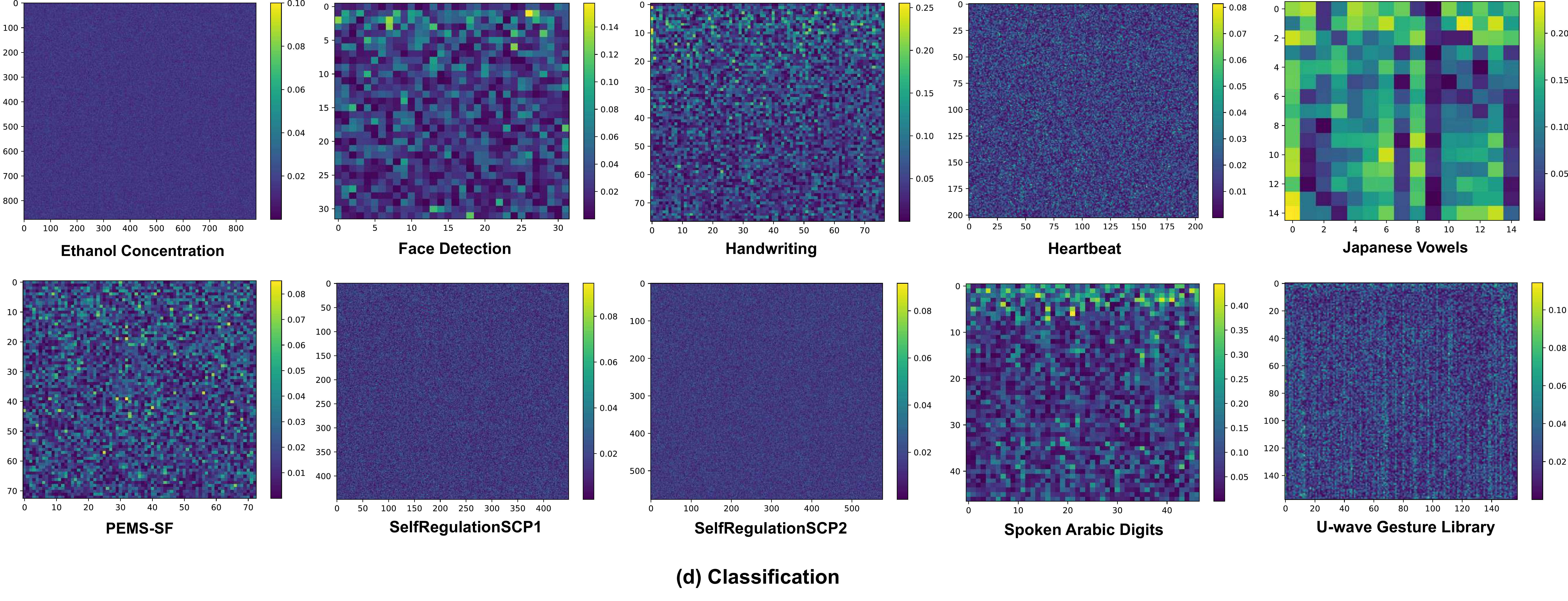}}
	\caption{
Representation Analysis Based on Visualization of Linear Weights. We utilize the linear layer to enable the establishment of a mapping relationship between the frequency components of the input and output sequences in the frequency - domain space. Each task exhibits a specific mapping pattern.
}\label{fig:repre_linear}
\end{center}
\vspace{-10pt}
\end{figure*}

We conducted comprehensive experiments on benchmarks composed of 4 tasks (forecasting, imputation, anomaly detection, and classification) and a total of 20 datasets. Fig.~\ref{fig:repre_linear} presents the visualization of the linear weights between the original and generated sequence in the frequency domain, and diverse tasks exhibit distinct active patterns. Specifically, in the tasks of prediction, imputation, and anomaly detection, the weight scores are concentrated around the diagonal positions. 
Specifically, in prediction, the weights are preponderantly distributed within the low-frequency components (the light blue regions in Fig.~\ref{fig:repre_linear} (a)). For imputation, the weights clustered on the mapping of the input and output sequences at identical frequency components (the diagonal regions in Fig.~\ref{fig:repre_linear}(b)). This indicates that for generative tasks, the frequency-domain components of the input and output sequences tend to establish a comprehensive mapping relationship. 
In classification Fig.~\ref{fig:repre_linear}(d), the weights are stochastically dispersed throughout the global scope.    Meanwhile, anomaly detection Fig.~\ref{fig:repre_linear}(c) amalgamates the characteristics of sundry tasks. Moreover, conventional decomposition and multi-scale augmentation algorithms fall short of effectively disentangling disparate fluctuation patterns, as the high-frequency pattern exists only in the seasonal and fine-grained segments and remains superposed and intertwined with other patterns.

\subsection{Intra-pattern Dependencies}\label{sec:intra_pattern}
\begin{figure*}[htbp]
\begin{center}
	\centerline{\includegraphics[width=2.0\columnwidth]{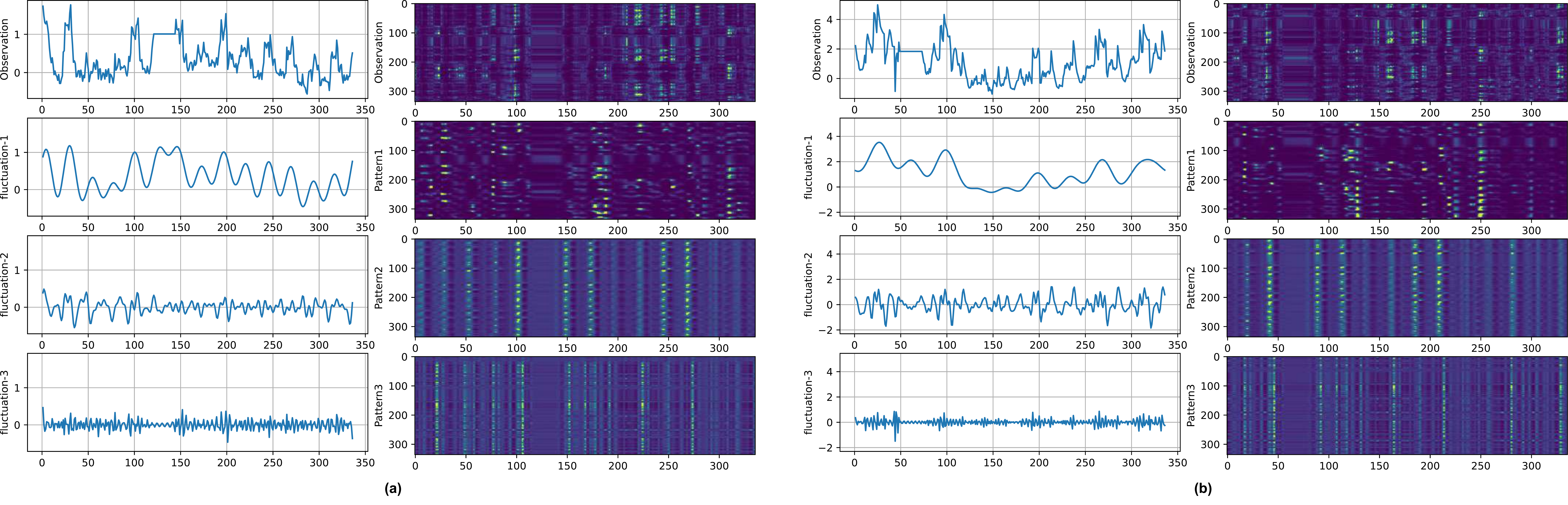}}
	\centerline{\includegraphics[width=2.0\columnwidth]{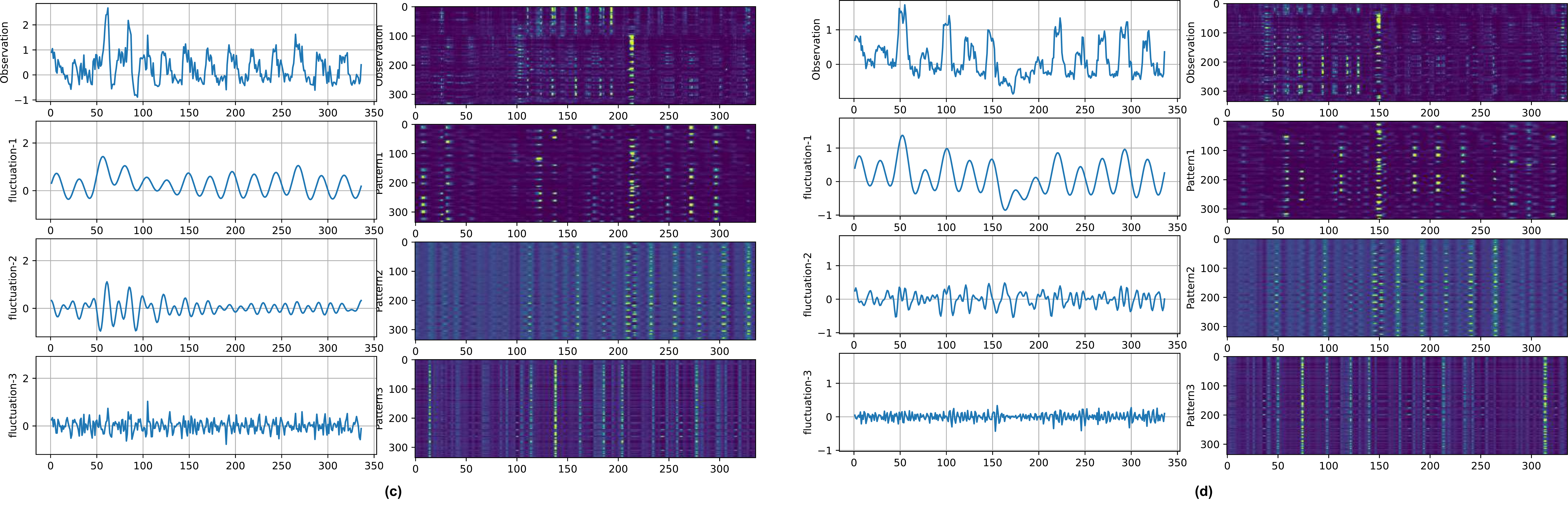}}
	\centerline{\includegraphics[width=2.0\columnwidth]{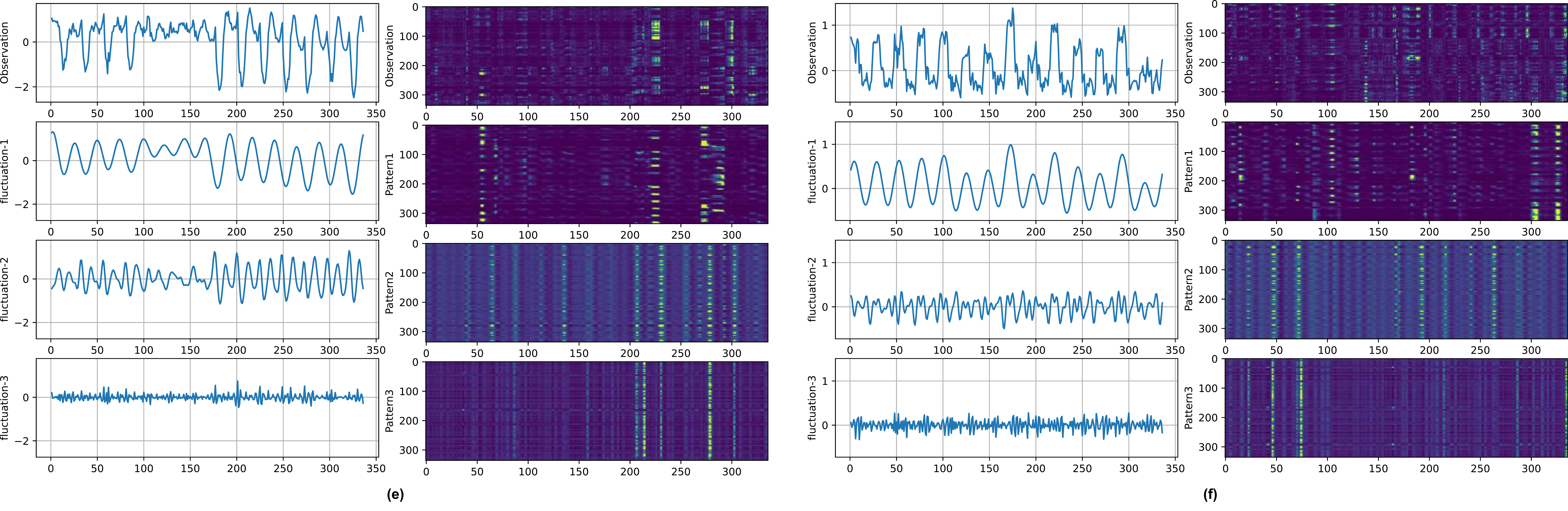}}
	\centerline{\includegraphics[width=2.0\columnwidth]{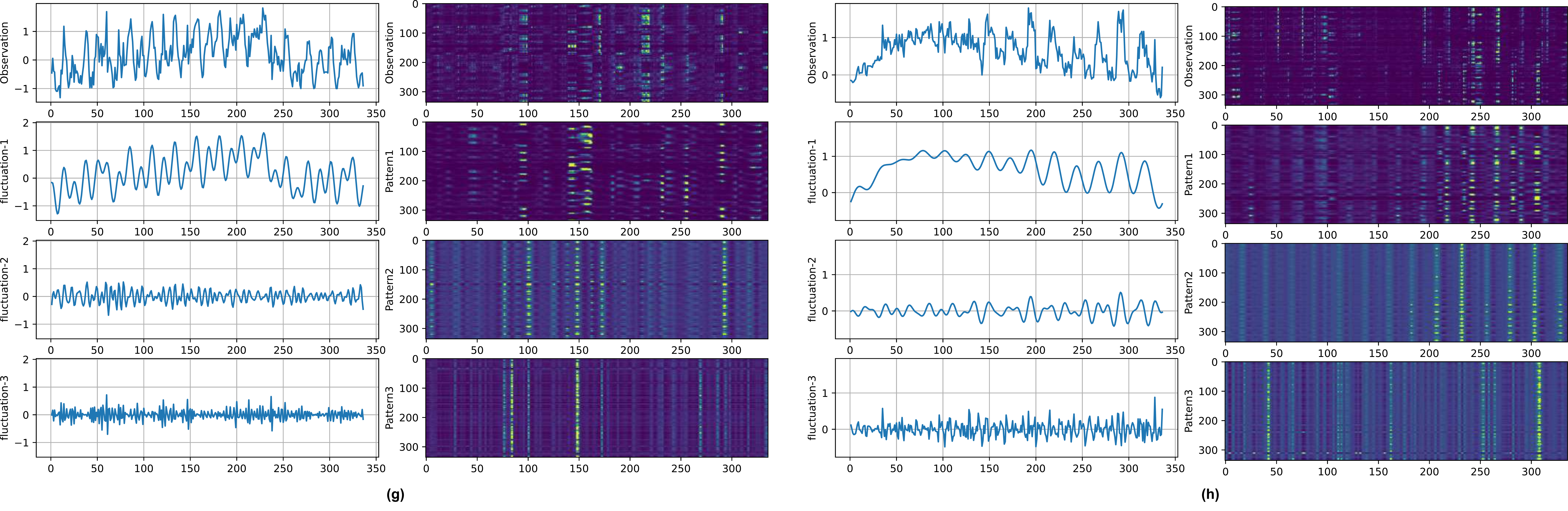}}
    \vspace{-5pt}
	\caption{
Visualization of the decomposition algorithm on the \textbf{ETTh1} dataset. Additionally, we present the point-wise attention scores calculated on the observed sequence and the three decomposed sequences.
}\label{fig:repre_attn4patterns_etth1}
\end{center}
\vspace{-10pt}
\end{figure*}

\begin{figure*}[htbp]
\begin{center}
	\centerline{\includegraphics[width=2.0\columnwidth]{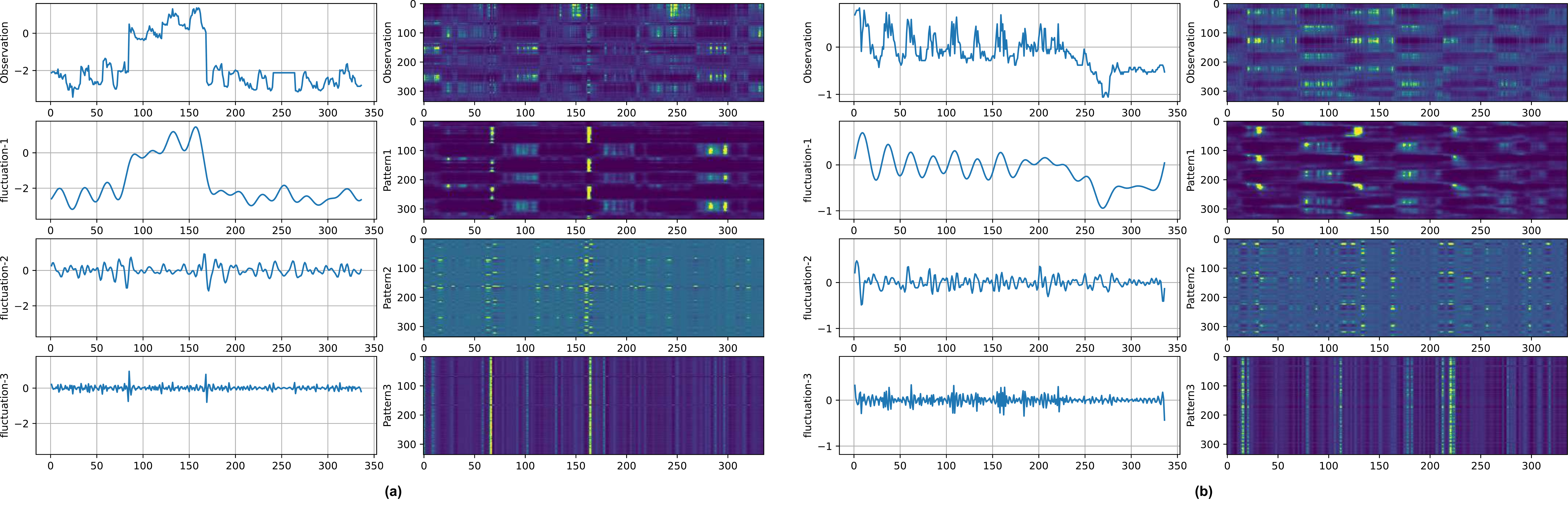}}
	\centerline{\includegraphics[width=2.0\columnwidth]{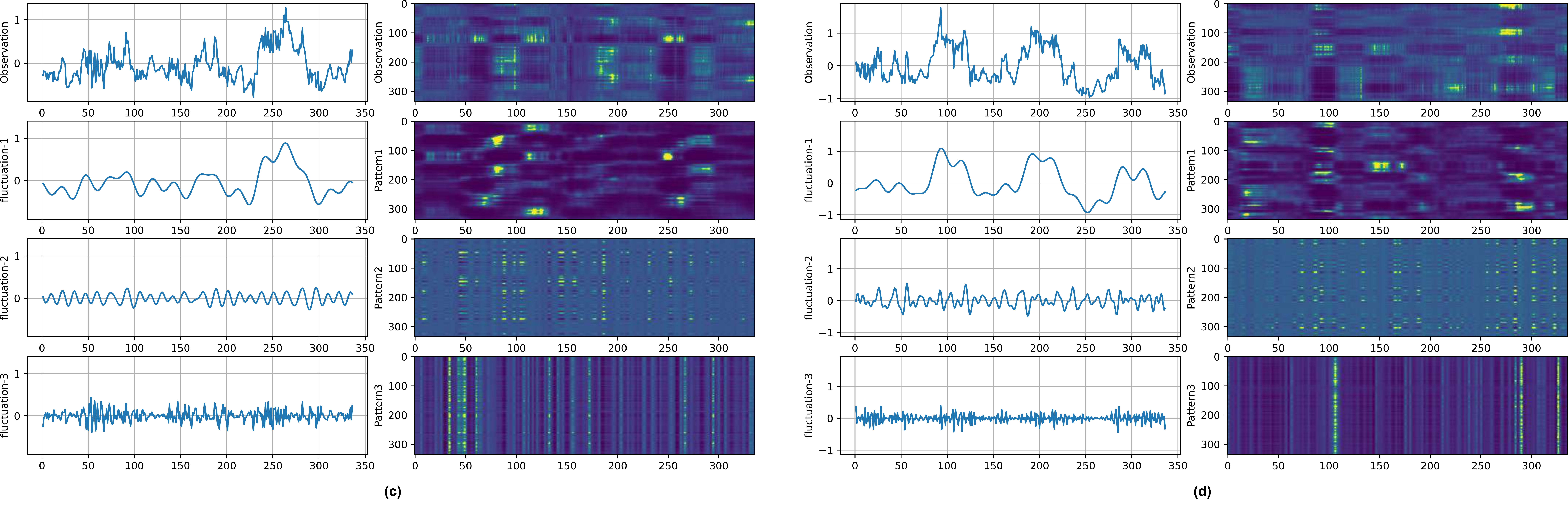}}
	\centerline{\includegraphics[width=2.0\columnwidth]{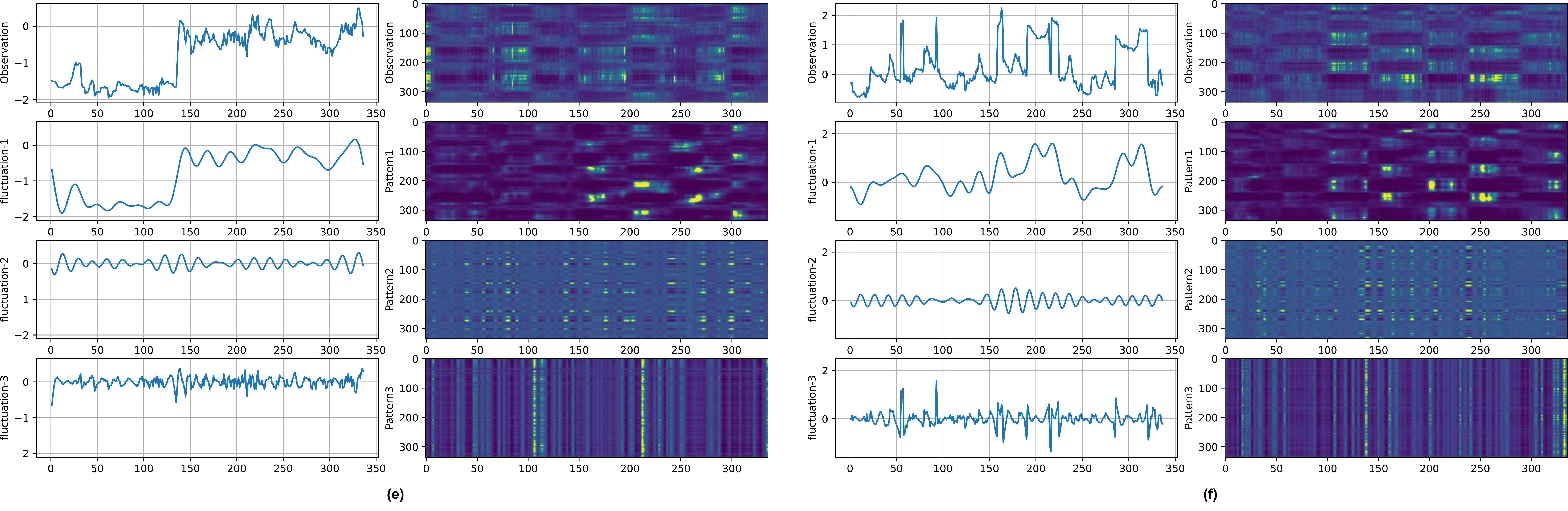}}
	\centerline{\includegraphics[width=2.0\columnwidth]{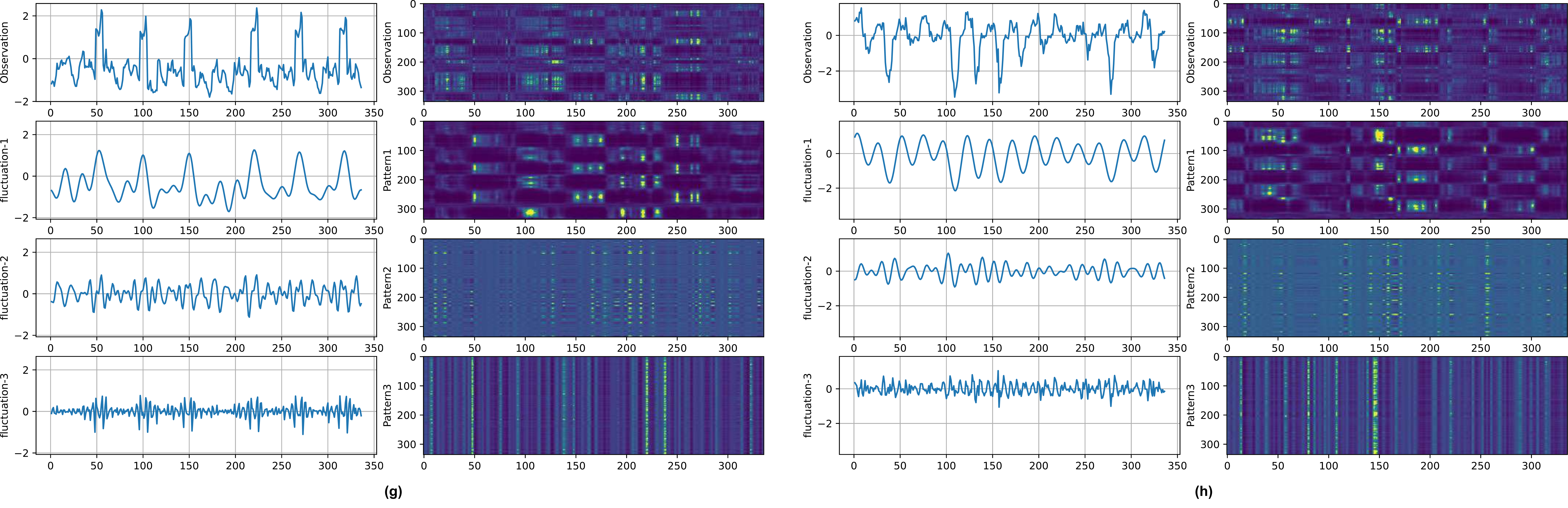}}
    \vspace{-5pt}
	\caption{
Visualization of the decomposition algorithm on the \textbf{ETTm1} dataset. Additionally, we present the point-wise attention scores calculated on the observed sequence and the three decomposed sequences.
}\label{fig:repre_attn4patterns_ettm1}
\end{center}
\vspace{-10pt}
\end{figure*}

In Fig.~\ref{fig:repre_attn4patterns_etth1} and \ref{fig:repre_attn4patterns_ettm1}, the original time series samples are decoupled into three independent fluctuation modes, and the fluctuation modes 1 to 3 exhibit the pattern characteristics from long-to-short period and low-to-high frequency respectively. We then compute the global dependencies between all time points in each fluctuation pattern, and the top plot shows the attention score visualization for each pattern. 
Specifically, in the original sequence, the dependencies between time points are chaotic, which is caused by the mutual coupling between multiple periodic patterns in the original sequence. In contrast, the decoupled fluctuation patterns show distinct periodic variation characteristics. Concretely, in the fluctuation pattern 1, the bright yellow blob shows a sparse distribution characteristic, which means that the attention mechanism learns the dependence between long periods from it, which we call the long wave pattern. However, in the fluctuation mode 3, the bright yellow blobs show the characteristics of clustered distribution, and the attention learns the dependence between short periods from it, which we call the short-wave mode.

\subsection{Inter-pattern Dependencies}\label{sec:inter_pattern}
\begin{figure*}[htbp]
\begin{center}
	\centerline{\includegraphics[width=2.0\columnwidth]{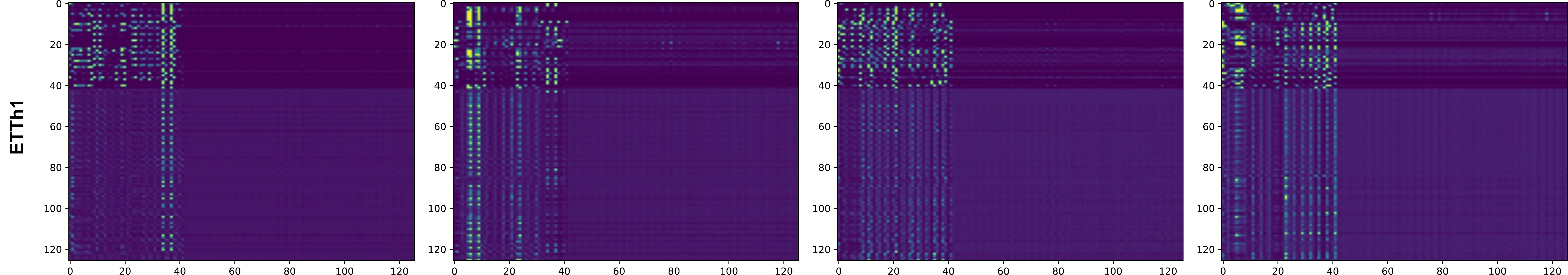}}
    \vspace{10pt}
	\centerline{\includegraphics[width=2.0\columnwidth]{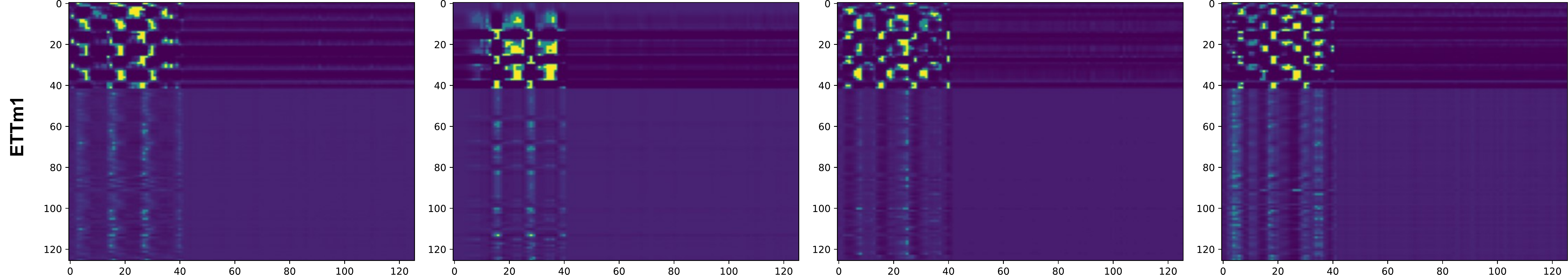}}
    \vspace{10pt}
	\centerline{\includegraphics[width=2.0\columnwidth]{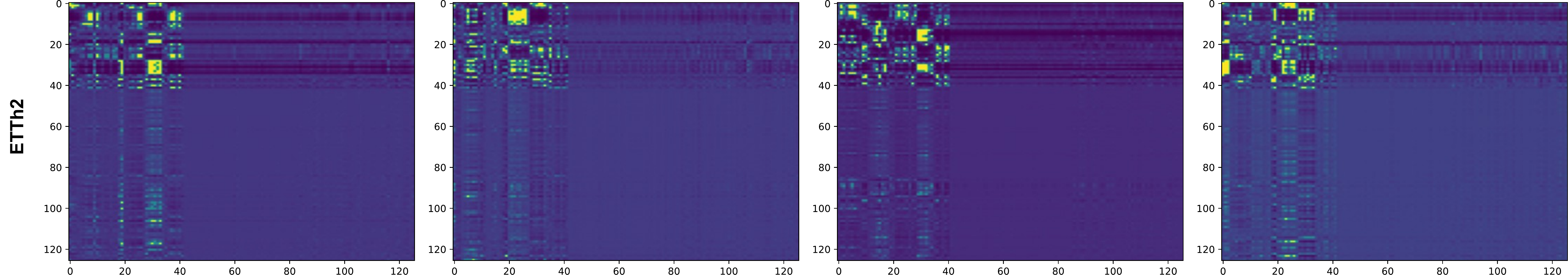}}
    \vspace{10pt}
	\centerline{\includegraphics[width=2.0\columnwidth]{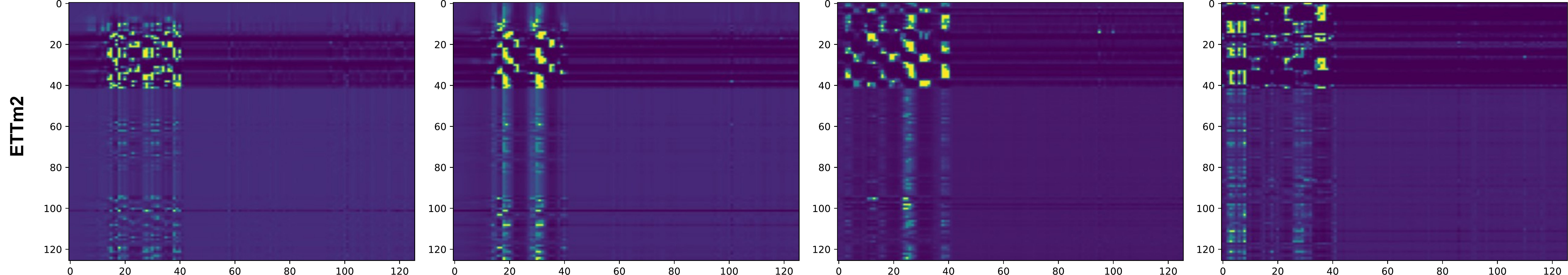}}
    \vspace{10pt}
	\centerline{\includegraphics[width=2.0\columnwidth]{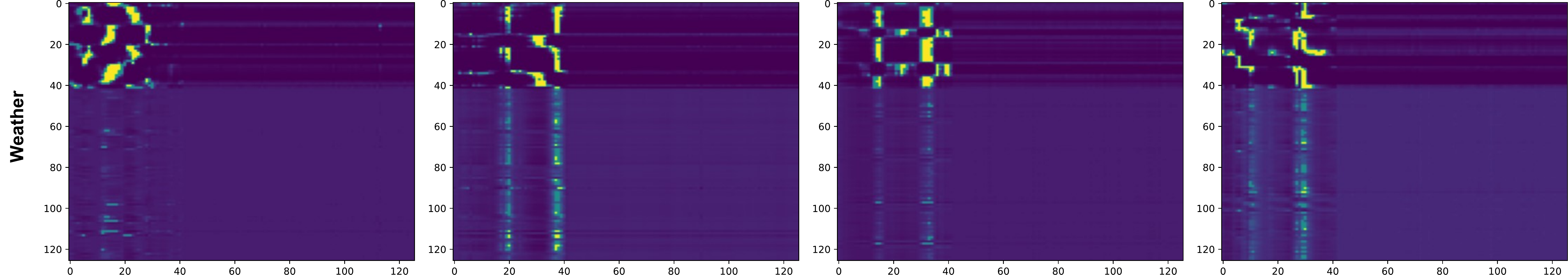}}
    \vspace{10pt}
	\centerline{\includegraphics[width=2.0\columnwidth]{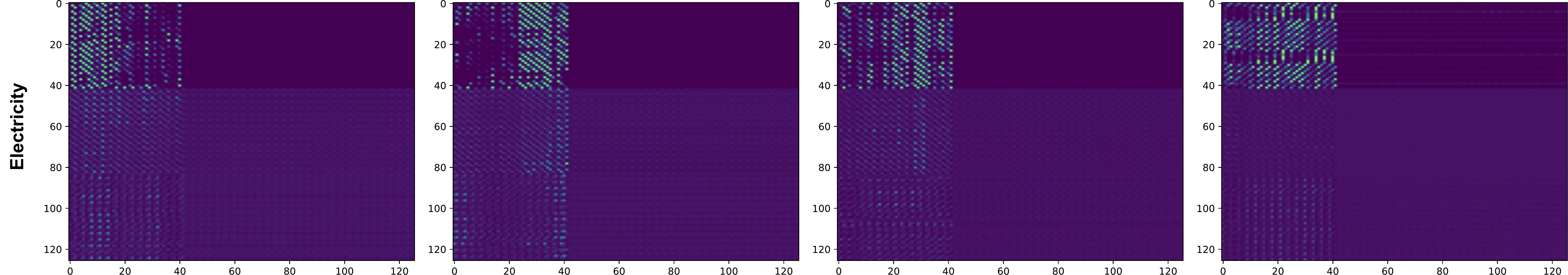}}
	\caption{
Attention scores among different fluctuation patterns were calculated and visualized in the \textbf{forecasting} task.
}\label{fig:repre_attn4tasks_forecasting}
\end{center}
\vspace{-10pt}
\end{figure*}

\begin{figure*}[htbp]
\begin{center}
	\centerline{\includegraphics[width=2.0\columnwidth]{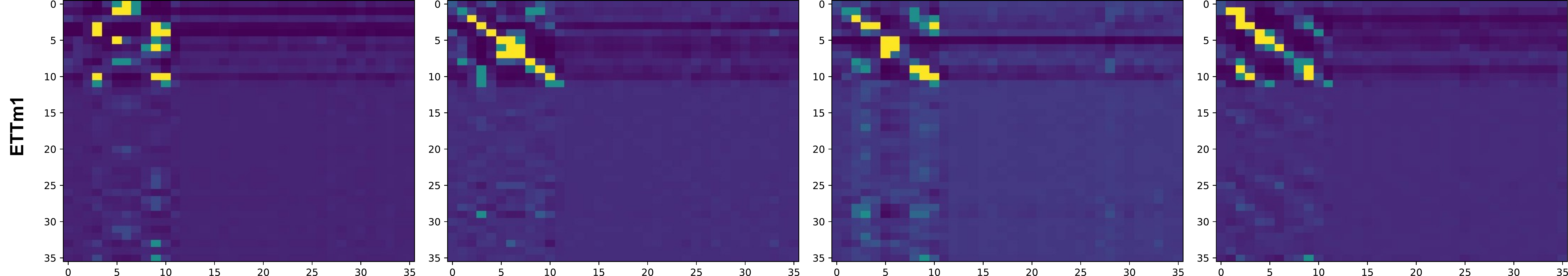}}
    \vspace{10pt}
	\centerline{\includegraphics[width=2.0\columnwidth]{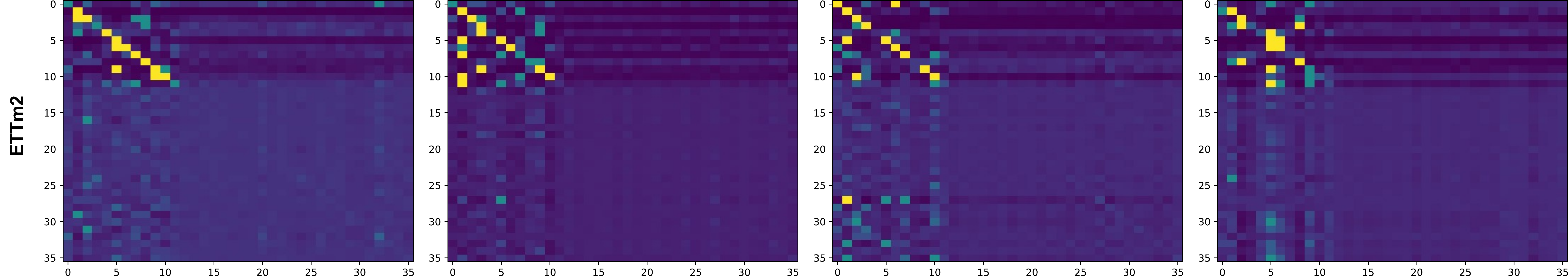}}
    \vspace{10pt}
	\centerline{\includegraphics[width=2.0\columnwidth]{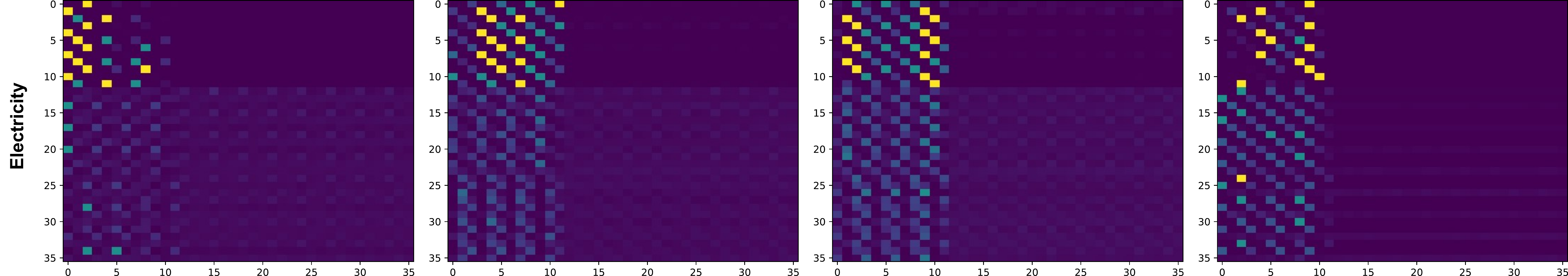}}
	\caption{
Attention scores among different fluctuation patterns were calculated and visualized in the \textbf{imputation} task.
}\label{fig:repre_attn4tasks_imputation}
\end{center}
\vspace{-10pt}
\end{figure*}

\begin{figure*}[htbp]
\begin{center}
	\centerline{\includegraphics[width=2.0\columnwidth]{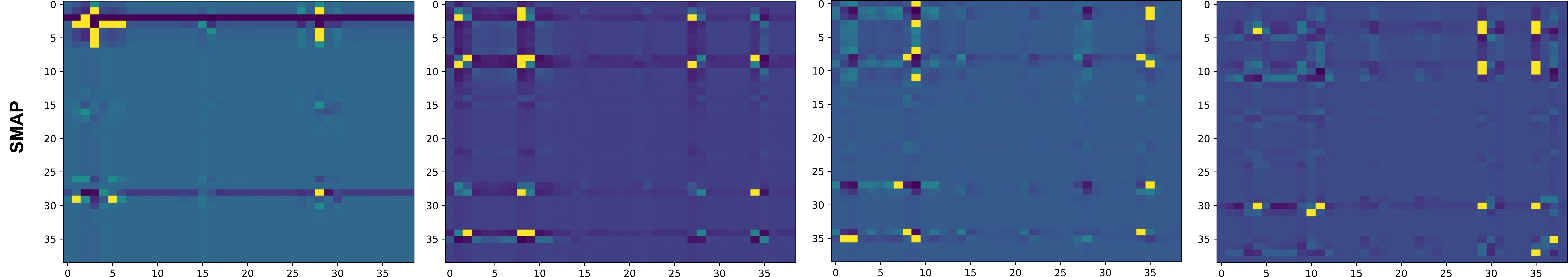}}
    \vspace{10pt}
	\centerline{\includegraphics[width=2.0\columnwidth]{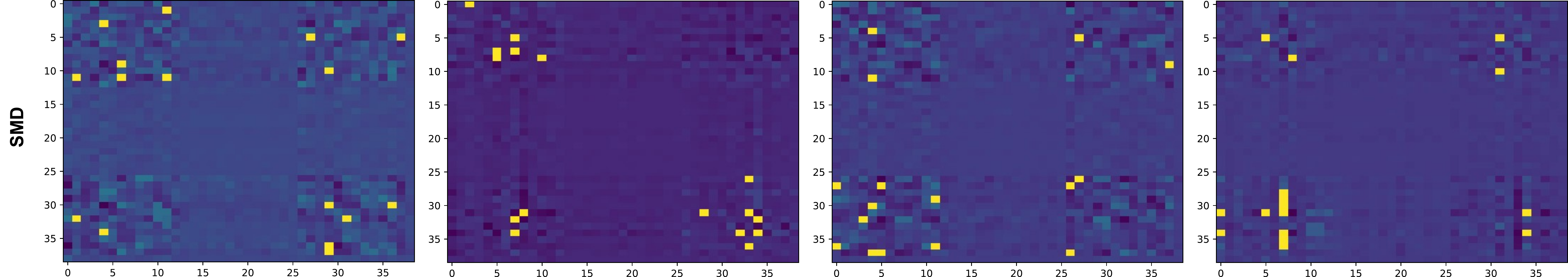}}
    \vspace{10pt}
	\centerline{\includegraphics[width=2.0\columnwidth]{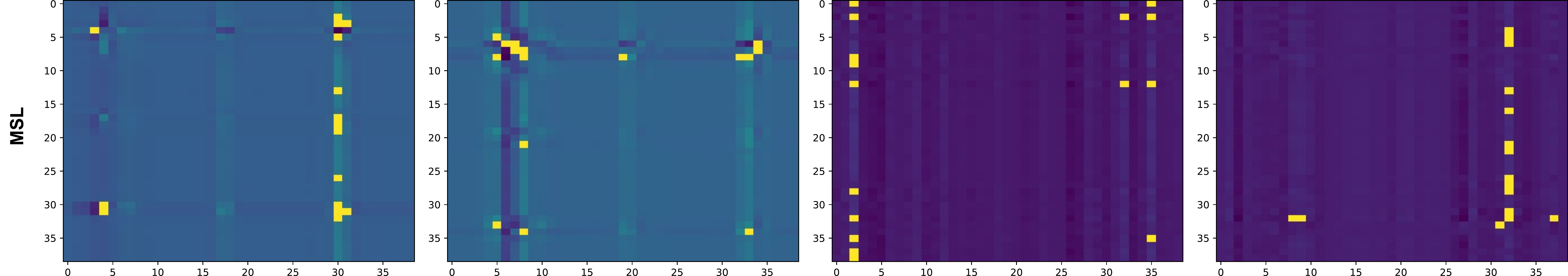}}
	\caption{
Attention scores among different fluctuation patterns were calculated and visualized in the \textbf{anomaly detection} task.
}\label{fig:repre_attn4tasks_anomaly}
\end{center}
\vspace{-10pt}
\end{figure*}

\begin{figure*}[htbp]
\begin{center}
	\centerline{\includegraphics[width=2.0\columnwidth]{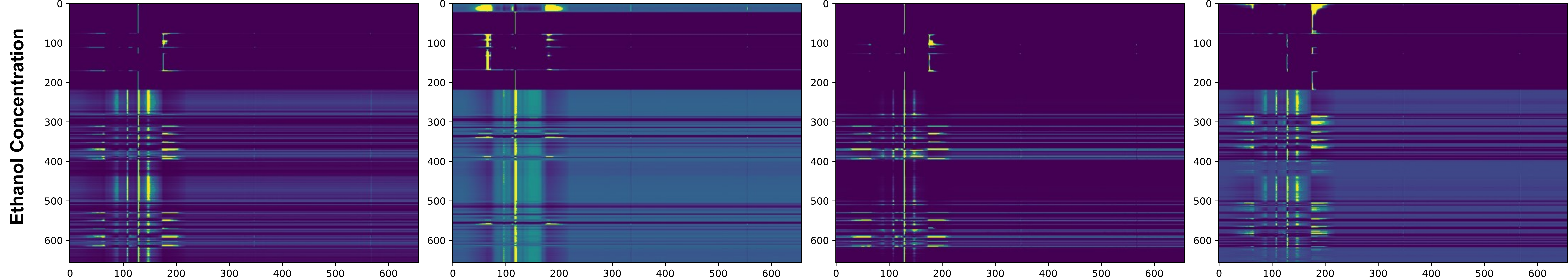}}
    \vspace{10pt}
	\centerline{\includegraphics[width=2.0\columnwidth]{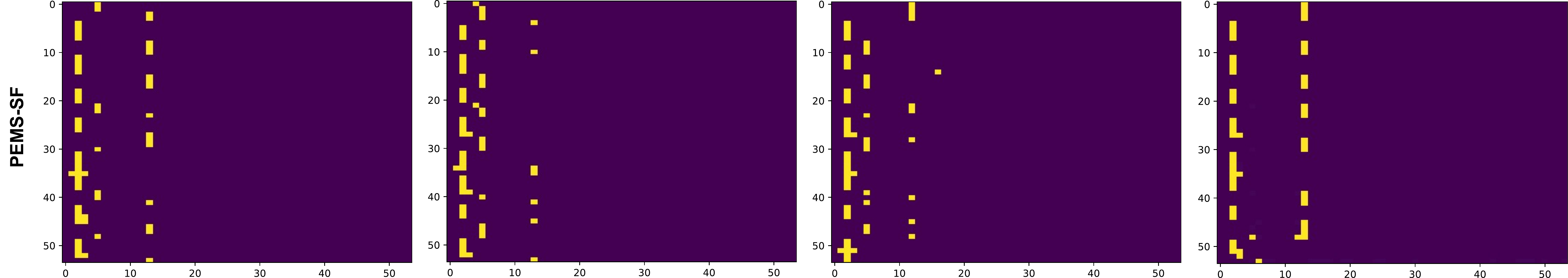}}
    \vspace{10pt}
	\centerline{\includegraphics[width=2.0\columnwidth]{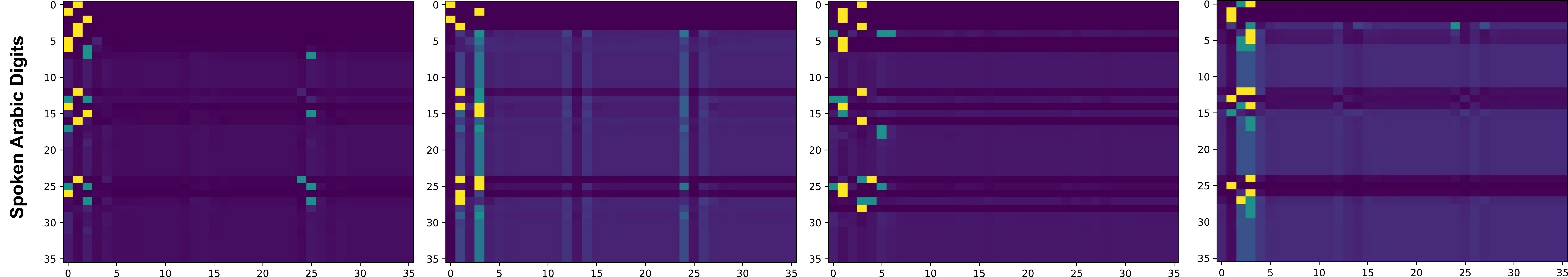}}
    \vspace{10pt}
	\centerline{\includegraphics[width=2.0\columnwidth]{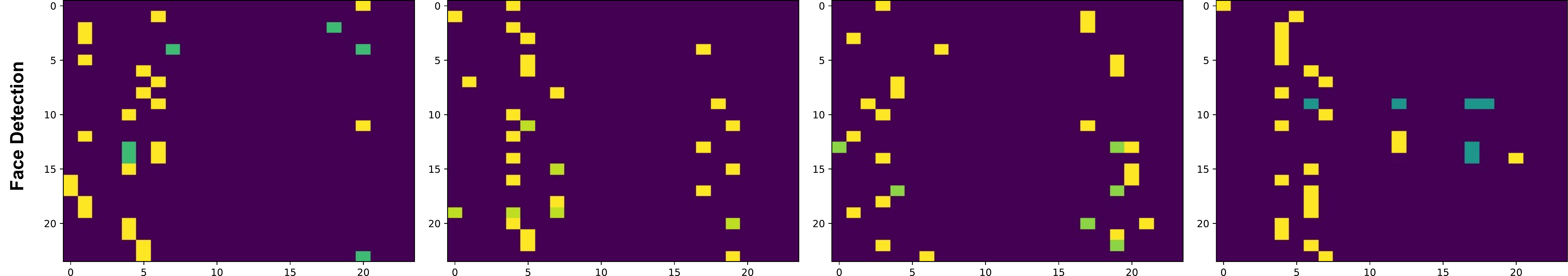}}
    \vspace{10pt}
	\centerline{\includegraphics[width=2.0\columnwidth]{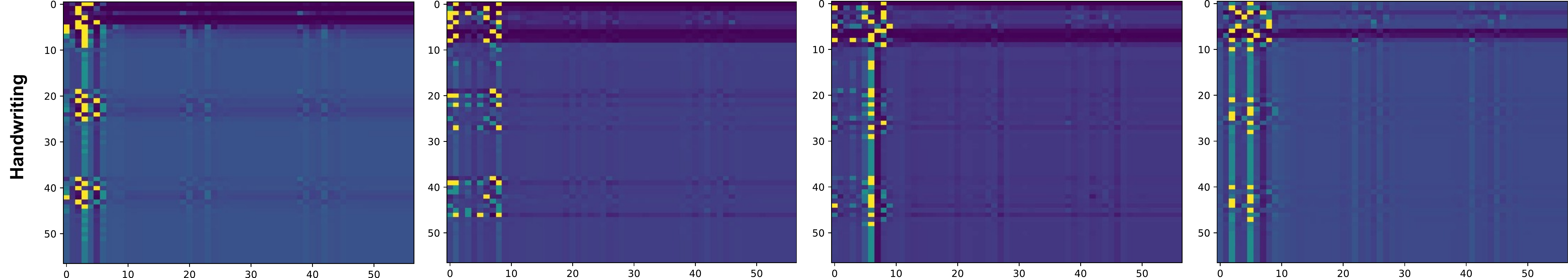}}
	\caption{
Attention scores among different fluctuation patterns were calculated and visualized in the \textbf{classification} task.
}\label{fig:repre_attn4tasks_classification}
\end{center}
\vspace{-10pt}
\end{figure*}

In Fig.~\ref{fig:repre_attn4tasks_forecasting}, Fig.~\ref{fig:repre_attn4tasks_imputation}, Fig.~\ref{fig:repre_attn4tasks_anomaly} and Fig.~\ref{fig:repre_attn4tasks_classification}, we present the visualization of the importance of different fluctuation patterns for specific downstream tasks. 
We first decouple the observed sequence into three fluctuating sequences. Subsequently, each of these fluctuating sequences undergoes a patch-based embedding operation to obtain three groups of pattern tokens, where each set contains 40 tokens. Finally, all 120 tokens are concatenated in the order from low-frequency pattern to high-frequency pattern, and then fed into the transformer, where the attention scores between these tokens in different tasks are calculated.

Specifically, in the forecasting and imputation tasks, we notice that the yellow bright spots only appear in the top left corner, which means that there is a strong dependence between the shortwave patterns (\begin{small}$x\in [0,40]$\end{small}) of the input sequence, \textbf{short wave (long period) patterns are crucial for both forecasting and imputation tasks}. This is also in line with cognition, since long-term trends are important for prediction tasks, and high-frequency oscillations in them are likely due to noise. In contrast, in the anomaly detection and classification tasks, the yellow bright spot distribution will appear in the small right corner and the bottom left corner, which means that the high-frequency perturbation is critical for the classification and anomaly detection tasks. It can be understood that abrupt signals are often caused by reasons such as mechanical failures, and therefore these high-frequency patterns are very important for discriminative tasks.

\section{Full Results} \label{section:full_results}
Due to the space limitation of the main text, we place the full results of all experiments in the following: long-term forecasting under full-shot setting in Table~\ref{tab:forecasting_fullshot}, long-term forecasting under few-shot setting in Table~\ref{tab:forecasting_fewshot}, long-term forecasting under zero-shot setting in Table~\ref{tab:forecasting_zeroshot_brief}, univariate short-term forecasting in Table~\ref{tab:short_m4}, multivariate short-term forecasting in Table~\ref{tab:short_pems}, anomaly detection in Table~\ref{tab:anomaly} and classification in Table~\ref{tab:classification}.
In addition, the complete ablation experiments of the enhancement strategy and the original baseline in the long-term forecasting task in Table~\ref{tab:forecasting_enhanced}

\begin{table*}[htbp]
    \centering
    \small
    \tabcolsep=0.8mm
    \renewcommand\arraystretch{1.1}
    \caption{Comparison of the complete performance with diverse prediction length on full-data long-term forecasting task. \textbf{Avg.} is averaged from all four prediction lengths. We boldface the best performance.}\label{tab:forecasting_fullshot}
    \vspace{0.2em}
    \resizebox{0.95\textwidth}{!}{
\begin{tabular}{cc|cc|cc|cc|cc|cc|cc|cc|cc|cc|cc|cc}
    \toprule
    \hline
    \multicolumn{2}{c}{\multirow{2}{*}{{Models}}} &
    \multicolumn{2}{c}{\textbf{Pets}} & 
    \multicolumn{2}{c}{DUET} & 
    \multicolumn{2}{c}{PDF} & 
    \multicolumn{2}{c}{iTrans.} & 
    \multicolumn{2}{c}{Pathformer} & 
    \multicolumn{2}{c}{FITS} & 
    \multicolumn{2}{c}{TimeMixer} & 
    \multicolumn{2}{c}{PatchTST} & 
    \multicolumn{2}{c}{Crossformer} & 
    \multicolumn{2}{c}{TimesNet} & 
    \multicolumn{2}{c}{DLinear} \\
    
    \multicolumn{2}{c}{} & 
    \multicolumn{2}{c}{{(\textbf{Ours})}} & 
    \multicolumn{2}{c}{{\cite{qiu2025duet}}} & 
    \multicolumn{2}{c}{{\cite{dai2024pdf}}} & 
    \multicolumn{2}{c}{{\cite{liu2023itransformer}}} & 
    \multicolumn{2}{c}{{\cite{chen2024pathformer}}} & 
    \multicolumn{2}{c}{{\cite{xu2023fits}}} & 
    \multicolumn{2}{c}{{\cite{Wang2024TimeMixerDM}}} & 
    \multicolumn{2}{c}{{\cite{Nie2023PatchTST}}} & 
    \multicolumn{2}{c}{{\cite{zhang2023crossformer}}} & 
    \multicolumn{2}{c}{{\cite{wu2023timesnet}}} & 
    \multicolumn{2}{c}{{\cite{Zeng2022DLinear}}} \\

    \cline{3-24} 
    
    \multicolumn{2}{c}{{Metric}} & 
    MSE & MAE & MSE & MAE & MSE & MAE & MSE & MAE & 
    MSE & MAE & MSE & MAE & MSE & MAE & MSE & MAE & MSE & MAE & MSE & MAE & MSE & MAE \\
    
    \hline 
    \hline
    
    \multicolumn{1}{c|}{\multirow{5}{*}{\rotatebox{90}{ETTh1}}} & 96 & 
    \textbf{0.351} & \textbf{0.372}
& 0.352 & 0.384 & 0.360 & 0.391 & 0.386 & 0.405 & 0.372 & 0.392 & 0.376 & 0.396
& 0.372 & 0.401 & 0.377 & 0.397 & 0.411 & 0.435 & 0.389 & 0.412 & 0.379 & 0.403 \\
    
    \multicolumn{1}{c|}{} & 192 & 
    \textbf{0.389} & \textbf{0.394}
& 0.398 & 0.409 & 0.392 & 0.414 & 0.424 & 0.440 & 0.408 & 0.415 & 0.400 & 0.418
& 0.413 & 0.430 & 0.409 & 0.425 & 0.409 & 0.438 & 0.440 & 0.443 & 0.408 & 0.419 \\
\hline
    \bottomrule
    \end{tabular}
    } 
    \vspace{-0.5cm}
\end{table*}

\begin{table*}[htbp]
    \centering
    \small
    \tabcolsep=0.6mm
    \renewcommand\arraystretch{1.5}
    \caption{Comparison of the complete performance with diverse prediction length on few-data (10\% data) long-term forecasting task. \textbf{Avg.} is averaged from all four prediction lengths, that $\{96, 192, 336, 720\} $. We boldface the best performance.}\label{tab:forecasting_fewshot}
    \vspace{0.2em}
    \resizebox{\textwidth}{!}{
    \begin{tabular}{cc|cc|cc|cc|cc|cc|cc|cc|cc|cc|cc|cc|cc}
    \toprule
    \hline
    \multicolumn{2}{c}{\multirow{2}{*}{{Models}}} &
    \multicolumn{2}{c}{{\textbf{\myformer}}} &
    \multicolumn{2}{c}{{TimeMixer}} &
    \multicolumn{2}{c}{{iTransformer}} &
    \multicolumn{2}{c}{{TiDE}} &
    \multicolumn{2}{c}{{DLinear}} &
    \multicolumn{2}{c}{{PatchTST}} &
    \multicolumn{2}{c}{{TimesNet}} &
    \multicolumn{2}{c}{{FEDformer}} &
    \multicolumn{2}{c}{{Autoformer}} &
    \multicolumn{2}{c}{{Stationary}} &
    \multicolumn{2}{c}{{ETSformer}} &
    \multicolumn{2}{c}{{LightTS}} \\

    \multicolumn{2}{c}{} & 
    \multicolumn{2}{c}{{(\textbf{Ours})}} & 
    \multicolumn{2}{c}{{\cite{Wang2024TimeMixerDM}}} &
    \multicolumn{2}{c}{{\cite{liu2023itransformer}}}& 
    \multicolumn{2}{c}{{\cite{Das2023LongtermTiDE}}} &
    \multicolumn{2}{c}{{\cite{Zeng2022DLinear}}} &
    \multicolumn{2}{c}{{\cite{Nie2023PatchTST}}} & 
    \multicolumn{2}{c}{{\cite{wu2023timesnet}}} & 
    \multicolumn{2}{c}{{\cite{zhou2022fedformer}}} & 
    \multicolumn{2}{c}{{\cite{wu2021autoformer}}} & 
    \multicolumn{2}{c}{{\cite{Liu2022NonstationaryTR}}} &  
    \multicolumn{2}{c}{{\cite{woo2022etsformer}}} & 
    \multicolumn{2}{c}{{\cite{Campos2023LightTSLT}}} \\

    \cline{3-26} 
    
    \multicolumn{2}{c}{{Metric}} & 
    MSE & MAE & MSE & MAE & MSE & MAE & MSE & MAE & MSE & MAE & MSE & MAE & MSE & MAE &
    MSE & MAE & MSE & MAE & MSE & MAE & MSE & MAE & MSE & MAE \\
    
    \hline 
    \hline
    
    \multicolumn{1}{c|}{\multirow{5}{*}{\rotatebox{90}{Weather}}} & 96 & 
    \textbf{0.150} & \textbf{0.200} &
    {0.159} & {0.212} &
    {0.191} & {0.250} &
    {0.163} & {0.220} &               
    {0.171} & {0.224} & 
    {0.165} & {0.215} &
    {0.184} & {0.230} &
    {0.188} & {0.253} &
    {0.221} & {0.297} &
    {0.192} & {0.234} &
    {0.199} & {0.272} &
    {0.217} & {0.269} \\
\hline
    \bottomrule
    \end{tabular}
    } 
    \vspace{-0.5cm}
\end{table*}

\begin{table}[htbp]
    \centering
    \small
    \tabcolsep=0.6mm
    \renewcommand\arraystretch{1.2}
    \caption{Comparison of the performance on \textbf{Zero Shot Forecasting} task. We present the model’s average performance across multiple prediction horizons.}\label{tab:forecasting_zeroshot_brief_without_llm}
    \vspace{0.3em}
    \resizebox{0.8\textwidth}{!}{
    \begin{tabular}{ccccccccccccccc}
    \toprule
    \hline
    \multicolumn{1}{c}{\multirow{2}{*}{Models}} &
    \multicolumn{2}{c}{\textbf{\myformer}} & 
    \multicolumn{2}{c}{TimeMixer} & 
    \multicolumn{2}{c}{LLMTime} & 
    \multicolumn{2}{c}{PatchTST} & 
    \multicolumn{2}{c}{TimesNet} & 
    \multicolumn{2}{c}{iTransformer} & 
    \multicolumn{2}{c}{TiDE} \\

    \multicolumn{1}{c}{} & 
    \multicolumn{2}{c}{{(\textbf{Ours})}} & 
    \multicolumn{2}{c}{{\cite{Wang2024TimeMixerDM}}} &
    \multicolumn{2}{c}{{\cite{llmtime}}} & 
    \multicolumn{2}{c}{{\cite{Nie2023PatchTST}}} & 
    \multicolumn{2}{c}{{\cite{wu2023timesnet}}} &
    \multicolumn{2}{c}{{\cite{liu2023itransformer}}} &
    \multicolumn{2}{c}{{\cite{Das2023LongtermTiDE}}} \\

    \cline{1-15} 
    
    \multicolumn{1}{c}{Metric} & 
    MSE & MAE & MSE & MAE & MSE & MAE & MSE & MAE & MSE & MAE & MSE & MAE & MSE & MAE\\
    
    \hline
    
    \multirow{1}{*}{\rotatebox{0}{ETT\scalebox{0.8}{h1$\rightarrow$h2}}} & 
    \textbf{0.385} & \textbf{0.405} & 
    0.427 & 0.424 & 0.992 & 0.708 & 0.380 & 0.405 & 0.421 & 0.431 & 0.481 & 0.474 & 0.593 & 0.582 \\

    \multirow{1}{*}{\rotatebox{0}{ETT\scalebox{0.8}{h1$\rightarrow$m2}}} & 
    \textbf{0.312} & \textbf{0.353} & 
    0.361 & 0.397 & 1.867 & 0.869 & 0.314 & 0.360 & 0.327 & 0.361 & 0.311 & 0.361 & 0.563 & 0.547 \\

    \multirow{1}{*}{\rotatebox{0}{ETT\scalebox{0.8}{h2$\rightarrow$h1}}} & 
    \textbf{0.512} & \textbf{0.493} & 
    0.679 & 0.577 & 1.961 & 0.981 & 0.565 & 0.513 & 0.865 & 0.621 & 0.552 & 0.511 & 0.588 & 0.556 \\

    \multirow{1}{*}{\rotatebox{0}{ETT\scalebox{0.8}{m1$\rightarrow$h2}}} & 
    \textbf{0.432} & \textbf{0.436} & 
    0.452 & 0.441 & 0.992 & 0.708 & 0.439 & 0.438 & 0.457 & 0.454 & 0.434 & 0.438 & 0.543 & 0.535 \\

    \multirow{1}{*}{\rotatebox{0}{ETT\scalebox{0.8}{m1$\rightarrow$m2}}} & 
    \textbf{0.293} & \textbf{0.326} & 
    0.329 & 0.357 & 1.867 & 0.869 & 0.296 & 0.334 & 0.322 & 0.354 & 0.324 & \textbf{0.331} & 0.534 & 0.527 \\
\hline
    \bottomrule
    \end{tabular}
    } 
\vspace{-3mm}
\end{table}

\begin{table*}[htbp]
    \centering
    \small
    \tabcolsep=0.8mm
    \renewcommand\arraystretch{1.5}
    \caption{Short-term forecasting results in the M4 dataset with a single variate. All prediction lengths are in $\left[ 6,48 \right]$. A lower SMAPE, MASE or OWA indicates a better prediction. }\label{tab:short_m4}
    \vspace{0.2em}
    \resizebox{\textwidth}{!}{
    \begin{tabular}{cc|c|cccccccccccccccccc}
    \toprule
    \hline
    
    \multicolumn{3}{c}{\multirow{2}{*}{Models}} & 
    \multicolumn{1}{c}{\rotatebox{0}{\scalebox{0.95}{\textbf{\myformer}}}}&
    \multicolumn{1}{c}{\rotatebox{0}{\scalebox{0.95}{{TimeMixer}}}}&
    \multicolumn{1}{c}{\rotatebox{0}{\scalebox{0.95}{{iTransformer}}}}&
    \multicolumn{1}{c}{\rotatebox{0}{\scalebox{0.95}{{TiDE}}}}&
    \multicolumn{1}{c}{\rotatebox{0}{\scalebox{0.95}{TimesNet}}} &
    \multicolumn{1}{c}{\rotatebox{0}{\scalebox{0.95}{{N-HiTS}}}} &
    \multicolumn{1}{c}{\rotatebox{0}{\scalebox{0.95}{{N-BEATS}}}} &
    \multicolumn{1}{c}{\rotatebox{0}{\scalebox{0.95}{PatchTST}}} &
    \multicolumn{1}{c}{\rotatebox{0}{\scalebox{0.95}{MICN}}} &
    \multicolumn{1}{c}{\rotatebox{0}{\scalebox{0.95}{FiLM}}} &
    \multicolumn{1}{c}{\rotatebox{0}{\scalebox{0.95}{LightTS}}} &
    \multicolumn{1}{c}{\rotatebox{0}{\scalebox{0.95}{DLinear}}} &
    \multicolumn{1}{c}{\rotatebox{0}{\scalebox{0.95}{FED.}}} & 
    \multicolumn{1}{c}{\rotatebox{0}{\scalebox{0.95}{Stationary}}} & 
    \multicolumn{1}{c}{\rotatebox{0}{\scalebox{0.95}{Auto.}}} \\
\end{tabular}
    } 
\end{table*}

\begin{table*}[htbp]
    \centering
    \small
    \tabcolsep=0.8mm
    \renewcommand\arraystretch{1.5}
    \caption{Short-term forecasting results in the PEMS datasets with multiple variates. All input lengths are 96 and prediction lengths are 12. A lower MAE, MAPE or RMSE indicates a better prediction.}\label{tab:short_pems}
    \vspace{0.2em}
    \resizebox{0.9\textwidth}{!}{
    \begin{tabular}{cc|c|cccccccccccccccccc}
    \toprule
    \hline
    
    \multicolumn{2}{c}{\multirow{2}{*}{Models}} &
    \multicolumn{1}{c}{\rotatebox{0}{\scalebox{0.95}{\textbf{\myformer}}}}&
    \multicolumn{1}{c}{\rotatebox{0}{\scalebox{0.95}{TimeMixer}}} &
    \multicolumn{1}{c}{\rotatebox{0}{\scalebox{0.95}{iTransformer}}} &
    \multicolumn{1}{c}{\rotatebox{0}{\scalebox{0.95}{TiDE}}} &
    \multicolumn{1}{c}{\rotatebox{0}{\scalebox{0.95}{SCINet}}} &
    \multicolumn{1}{c}{\rotatebox{0}{\scalebox{0.95}{Crossformer}}} &
    \multicolumn{1}{c}{\rotatebox{0}{\scalebox{0.95}{PatchTST}}} &
    \multicolumn{1}{c}{\rotatebox{0}{\scalebox{0.95}{TimesNet}}} &
    \multicolumn{1}{c}{\rotatebox{0}{\scalebox{0.95}{MICN}}} &
    \multicolumn{1}{c}{\rotatebox{0}{\scalebox{0.95}{DLinear}}} &
    \multicolumn{1}{c}{\rotatebox{0}{\scalebox{0.95}{FEDformer}}} & 
    \multicolumn{1}{c}{\rotatebox{0}{\scalebox{0.95}{Stationary}}} & 
    \multicolumn{1}{c}{\rotatebox{0}{\scalebox{0.95}{Autoformer}}} \\
    
    \multicolumn{2}{c}{}&\multicolumn{1}{c}{\scalebox{0.95}{\textbf{(Ours)}}} &
    \multicolumn{1}{c}{\scalebox{0.95}{\cite{Wang2024TimeMixerDM}}}&
    \multicolumn{1}{c}{\scalebox{0.95}{\cite{liu2023itransformer}}}&
    \multicolumn{1}{c}{\scalebox{0.95}{\cite{Das2023LongtermTiDE}}}&
    \multicolumn{1}{c}{\scalebox{0.95}{\cite{Liu2021SCINetTS}}} &
    \multicolumn{1}{c}{\scalebox{0.95}{\cite{Du2021CrossDomainGD}}} &
    \multicolumn{1}{c}{\scalebox{0.95}{\cite{Nie2023PatchTST}}} &
    \multicolumn{1}{c}{\scalebox{0.95}{\cite{wu2023timesnet}}} &
    \multicolumn{1}{c}{\scalebox{0.95}{\cite{wang2023micn}}} &
    \multicolumn{1}{c}{\scalebox{0.95}{\cite{Zeng2022DLinear}}} &
    \multicolumn{1}{c}{\scalebox{0.95}{\cite{zhou2022fedformer}}} &
    \multicolumn{1}{c}{\scalebox{0.95}{\cite{Liu2022NonstationaryTR}}}&
    \multicolumn{1}{c}{\scalebox{0.95}{\cite{wu2021autoformer}}} \\
\hline
    \bottomrule
    \end{tabular}
    } 
    \vspace{-0.5cm}
\end{table*}

\begin{table*}[htbp]
    \centering
    \small
    \tabcolsep=0.8mm
    \renewcommand\arraystretch{1.5}
    \caption{Full results for the anomaly detection task. The P, R and F1 represent the precision, recall and F1-score (\%) respectively. F1-score is the harmonic mean of precision and recall. A higher value of P, R and F1 indicates a better performance.}\label{tab:anomaly}
    \vspace{0.2em}
    \resizebox{1.0\textwidth}{!}{
    \begin{tabular}{cc|ccccccccccccccccccc}
    \toprule
    \hline
    
    \multicolumn{2}{c}{\scalebox{0.8}{Datasets}} & 
    \multicolumn{3}{c}{\scalebox{0.8}{\rotatebox{0}{SMD}}} &
    \multicolumn{3}{c}{\scalebox{0.8}{\rotatebox{0}{MSL}}} &
    \multicolumn{3}{c}{\scalebox{0.8}{\rotatebox{0}{SMAP}}} &
    \multicolumn{3}{c}{\scalebox{0.8}{\rotatebox{0}{SWaT}}} & 
    \multicolumn{3}{c}{\scalebox{0.8}{\rotatebox{0}{PSM}}} & \scalebox{0.8}{Avg F1} \\

    \cline{3-18}
    
    \multicolumn{2}{c}{\scalebox{0.8}{Metrics}} & 
    \scalebox{0.8}{P} & \scalebox{0.8}{R} & \scalebox{0.8}{F1} & 
    \scalebox{0.8}{P} & \scalebox{0.8}{R} & \scalebox{0.8}{F1} & 
    \scalebox{0.8}{P} & \scalebox{0.8}{R} & \scalebox{0.8}{F1} & 
    \scalebox{0.8}{P} & \scalebox{0.8}{R} & \scalebox{0.8}{F1} & 
    \scalebox{0.8}{P} & \scalebox{0.8}{R} & \scalebox{0.8}{F1} & 
    \scalebox{0.8}{(\%)}\\

    \hline
    
    \scalebox{0.85}{LSTM} &
    \scalebox{0.85}{\cite{Hochreiter1997LongSM}} %      
    & \scalebox{0.85}{78.52} & \scalebox{0.85}{65.47} & \scalebox{0.85}{71.41} 
    & \scalebox{0.85}{78.04} & \scalebox{0.85}{86.22} & \scalebox{0.85}{81.93}
    & \scalebox{0.85}{91.06} & \scalebox{0.85}{57.49} & \scalebox{0.85}{70.48} 
    & \scalebox{0.85}{78.06} & \scalebox{0.85}{91.72} & \scalebox{0.85}{84.34} 
    & \scalebox{0.85}{69.24} & \scalebox{0.85}{99.53} & \scalebox{0.85}{81.67}
    & \scalebox{0.85}{77.97} \\ 
    
    \scalebox{0.85}{Transformer} &
    \scalebox{0.85}{\cite{Vaswani2017AttentionIA}} %   
    & \scalebox{0.85}{83.58} & \scalebox{0.85}{76.13} & \scalebox{0.85}{79.56} 
    & \scalebox{0.85}{71.57} & \scalebox{0.85}{87.37} & \scalebox{0.85}{78.68}
    & \scalebox{0.85}{89.37} & \scalebox{0.85}{57.12} & \scalebox{0.85}{69.70} 
    & \scalebox{0.85}{68.84} & \scalebox{0.85}{96.53} & \scalebox{0.85}{80.37}  %%
    & \scalebox{0.85}{62.75} & \scalebox{0.85}{96.56} & \scalebox{0.85}{76.07}
    & \scalebox{0.85}{76.88} \\
\end{tabular}
    } 
    \vspace{-0.5cm}
\end{table*}

\begin{table*}[htbp]
    \centering
    \small
    \tabcolsep=0.0mm
    \renewcommand\arraystretch{1.7}
    \caption{Full results for the classification task. We report the classification accuracy (\%) as the result. The standard deviation is within 0.1\%.}\label{tab:classification}
    \vspace{0.2em}
    \resizebox{\textwidth}{!}{
    \begin{tabular}{c|ccccccccccccccccccccccccccccccccccc}
    \toprule
    \hline

    \multirow{3}{*}{{Datasets/Models}} & 
    \multicolumn{3}{c}{{Classical methods}} & 
    \multicolumn{3}{c}{{RNN}} &
    \multicolumn{1}{c}{{TCN}} &
    \multicolumn{10}{c}{{Transformers}} & 
    \multicolumn{3}{c}{{MLP}} & 
    \multicolumn{1}{c}{{CNN}}\\
    
    \cmidrule(lr){2-4}\cmidrule(lr){5-7}\cmidrule(lr){8-8}\cmidrule(lr){9-18}\cmidrule(lr){19-21}\cmidrule(lr){22-22} &
    
    \scalebox{0.8}{DTW} & \scalebox{0.8}{XGBoost} & \scalebox{0.8}{Rocket} & \scalebox{0.8}{LSTM} & \scalebox{0.8}{LSTNet} & 
    \scalebox{0.8}{LSSL} & \scalebox{0.8}{TCN} & \scalebox{0.9}{Trans.} & \scalebox{0.9}{Re.} & \scalebox{0.9}{In.} & 
    \scalebox{0.9}{Pyra.} & \scalebox{0.9}{Auto.} & \scalebox{0.9}{Stat.} & \scalebox{0.9}{FED.} & \scalebox{0.9}{ETS.} & 
    \scalebox{0.9}{Flow.} & \scalebox{0.8}{iTrans.} & \scalebox{0.8}{DLinear} & \scalebox{0.8}{LightTS.} & 
    \scalebox{0.8}{TiDE} & \scalebox{0.8}{TimesNet} & {\textbf{\myformer}} \\
\hline
    \bottomrule
    \end{tabular}
    } 
    \vspace{-0.5cm}
\end{table*}

\begin{table*}[htbp]
    \centering
    \small
    \tabcolsep=0.6mm
    \renewcommand\arraystretch{1.7}
    \caption{forecasting enhanced.}\label{tab:forecasting_enhanced}
    \vspace{0.2em}
    \resizebox{\textwidth}{!}{
    \begin{tabular}{cc|cc|cc|cc||cc|cc|cc||cc|cc|cc||cc|cc|cc}
    \toprule
    \hline
    \multicolumn{2}{c}{\multirow{2}{*}{Methods}} &
    \multicolumn{6}{c}{\scalebox{1.2}{PatchTST~\cite{Nie2023PatchTST}}} & 
    \multicolumn{6}{c}{\scalebox{1.2}{TimeMixer~\cite{Wang2024TimeMixerDM}}} & 
    \multicolumn{6}{c}{\scalebox{1.2}{DLinear~\cite{Zeng2022DLinear}}} & 
    \multicolumn{6}{c}{\scalebox{1.2}{TimesNet~\cite{wu2023timesnet}}} \\
    
    \cline{3-26} 
    
    \multicolumn{2}{c}{Metric} &
    \multicolumn{2}{c|}{Original} & 
    \multicolumn{2}{c|}{\textbf{Enhance}} & 
    \multicolumn{2}{c||}{\textbf{Improve}} & 
    \multicolumn{2}{c|}{Original} & 
    \multicolumn{2}{c|}{\textbf{Enhance}} & 
    \multicolumn{2}{c||}{\textbf{Improve}} & 
    \multicolumn{2}{c|}{Original} & 
    \multicolumn{2}{c|}{\textbf{Enhance}} & 
    \multicolumn{2}{c||}{\textbf{Improve}} & 
    \multicolumn{2}{c|}{Original} & 
    \multicolumn{2}{c|}{\textbf{Enhance}} & 
    \multicolumn{2}{c}{\textbf{Improve}} \\

    \cline{3-26} 
    
    \multicolumn{2}{c}{} & 
    MSE & MAE & MSE & MAE & MSE & MAE & MSE & MAE & 
    MSE & MAE & MSE & MAE & MSE & MAE & MSE & MAE & MSE & MAE & MSE & MAE & MSE & MAE & MSE & MAE \\
    
    \hline 
    \hline
    
    \multicolumn{1}{c|}{\multirow{5}{*}{\rotatebox{90}{Weather}}} & 96 & 
    {0.186} & {0.227} &
    \textbf{0.146} & \textbf{0.194} &
    \red{\textbf{21.5\%}} & \red{\textbf{14.5\%}} &
    {0.163} & {0.209} &
    \textbf{0.147} & \textbf{0.199} &
    \red{\textbf{9.8\%}} & \red{\textbf{4.8\%}} &
    {0.195} & {0.252} &
    \textbf{0.173} & \textbf{0.222} &
    \red{\textbf{13.5\%}} & \red{\textbf{11.9\%}} &
    {0.172} & {0.22} &
    \textbf{0.156} & \textbf{0.195} &
    \red{\textbf{9.2\%}} & \red{\textbf{11.4\%}} \\
\hline 
    \bottomrule
    \end{tabular}
    } 
    \vspace{-0.5cm}
\end{table*}

\section{Showcases}\label{appendix:showcase}
\begin{figure*}[t]
\begin{center}
	\centerline{\includegraphics[width=2.0\columnwidth]{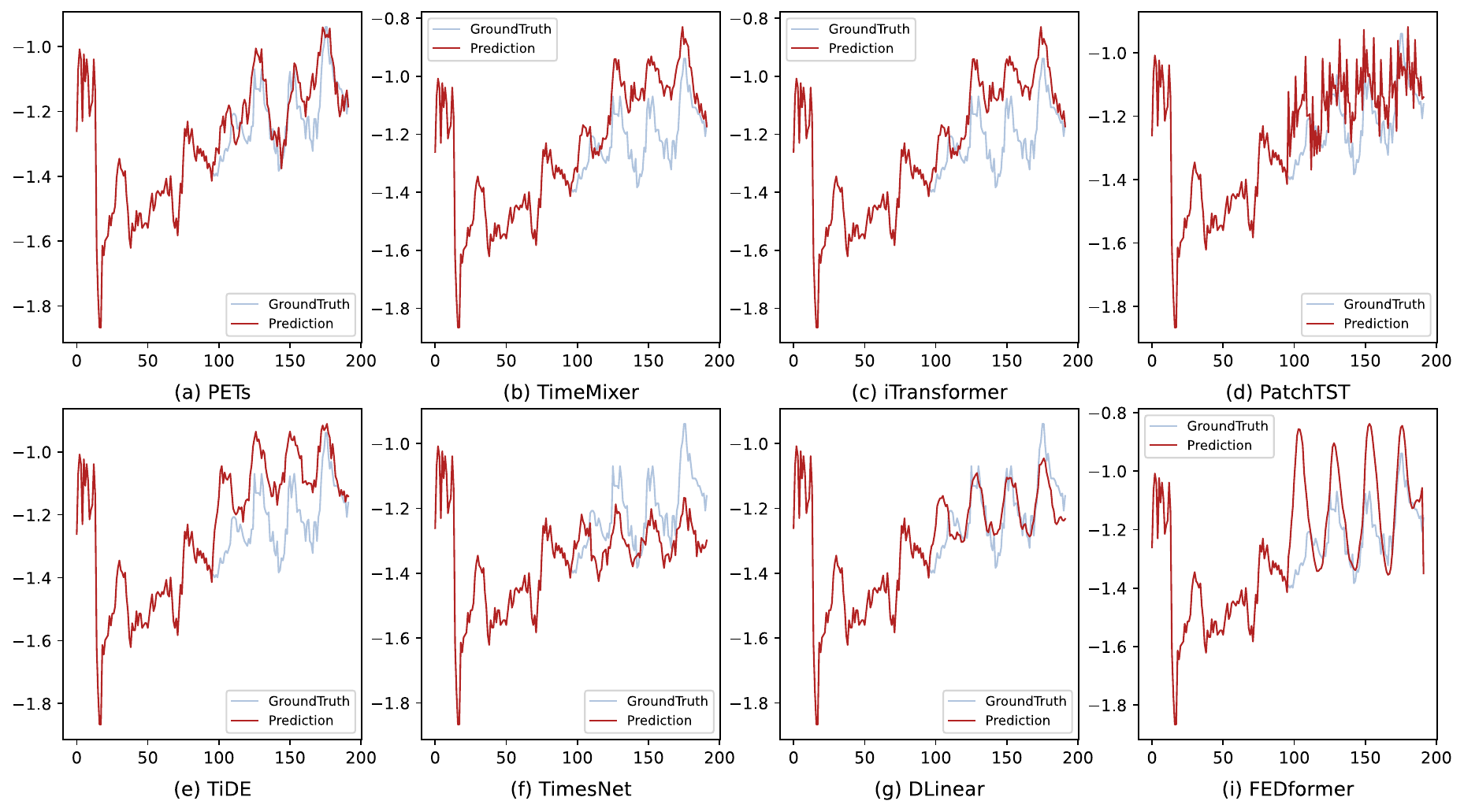}}
    \vspace{10pt}
	\centerline{\includegraphics[width=2.0\columnwidth]{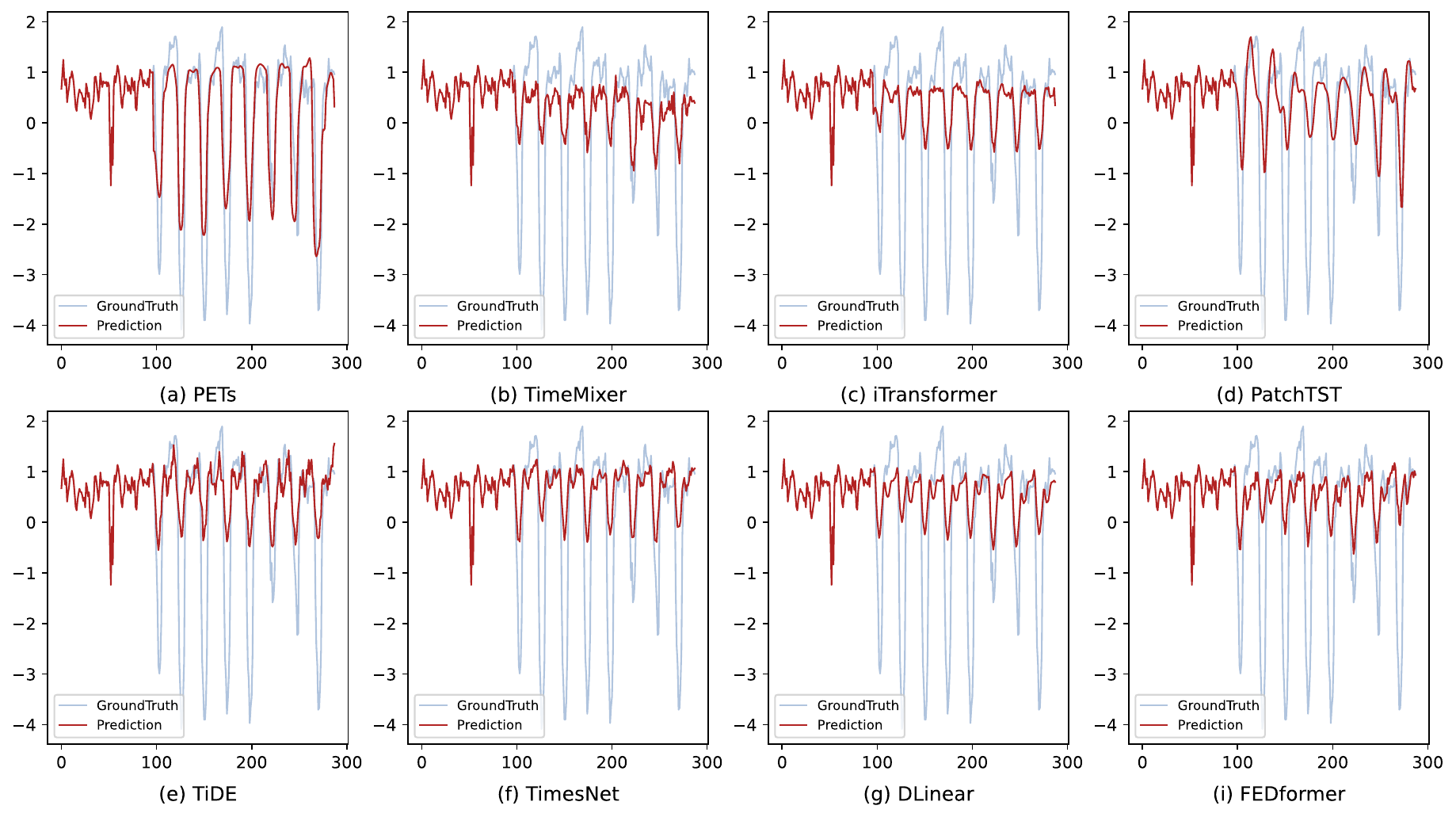}}
	\caption{
Prediction cases from \textbf{ETTh1} by different models under the input-96-predict-96 settings. 
}\label{fig:results_etth1_96_96}
\end{center}
\end{figure*}

\begin{figure*}[t]
\begin{center}
	\centerline{\includegraphics[width=2.0\columnwidth]{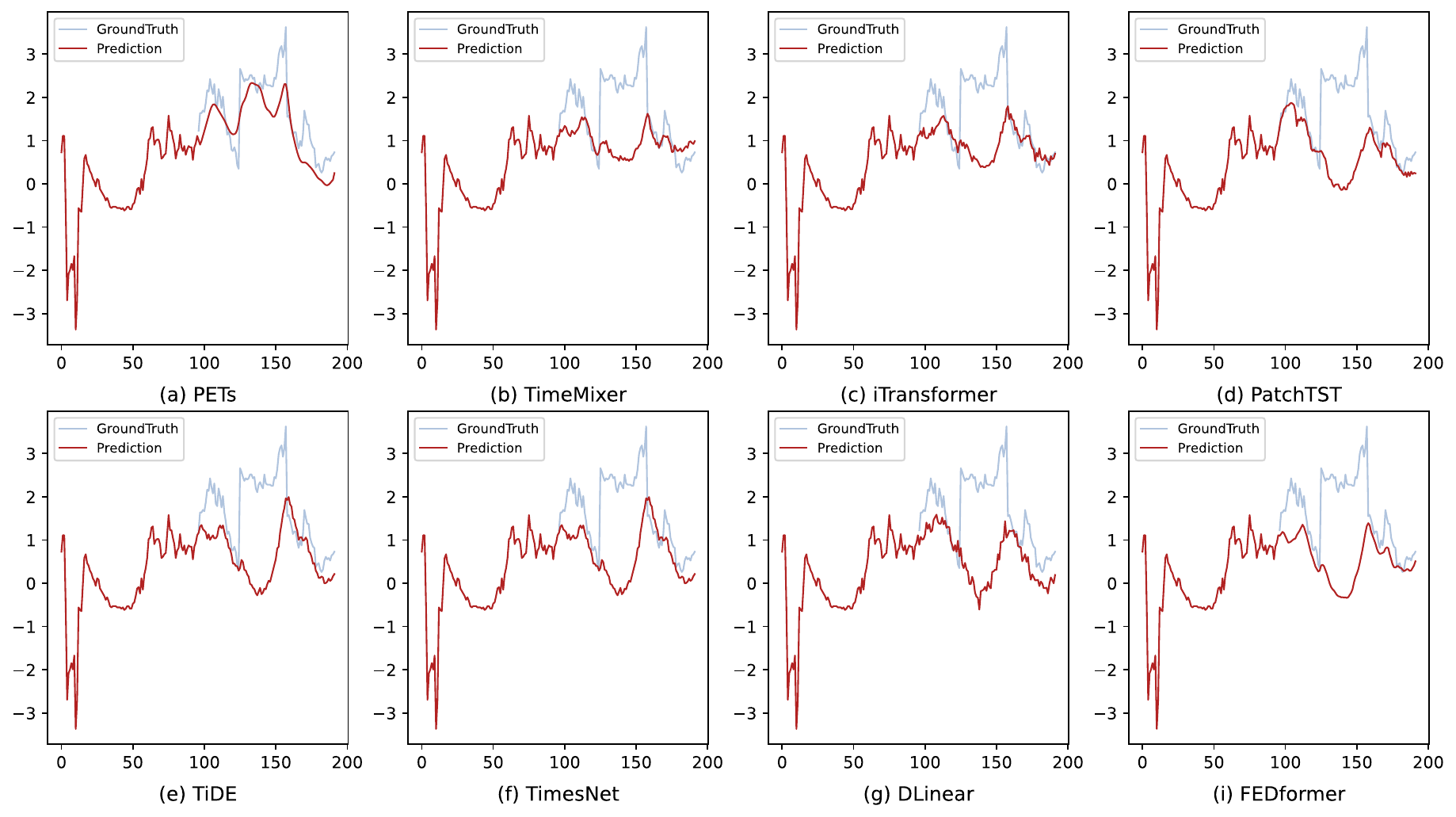}}
    \vspace{10pt}
	\centerline{\includegraphics[width=2.0\columnwidth]{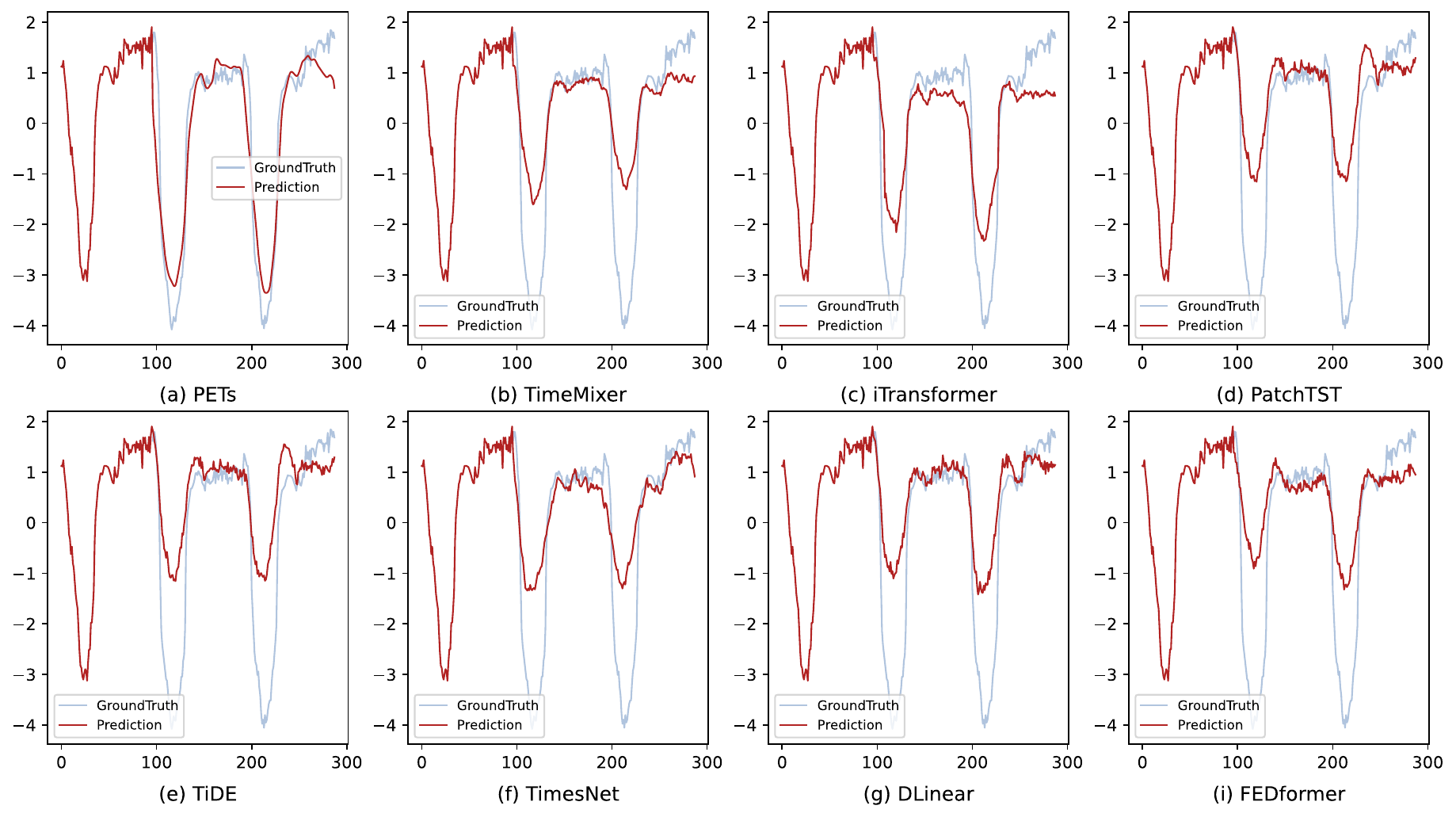}}
	\caption{
Prediction cases from \textbf{ETTm1} by different models under the input-96-predict-96 settings. 
}\label{fig:results_ettm1_96_96}
\end{center}
\end{figure*}

\begin{figure*}[t]
\begin{center}
	\centerline{\includegraphics[width=2.0\columnwidth]{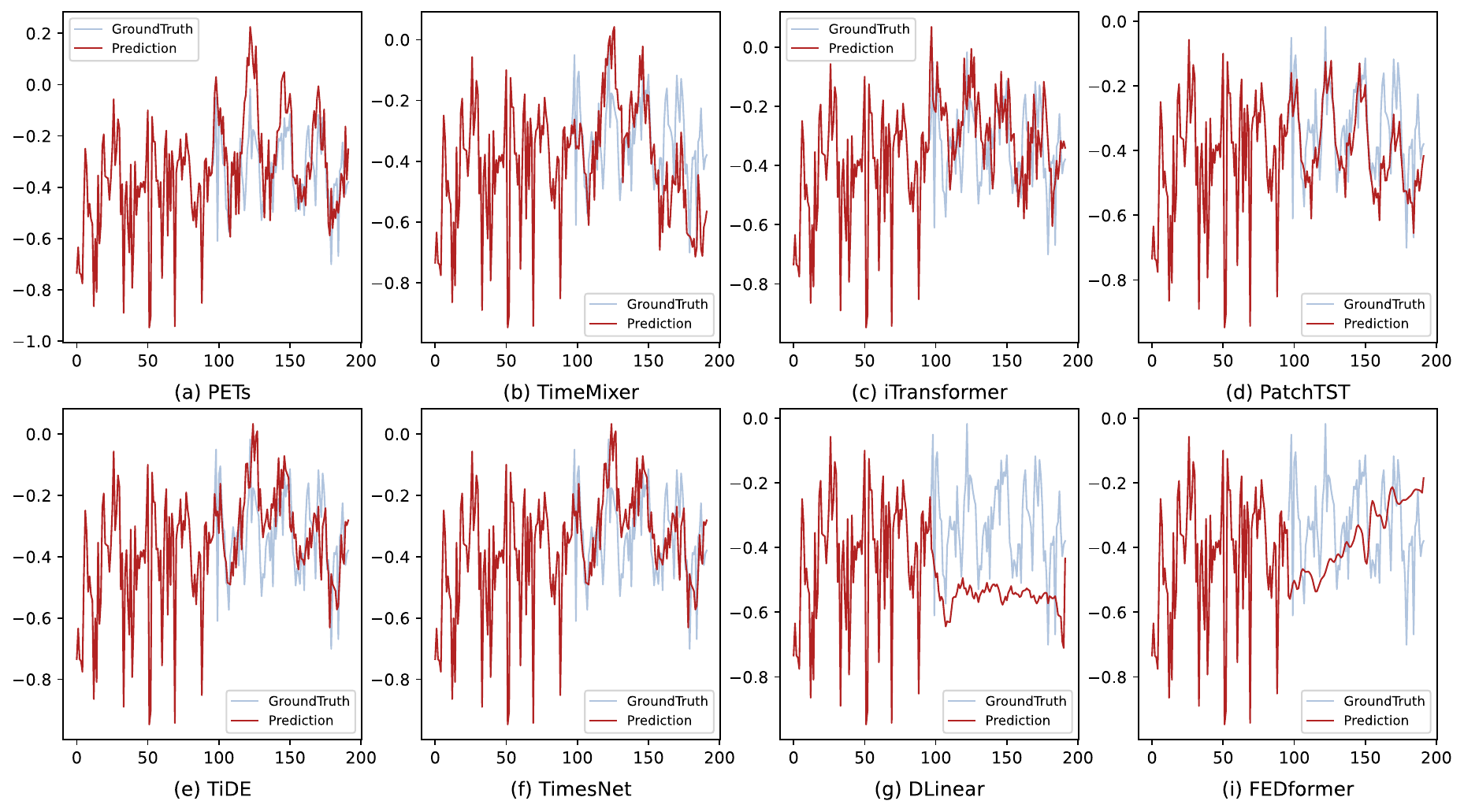}}
    \vspace{10pt}
	\centerline{\includegraphics[width=2.0\columnwidth]{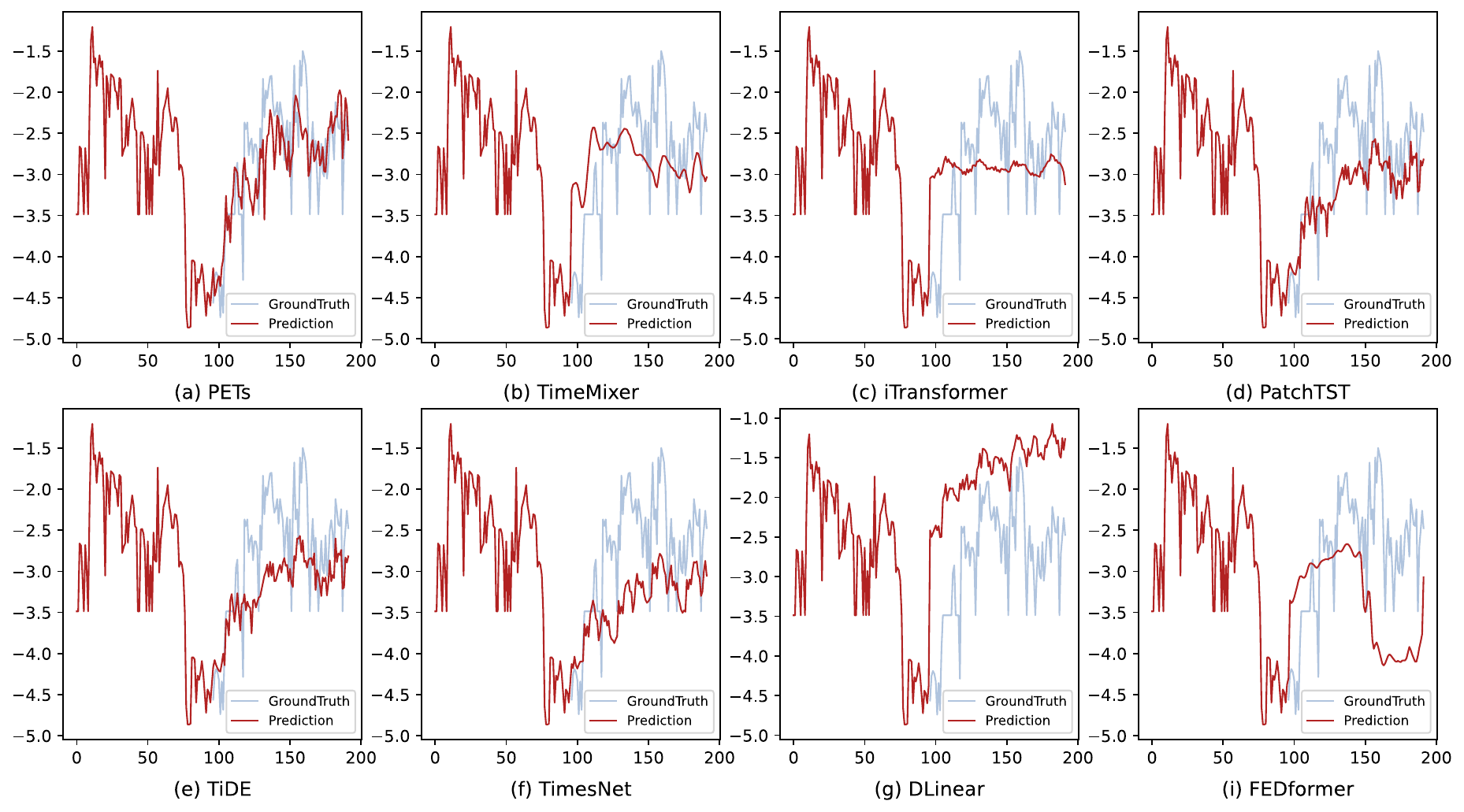}}
	\caption{
Prediction cases from \textbf{ETTh2} by different models under the input-96-predict-96 settings. 
}\label{fig:results_etth2_96_96}
\end{center}
\end{figure*}

\begin{figure*}[t]
\begin{center}
	\centerline{\includegraphics[width=2.0\columnwidth]{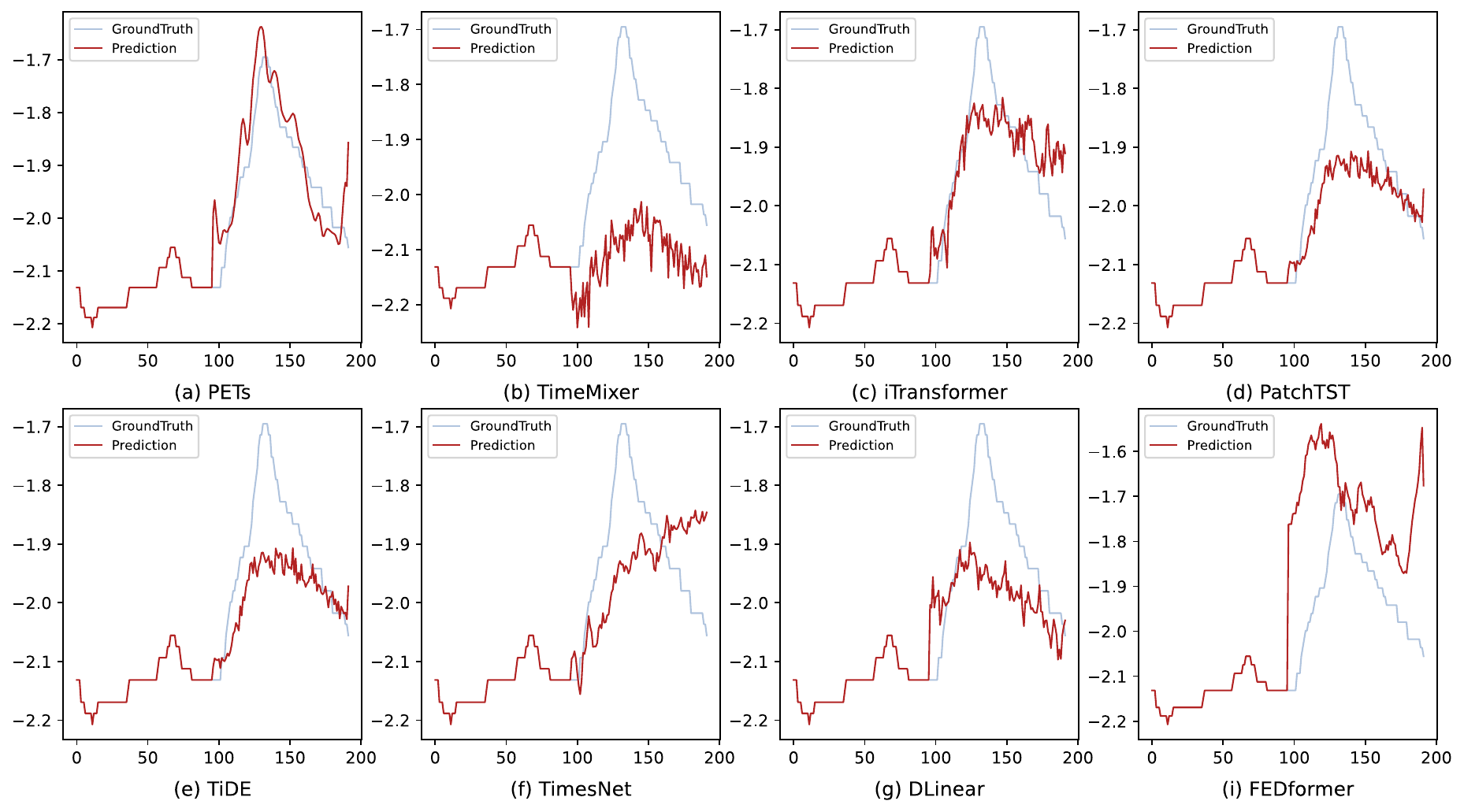}}
    \vspace{10pt}
	\centerline{\includegraphics[width=2.0\columnwidth]{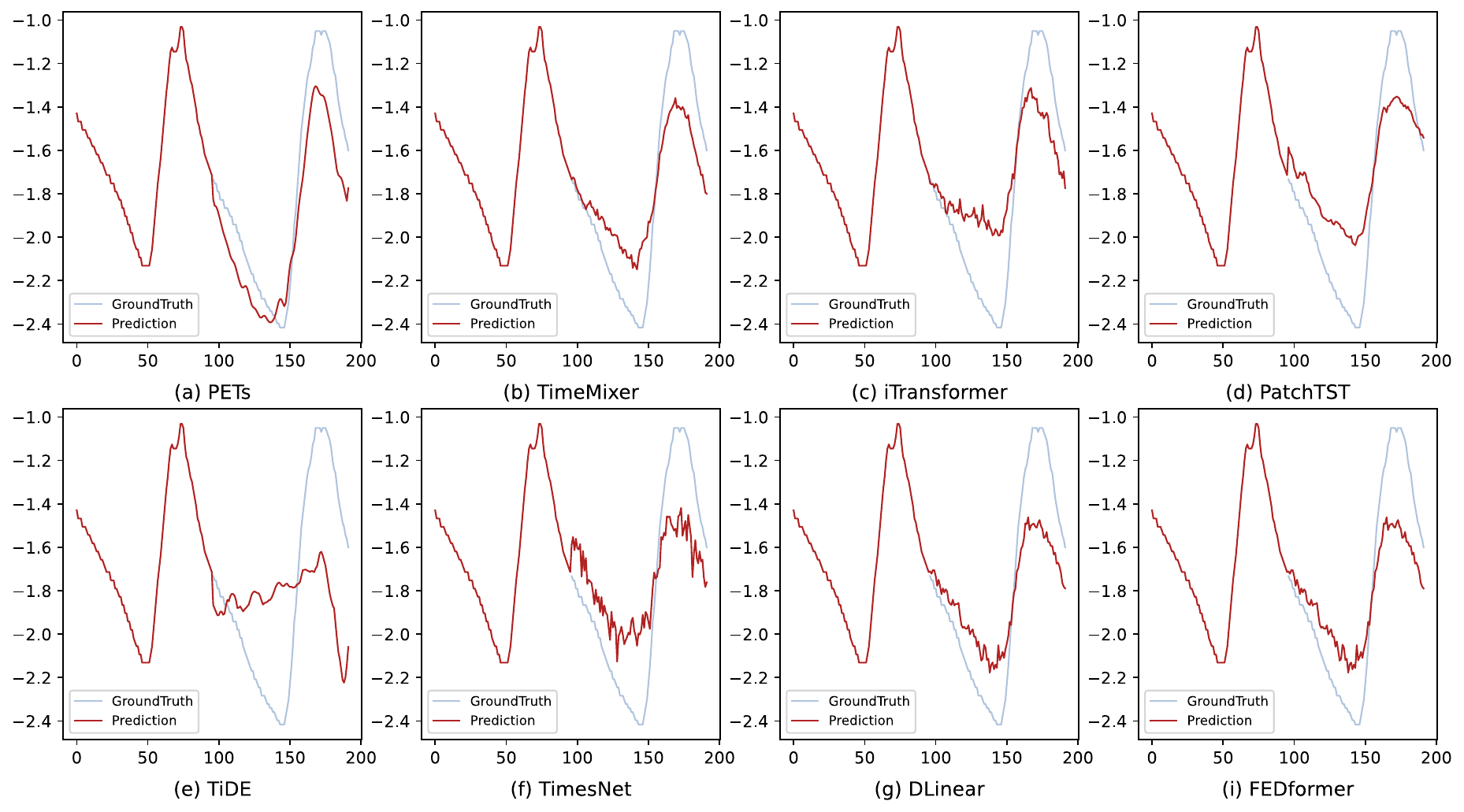}}
	\caption{
Prediction cases from \textbf{ETTm2} by different models under the input-96-predict-96 settings. 
}\label{fig:results_ettm2_96_96}
\end{center}
\end{figure*}

\begin{figure*}[t]
\begin{center}
	\centerline{\includegraphics[width=2.0\columnwidth]{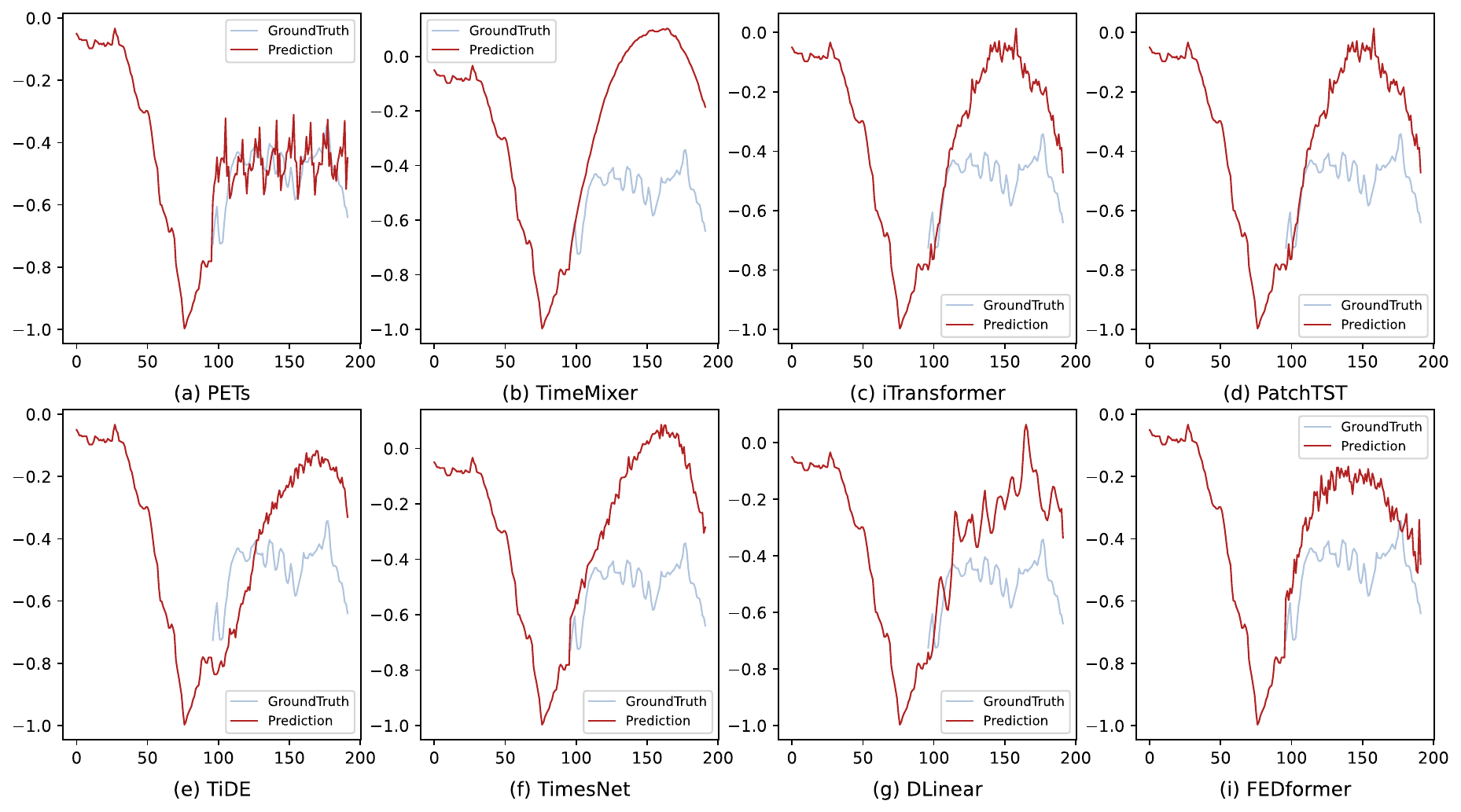}}
    \vspace{10pt}
	\centerline{\includegraphics[width=2.0\columnwidth]{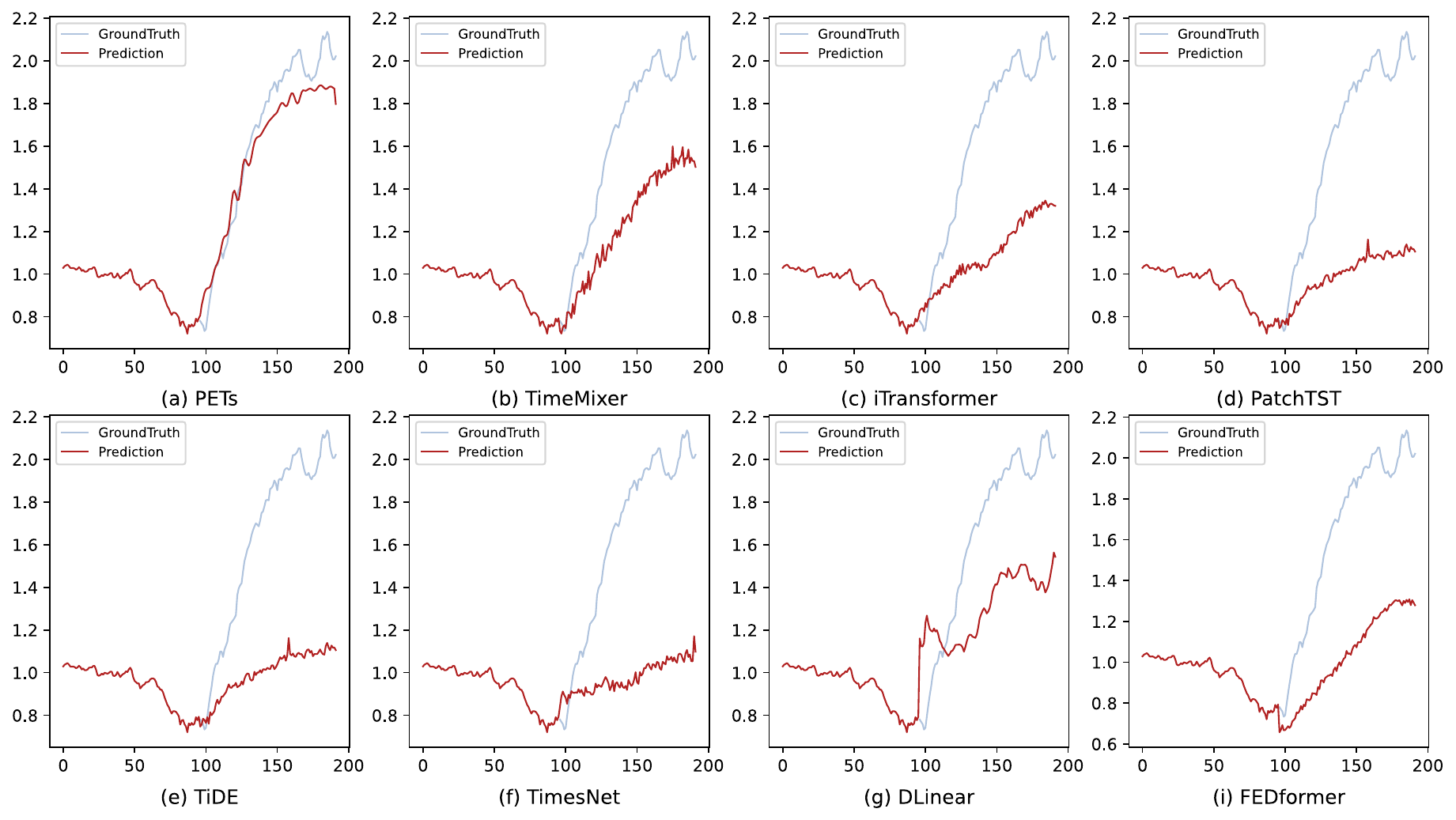}}
	\caption{
Prediction cases from \textbf{Weather} by different models under the input-96-predict-96 settings. 
}\label{fig:results_weather_96_96}
\end{center}
\end{figure*}

\begin{figure*}[t]
\begin{center}
	\centerline{\includegraphics[width=2.0\columnwidth]{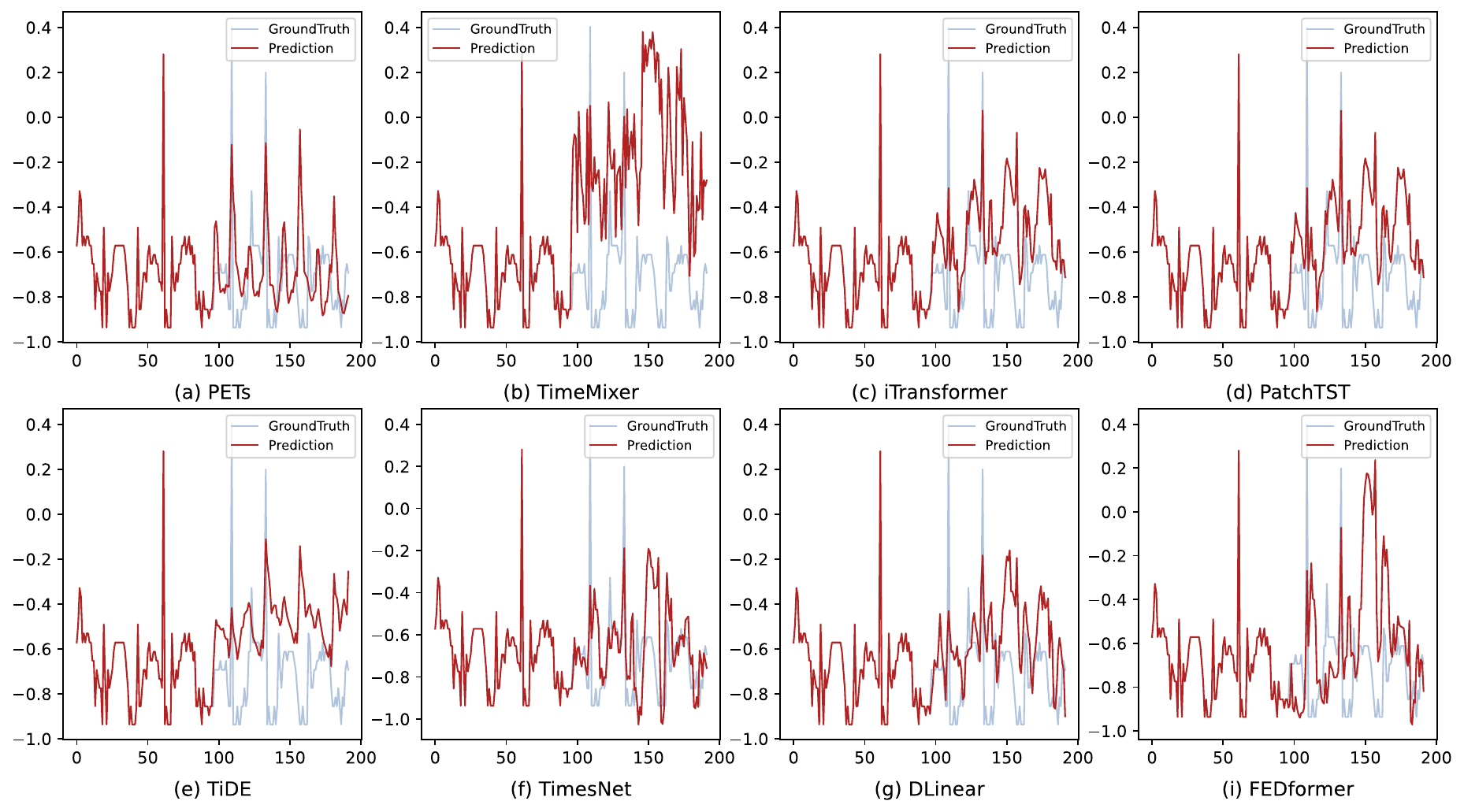}}
    \vspace{10pt}
	\centerline{\includegraphics[width=2.0\columnwidth]{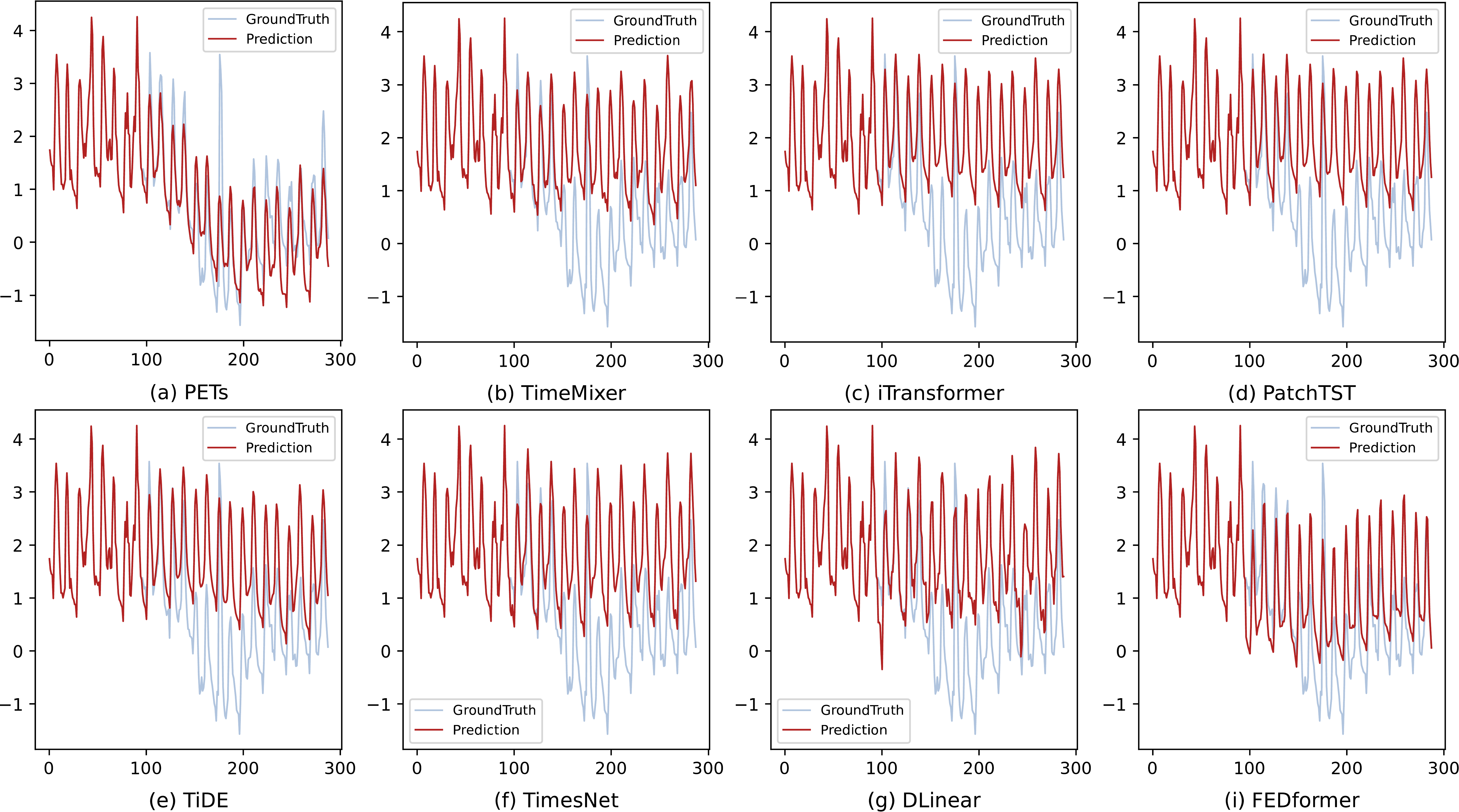}}
	\caption{
Prediction cases from \textbf{Electricity} by different models under the input-96-predict-96 settings. 
}\label{fig:results_ecl_96_96}
\end{center}
\end{figure*}

\begin{figure*}[t]
\begin{center}
	\centerline{\includegraphics[width=2.0\columnwidth]{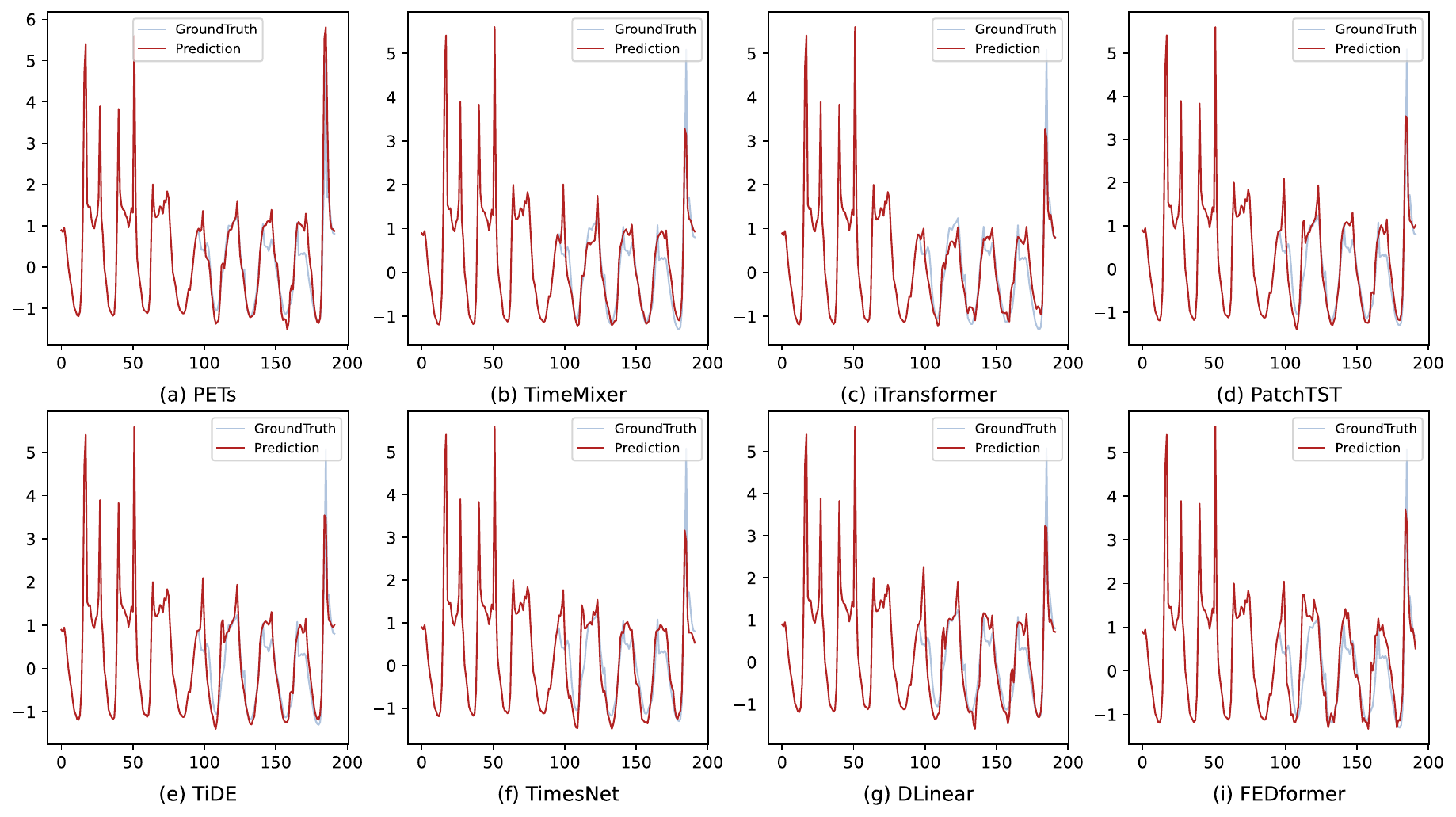}}
    \vspace{10pt}
	\centerline{\includegraphics[width=2.0\columnwidth]{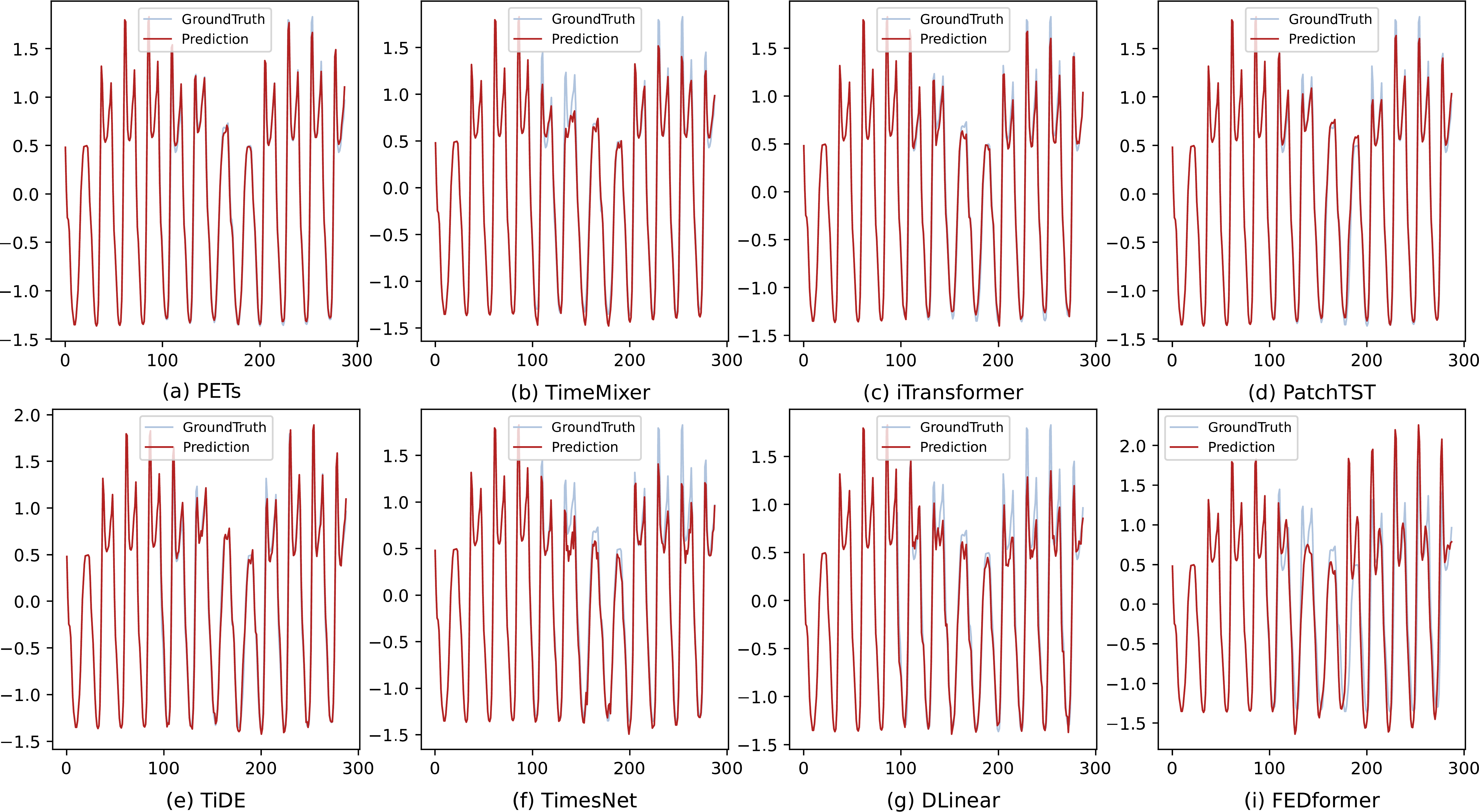}}
	\caption{
Prediction cases from \textbf{Traffic} by different models under the input-96-predict-96 settings. 
}\label{fig:results_traffic_96_96}
\end{center}
\end{figure*}

To assess the performance of various models, we perform a qualitative comparison by visualizing the final dimension of the forecasting results derived from the test set of each dataset (Fig.~\ref{fig:results_etth1_96_96},~\ref{fig:results_ettm1_96_96},~\ref{fig:results_etth2_96_96},~\ref{fig:results_ettm2_96_96},~\ref{fig:results_weather_96_96},~\ref{fig:results_ecl_96_96},~\ref{fig:results_traffic_96_96}). Among the various models, \myformer exhibits superior performance.

\vfill

\end{document}